%% file: rfm_change_detection_preprints.tex
\documentclass[pdftex,10pt,a4paper]{article}
\usepackage{afterpage}
\usepackage{amsmath}
\usepackage{amsfonts}
\usepackage{amssymb}
\usepackage{txfonts} 
\DeclareMathAlphabet{\mathpzc}{OT1}{pzc}{m}{it} 
\usepackage{bbm,bbold}
\usepackage{mathptmx}       
\usepackage{makeidx}         
\usepackage{graphicx}        
\graphicspath{{./}}
\usepackage{multicol}        
\usepackage{acronym}
\usepackage{multirow}
\usepackage{color, soul}

\usepackage[caption=false, font=footnotesize]{subfig}
\usepackage{setspace}

\usepackage{footnote}
\usepackage{tablefootnote}
\makesavenoteenv{tabular}
\makesavenoteenv{table}

\usepackage{nameref}
\usepackage{natbib}
\DeclareMathAlphabet{\mathcal}{OMS}{cmsy}{m}{n}
\SetMathAlphabet{\mathcal}{bold}{OMS}{cmsy}{b}{n}
\usepackage{adjustbox}
 
\title{Feature-wise change detection and robust indoor positioning using \acs{ransac}-like approach}

\author{Caifa Zhou (caifa.zhou@geod.baug.ethz.ch)}
\begin{document}
	\maketitle
\input{./cz_acronyms.tex}
\input{./cz_command.tex}


\begin{abstract}
	{Fingerprinting-based positioning, one of the promising indoor positioning solutions, has been broadly explored owing to the pervasiveness of sensor-rich mobile devices, the prosperity of opportunistically measurable location-relevant signals and the progress of data-driven algorithms. One critical challenge is to control and improve the quality of the reference fingerprint map (RFM), which is built at the offline stage and applied for online positioning. The key concept concerning the quality control of the RFM is updating the RFM according to the newly measured data. Though varies methods have been proposed for adapting the RFM, they approach the problem by introducing extra-positioning schemes (e.g. PDR or UGV) and directly adjust the RFM without distinguishing whether critical changes have occurred. This paper aims at proposing an extra-positioning-free solution by making full use of the redundancy of measurable features. Loosely inspired by random sampling consensus (RANSAC), arbitrarily sampled subset of features from the online measurement are used for generating multi-resamples, which are used for estimating the intermediate locations. In the way of resampling, it can mitigate the impact of the changed features on positioning and enables to retrieve accurate location estimation. The user’s location is robustly computed by identifying the candidate locations from these intermediate ones using modified Jaccard index (MJI) and the feature-wise change belief is calculated according to the world model of the RFM and the estimated variability of features. In order to validate our proposed approach, two levels of experimental analysis have been carried out. On the simulated dataset, the average change detection accuracy is about 90\%. Meanwhile, the improvement of positioning accuracy within 2~m is about 20\% by dropping out the features that are detected as changed when performing positioning comparing to that of using all measured features for location estimation. On the long-term collected dataset, the average change detection accuracy is about 85\%.}
	
\end{abstract}

\section{Introduction}\label{sec:intro}
Emerging from about two decades ago, indoor object/pedestrian tracking systems have been attracting the attention from both academia and industry. Especially in the community of academia, the researchers have endeavored for designing the indoor positioning system which is capable of yielding accurate localization services with low cost. Tens of different signal sources, such as ultrasound (\eg Active Bat \citep{addlesee2001implementing}, Dolphin \citep{minami2004dolphin}, or Cricket \citep{kolodziej2006local}), infrared (Active Badge \citep{want1992active}, \citep{Wang2017}), RF signals (\eg RFID \cite{hightower2000spoton,xu2017rfid}, UWB \citep{alarifi2016ultra}, WLAN \citep{Padmanabhan2000,Youssef2008, he2016wi}, BLE \citep{zhuang2016smartphone}, FM \citep{cong2018environmentally}, cellular networks \citep{schloemann2016toward}), machine visions \citep{guan2016vision}, magnetic field \citep{he2018geomagnetism}, inertial sensors \citep{elloumi2016indoor} and visible light \citep{vadeny2016vslam} have been explored and exploited to infer the position based on range, angular or feature-based positioning approaches \citep{mautz2012indoor}. These investigated systems can role as the replacement/supplement of GNSS when navigating in indoor environments. Though they can provide positioning services at varies accuracy, range and cost, the promising indoor positioning solution for pedestrians must require low cost (\eg the infrastructure or deployment burden), fairly good accuracy, and large range (or good scalability). Recent research trend reveals one type of favorable indoor positioning system, which is feature-based (\ie fingerprinting-based) indoor positioning system (\acs{fips}) owing to the popularity of mobile devices with abundant built-in and off-the-shelf sensors (\eg \acsp{imu} or WiFi/BLE module), the development of data processing and interpretation (\eg crowd sourcing and machine learning), and the availability of opportunistically measurable location-relevant signals \cite{he2016wi, pei2016survey}.

The challenges of establishing such an \acs{fips} also originate in the prevalence of varies sensed data from heterogeneous devices and from diverse opportunistic signals. An \acs{fips} consists of two phases: offline referencing phase and online positioning phase. At the referencing phase, the core work is to construct the world model for representing the relationship between the measurable features and the corresponding locations. It is used as the \acf{rfm} (a more general terminology than radio map) for online positioning, which matches the current measured features to the \acs{rfm} for inferring users' position. The labor cost of \acs{rfm} collection, the computation cost of positioning and the availability of positioning approaches are no longer the main bottlenecks which restrict the applications of \acsp{fips} benefiting from the crowdsourcing way of collecting the \acs{rfm}, the power of computational resource on mobile devices and the comprehensiveness of data-driven algorithms  \cite{he2016wi, pei2016survey}. One critical challenge is to improve the quality of the \acs{rfm}, including i) fusing the \acs{rfm} contributed by multi-users/devices \citep{pei2016survey}, and ii) keeping the \acs{rfm} up-to-date \citep{Peng2018}, in order to guarantee the positioning performance in the long-term.

In this paper, we focus on maintaining the \acs{rfm} and divide the \acs{rfm} maintenance task into two parts: i) detecting the changes occurred in the online measured features comparing to an available \acs{rfm}; and ii) incrementally adapting the \acs{rfm} according to the detected changes and update criteria. Our way of treating the \acs{rfm} maintenance differs from the previous publications (see \secPref\ref{sec:related}, \tabPref\ref{tab:sum_rfm_update}). Most of them have purely addressed the latter, \ie adapting the \acs{rfm} via some regression or interpolation approaches (\eg Kriging or \acf{gpr}) using repetitively newly collected or crowd-sourced data without taking how the newly measured features differ from those of in the available \acs{rfm} into account. In this way, the quality of the updated \acs{rfm} cannot be ensured. In addition, most of their proposals depend on the extra indoor positioning technique for collecting the data used for updating the  \acs{rfm}. From this perspective, we recognize the dilemma that i) introducing extra hardware for positioning is contradict to the low-cost characteristic of \acsp{fips}; and ii) existing of other indoor localization techniques makes the \acs{fips} looking redundant though one tendency in the development of indoor positioning systems is to combine several positioning technologies in order to provide sufficient localization performance. Therefore, we explore the solution for achieving the \acs{rfm} maintenance without the supervision of extra position information. I.e. we aim at proposing an unsupervised, more precisely a self-supervised approach.

Our extra-positioning-free solution includes two assumptions: i) the opportunistically measurable features are redundant; and ii) the critical changes have occurred to the minority part of all measurable features. The former assumption means that more than one combination of the features can achieve accurate positioning. The latter ensures that the estimated position is moderately good in case that the features have been partially changed. It is the base for detecting the changes by comparing the measured features to what represented by the world model of the \acs{rfm}. This idea is similar to what is defined as novelty detection \citep{pimentel2014review}, which is for recognizing that the difference between the real-time measured data and the data during training. Our proposal targets for feature-wisely detecting the changes and for recovering positioning performance. We thus can provide accurate positioning to the user when changes have been occurred and extract the change status of each feature, which can be used for cumulatively and tactically updating the \acs{rfm}.

Our proposal is loosely inspired by \acf{ransac} \citep{choi1997performance}. It starts with constructing sampled measurements by arbitrarily sampling a subset of the features from the online measured ones. These estimated intermediate locations are obtained by matching these sampled observations to the \acs{rfm}. Owing to the assumptions. we expect that the resample consisting of the subset of unchanged features can be located close to the ground truth of the user (though unknown) via a suitable fingerprint matching approach. We then use the similarity metric, \acf{mji}, for identifying the most probable candidate locations which are closest to the ground truth. In this way we thus can achieve good positioning accuracy. Meanwhile, the change belief of each feature is approximated by modeling the variability of features and by comparing to the online measurements with the expected one that is retrieved from the world model of the \acs{rfm} at the estimated location. These detected change status of features can be used for accumulatively and tactically updating the \acs{rfm} and for refining the online positioning. On the simulated dataset, we can detect different simulated changes and the average value of area under the \acf{roc} curve (\acs{auc}) is about 0.9, \ie the achievable change detection accuracy is about 90\%. Meanwhile, the positioning accuracy can be improved by dropping out features that are detected as change when performing the fingerprint matching with the \acs{rfm}. The improvement of positioning accuracy within 2~m and 4~m are up about 20\% and 10\% comparing to that of matching the full measurement to the \acs{rfm}. Regarding the long-term dataset \citep{mendoza2018long}, we use it for validating the generalization capability of our approach. Though there are no apparent dependences between the change detection performance and spatial and temporal variations on the long-term dataset, we have found out that our approach is capable of detecting the changes over time and the average \acs{auc} value is about 85\%.

The remaining of the paper is organized as follows: \secPref\ref{sec:related} describes the related progress of the development of \acsp{fips}, especially focusing on \acs{rfm} maintenance. The fundamentals and detailed analysis of the proposed feature-wise change detection and robust positioning are presented in \secPref\ref{sec:fundamentals} and \secPref\ref{sec:robust_fbp}, respectively. The configuration of experimental analysis and results are given in \secPref\ref{sec:experimentalResults}.

%
\section{Related work}\label{sec:related}
In this section, we describe the progress of \acsp{fips}, including positioning technologies and approaches, \acs{rfm} creation/generation, and \acs{rfm} maintenance. Regarding the first two topics, we refer the readers to several reviews and publications, which have comprehensively discussed them. Mautz has talked about indoor positioning technologies in \citep{mautz2012indoor} and more recent discussions about them can be found in \citep{he2016wi,brena2017evolution}. In \citep{retscher2016,TORRESSOSPEDRA20159263,8115922} the authors have given a comprehensive comparison of different feature-based indoor positioning algorithms using varies similarity/dissimilarity metrics. A relative complete review of the methods for \acs{rfm} creation/generation can be found in \eg \citep{he2016wi,Zhou2018}. We exclusively focus on the current advancement of \acs{rfm} maintenance in this section. 

We start the review of the \acs{rfm} maintenance work with clarifying the terminology. The most publications regarding this topic were published in past four years but several keywords are used: maintenance, adaptation, update, and renovation. The first three words are more frequently used than renovation and have been utilized exchangeable in the publications.

Taniuchi and Maekawa proposed a method for automatically update the \acs{rfm} based on \acf{pdr} using the \acsp{imu}' data from the mobile devices \citep{Taniuchi:2015:AUI:2737797.2667226}. In their proposal, it contains the change detection step for determining when the feature value have been critically changes in a sequence of measurements using \acf{bic}. Tang \etal~introduced a fast fingerprint collection platform using \acfp{ugv} with LiDAR and other sensors with \acf{slam} \citep{s150305311}. Both of them have not considered the maintenance scheme and replace the available \acs{rfm} with the newly measured one.  In \citep{7565565} the authors have come up with a system for detecting the altered signals based on affinity propagation clustering (APC) algorithm and for updating the altered features with new values using \acs{gpr}. The \acf{ldplm} was employed for estimating the parameters of the Gaussian process when applying \acs{gpr}.

In 2018, varies \acs{rfm} update schemes have been introduced. In \citep{8168245}, it has described a method based on the evolutionary algorithm for updating the available \acs{rfm} using the crowd-sourced data which has no ground truth. The updating process relies on \acs{ldplm} and the parameters of the RF propagation model are estimated using the unlabeled data and their positions which are obtained using the available \acs{rfm}. Sun \etal~has proposed the \acs{rfm} update using crowd-sourced data but the updating process is realized Voronoi-wisely using neural network-based regression \citep{Sun2018}. In \citep{10.1007/978-3-319-71470-7_1} and \citep{8533825}, the authors have approached the \acs{rfm} update task from other aspects. They focus on minimizing the workload of re-collecting the data for the update by optimally selecting the locations where the data collection should be performed using the evolutionary algorithm and on mitigating the influence of missing features based on imputation schemes. Kernelized \acf{svr} has been applied for adapting the newly measured features to the available \acs{rfm}. The work extended from \citep{s150305311}, Peng \etal~has applied Kriging approach for updating the \acs{rfm} by interpolating over the \acs{ugv} measured data \citep{Peng2018}. All above works have not discussed the detecting the changes that might have occurred. They directly update the \acs{rfm} using a selected regression approach using the newly collected data.

One obvious tendency of \acs{rfm} update is the utilization of the \acs{pdr} \citep{8003484,yang2018ekf,8445663,Zou2017}. Based on the measurements gained together with \acs{pdr}, different approaches (\eg \acf{plsr} \citep{8003484}, \acs{gpr} with \acf{ekf} \citep{yang2018ekf}, fireworks algorithm \citep{8445663}, or polynomial surface fitting mean (PSFM) method \citep{Zou2017}) have been used for adapting the \acs{rfm}.

To sum-up, our extra-positioning-free solution for achieving robust positioning and feature-wise change detecting approaches the \acs{rfm} update problem in a different perspective comparing to previous work. In this way, we can reduce the mutual dependence between varies sensing and positioning techniques and accomplish the \acs{rfm} task in the self-supervised way.

\begin{table}[!htb]
		\centering
		\caption{Summary of works related to \acs{rfm} update}
		\label{tab:sum_rfm_update}
		\begin{adjustbox}{angle=90}
		\begin{tabular}{ccccccc}
			\hline 
			&Is extra & Extra tech.  & Is detection & Detection algo. & Update appr. & Year \\ 
			\hline 
			\citep{Taniuchi:2015:AUI:2737797.2667226} & Yes & PDR  &  Yes  & \acs{bic}  & -- & \multirow{2}{*}{2015}  \\ 
			\citep{s150305311} & Yes & \shortstack{\acs{ugv} (LiDAR, \acsp{imu}) \\ with \acs{slam}} & No & -- & -- &\\ 
			\hline 
			 \citep{7565565} & No & -- & Yes & APC & \acs{ldplm} with \acs{gpr} & 2016 \\ 
			 \hline
			 \citep{Zou2017} & Yes & PDR & No & -- & PSFM with \acs{gpr} & 2017 \\ 
			\hline 
			\citep{8168245} & Yes & Crowd-sourcing & No & --  & \acs{ldplm} & \multirow{8}{*}{2018} \\
			\citep{Sun2018}& Yes & Crowd-sourcing  & No & --  & \shortstack{Voronoi diagram with \\neural network-based regression} &  \\ 
			\citep{10.1007/978-3-319-71470-7_1}& Yes & \shortstack{Re-collection at \\optimized locations} & No &--  & \acs{svr} &  \\ 
			\citep{8533825}& Yes & \shortstack{Re-collection \\with data imputation} & No & -- & \acs{svr} &  \\ 
			\citep{Peng2018}& Yes & \shortstack{\acs{ugv} (LiDAR, \acsp{imu}) \\ with \acs{slam}}& No & -- & Kriging &  \\ 
			\citep{8003484}& Yes & \acs{pdr}  & No & --  & \acs{plsr} &  \\ 
			\citep{yang2018ekf}&Yes & \acs{pdr}  & No & --  & \acs{ekf} with \acs{gpr} &  \\ 
			\citep{8445663}&Yes & \acs{pdr}  & No & --  & \shortstack{Fireworks algorithm\\ with \acs{gpr}} &  \\ 
			\hline 
		\end{tabular}
	\end{adjustbox}
\end{table}

\section{Fundamental statement}\label{sec:fundamentals}
We first describe the usage of notations in this paper then briefly discuss the proposed robust positioning and change detection pipeline. 
\subsection{Notation}\label{subsec:notation}
\begin{itemize}
	\item A normal letter (\eg $ a $ or $ M $) denotes a scalar.
	\item A blackboard bold  letter (\eg $ \setSym{g} $ or $ \setSym{O} $) denotes a set.
	\item Bold lowercase (\eg $ \vecSym{l} $)  denotes a vector. 
	\item A calligraphic letter (\eg $ \funSym{f} $ or $ \funSym{F} $) denotes a functional mapping.
	\item The superscript of a symbol is used for identifying different meaning a symbol.
	\item The subscript of a symbol denotes the indexing variable. 
\end{itemize}
\subsection{An overview of the proposal}\label{subsec:fbpProb}
Herein we formulate the symbols from the perspective of the measured features. Each measured feature is uniquely identifiable and has a measured value, \eg an \acf {ap} can be identified by the \acf{mac} and has a \acf{rss}. It is thus formulated as a pair of attribute $ a $ and value $ v $, \ie $ (a, v) $. A measurement (\ie fingerprint) $ \setSymScript{O}{i}{u} $ taken by user $ \mathrm{u} $ at location/time $ i $ consists of a set of paired measured features, \ie $ \setSymScript{O}{i}{u} := \{(\norSymScript{a}{ik}{u}, \norSymScript{v}{ik}{u}) | \norSymScript{a}{ik}{u}\in\setSym{A}; \norSymScript{v}{ik}{u}\in\convSetSym{R};k\in\intSet{\norSymScript{N}{i}{u}} \} $,  where $ \setSym{A} $ is the complete set of the identifiers of all available features and $ N_i^{\texSym{u}} $ ($ N_i^{\texSym{u}} = |\setSymScript{O}{i}{u}|$) is the number of features observed by $ \texSym{u} $ at $ i $. A set of keys of $ \setSymScript{O}{i}{u} $ is defined as $ \setSymScript{A}{i}{u} := \{a_{ik}^{\texSym{u}}|\exists (a_{ik} ^{\texSym{u}}, v_{ik})\in \setSymScript{O}{i}{u}\} $ ($ \setSymScript{A}{i}{u} \subseteq \setSym{A}$). The positioning process consists of inferring the estimated user location $ \vecSymScript{\hat{l}}{i}{u}=\funSym{f}(\setSymScript{O}{i}{u}) (\vecSymScript{\hat{l}}{i}{u}\in\convSetSym{R}^{d}) $ as a function of the fingerprint and the \acs{rfm}, where $\funSym{f}$ is a suitable mapping algorithm from fingerprint to location, \ie $ \funSym{f}:\setSymScript{O}{i}{u}\mapSym \vecSymScript{\hat{l}}{i}{u}$. The \acs{rfm} $ \funSym{F} $ represents the relationship between the location $ \vecSym{l} $ and the fingerprint $ \setSym{O} $, \ie $ \funSym{F}:\vecSym{l}\mapSym\setSym{O} | \vecSym{l}\in\setSym{G} $ throughout the \acs{roi} $ \setSym{G} $. In reality, a discrete \acs{rfm} $ \setSym{F} := \{(\vecSym{l}_j, \setSymScript{\tilde{O}}{j}{})| \vecSym{l}_j\in \setSym{G}, j\in\intSet{|\setSym{F}|} \} $ (where $ \setSymScript{\tilde{O}}{j}{} = \funSym{F}(\vecSym{l}_j) $) is collected at different locations within the \acs{roi} $ \setSym{G}$.

\begin{figure}[!htb]
	\centering
	\includegraphics[width=0.85\columnwidth]{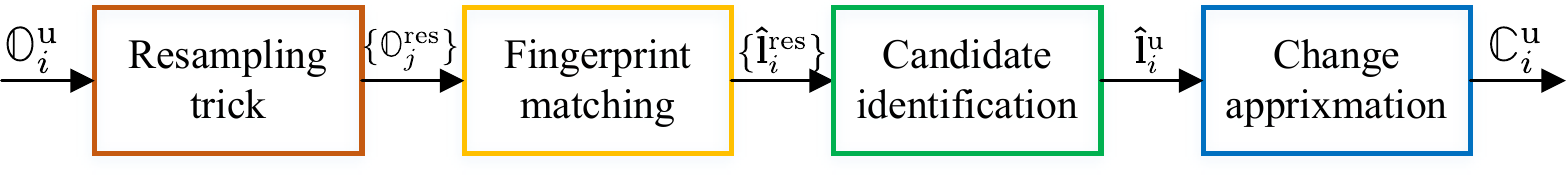}
	\caption{System flowchart of the proposed approach}
	\label{fig:system_ransac}
\end{figure}

Our joint robust positioning and feature-wise change detection works in the following procedures (\figPref\ref{fig:system_ransac}). Given a user measured fingerprint $ \setSymScript{O}{i}{u} $ at location/time $ i $, $ \norSymScript{N}{}{res} $ sampled observations $ \norSymScript{\{\setSymScript{O}{i}{res}\}}{i=1}{\norSymScript{N}{}{res}} $ are constructed by randomly (\eg uniformly) sampling \footnote{The analysis of varies sampling schemes is out-of-scope to this paper. A complete discussion about resampling can be found in \eg \citep{scott2015multivariate}.} a given ratio $ \norSymScript{\alpha}{}{res} $ (\ie $ \lceil\norSymScript{\alpha}{}{res}\cdot|\setSymScript{O}{i}{u}|\rceil$ signals) of those measured signals in $ \setSymScript{O}{i}{u} $. These resamples are used for robustly estimating the users' position $ \vecSymScript{\hat{l}}{i}{u} $ by matching them to the \acs{rfm} $ \setSym{F}$ using the selected positioning approach $ \funSym{f} $, intermediate locations $ \setSymScript{\hat{L}}{}{res}:= \{\vecSymScript{\hat{l}}{j}{res}\}_{j=1}^{\norSymScript{N}{}{res}} $ , where $ \vecSymScript{\hat{l}}{j}{res}=\funSym{f}(\setSymScript{O}{j}{res}) $. These temporally estimated locations $ \setSymScript{\hat{L}}{}{res} $ are treated as alternatives for refining the position estimation. We then retrieve the candidate locations from these alternatives by comparing the expected observations inferred from the world model of the \acs{rfm} and the one measured by the user. The retrieving scheme is designed in a way such that it takes the ones closest to the ground truth (though unknown during online positioning phase) as the candidates and they are used for computing $ \vecSymScript{\hat{l}}{i}{u} $. Once $ \vecSymScript{\hat{l}}{i}{u} $ is estimated, the change status of the measurable features are flagged through the change belief approximation.

\section{Joint robust positioning and feature-wise change detection}\label{sec:robust_fbp}
Our scheme consists of four steps (\figPref\ref{fig:system_ransac}): resampling, fingerprint matching, candidate identifying and change approximating and they are discussed in the following sections. For the resampling and fingerprint matching, we provide further remarks on them for clarifying our contributions and the differences comparing to the typical way of carrying out these two steps. Regarding the candidate position identification and change belief approximation, we formally formulate them out mathematically for mitigating the background gap and for backing our arguments.

\subsection{Resampling trick}
Comparing to the resampling tricks used in indoor positioning for assessing the positioning uncertainty and for comparing positioning algorithms \citep{4782964}, among which the sampling is carried out measurement-wise over the \acs{rfm}, our resampling trick is carried in the feature space of the online measurement instead of performing resampling measurement-wisely over the \acs{rfm}. The reason why we carry out the resampling in the feature space is two-fold: i) it makes full use of the redundancy of opportunistically observable location-relevant signals; and ii) it forms examples consisting of merely or predominantly the stable features of the online measurement. This supports that using the subset of the measured features can reach the sufficient position accuracy in case that the sampled observation consists of the stable features. In addition, using a subset of the measured features contributes to the dispersiveness (\ie variance) of the estimated positions. To a certain level of variance it is benefit for achieving the robust positioning. Because the variance makes the positioning process possible to locate close to the ground truth location (though unknown at the online positioning phase). The dispersiveness of the locations estimated using these resamples are influenced by the sampling ratio $ \norSymScript{\alpha}{}{res} $ (see examples in \figPref\ref{fig:cmp_distribution}). We present the quantitative metric for describing the dispersiveness and present our way of identifying the proper sampling ratio in the next subsection.

\subsection{Fingerprint matching in the subspace}\label{subsec:fin_match}
Fingerprint matching is the key to estimating the position, which can be accomplished by varies machine learning algorithms (\eg \acs{knn}, SVR, decision trees and random forest, neural networks) \citep{Padmanabhan2000,8115881,he2016wi}. We consider the fingerprint matching from the perspective of how to handle the case that the variant number of measurable features are sampled in the resampling step. We regard this problem in two cases: i) vector-based or ii) non-vector-based fingerprint matching.

The former case, the widely used one, requires to handle the variant measurable features problem specifically. Because machine learning algorithms used for positioning take vectorized features (\ie the measured features are represented as a vector) as input. In \citep{7676155,7565565}, He \etal~applied similar random sampling scheme in the feature space for detecting alterations, however, they carried out the fingerprint matching by filling in a missing value indicator to those unsampled features. This way can solve the varying dimension problem after sampling, but it introduces the extra bias to the dissimilarity measure and reduces the informativeness and discriminativeness of the features \citep{Murphy:2012:MLP:2380985,battaglia2018relational}. Our vector-based fingerprint matching for dealing with unsampled features is to omit those features from the \acs{rfm}. I.e., we drop out the corresponding unsampled features from the \acs{rfm}. This dropout idea, broadly used in training over-parameterized model (\eg deep neural networks) \citep{gal2016dropout}, is motivated by the essential idea of change detection because those unsampled features are assumed as the features in which changes have occurred. Therefore, we should not take them into account when matching the fingerprint with the \acs{rfm}.

\begin{figure}[!t]
	\centering
	\subfloat[\acs{knn}]{
		\label{subfig:knn_ecdf_missing}
		\includegraphics[width=0.4\linewidth]{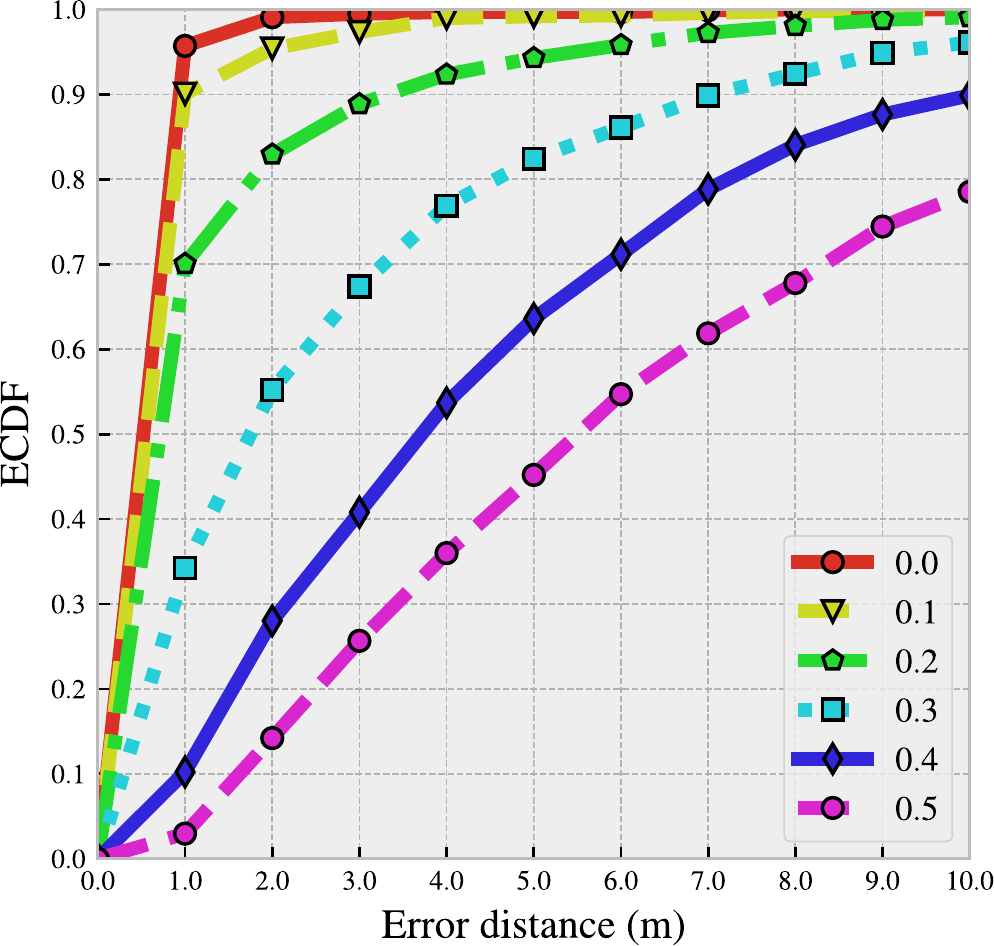}}\hspace{3ex}
	\subfloat[\acs{knn} with \acs{cdm}]{
		\label{subfig:cdm_ecdf_missing}
		\includegraphics[width=0.4\linewidth]{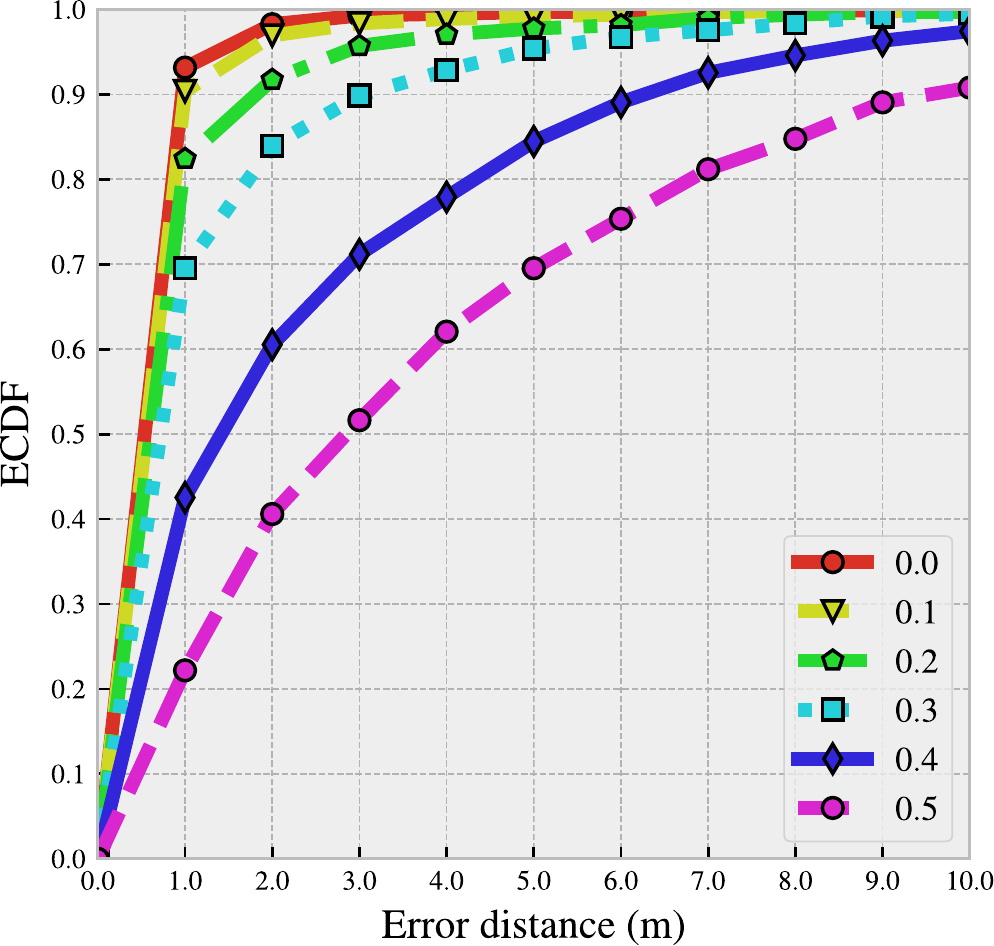}}
	\caption{Empirical cumulative distribution function (\acs{ecdf}) \acf{wrt} the positioning error on dataset \textit{HIL-R1}. We simulate sampling processing in the feature space by arbitrarily forcing a certain percentage of measurable features as missed from 10\% to 50\%. I.e. setting their measured value to be a missing indicator. More details about the experimental configuration can be found in \secPref\ref{subsec:config_exp}.}
	\label{fig:cmp_ecdf_missing}
\end{figure}

For the latter case of non-vector-based fingerprint matching, it has been firstly discussed in \citep{Zhou2018ipin} and is named \acf{cdm}, which requires no specific treatment for dealing with the varying number of measurable features. We take \acs{cdm} as a variant to the typical \acl{fbp} method (\eg \acs{knn}). In \figPref\ref{fig:cmp_ecdf_missing}, we compare the positioning performance of \acs{knn} and \acs{knn} with \acs{cdm}when a subset of measurable features is missing. Though both of them are negatively affected by the missing features, the degrading of positioning performance of \acs{knn} with \acs{cdm} is much less than that of \acs{knn}. This missing feature situation is equivalent of sampling a subset of features in the previous step. In order to avoid over-complicating the experimental analysis, we select \acs{knn} with \acs{cdm} as the representative positioning approach.

\begin{figure}[!h]
	\centering
	\subfloat[5\%]{
		\label{subfig:distribution_pt_5}
		\includegraphics[width=0.3\linewidth]{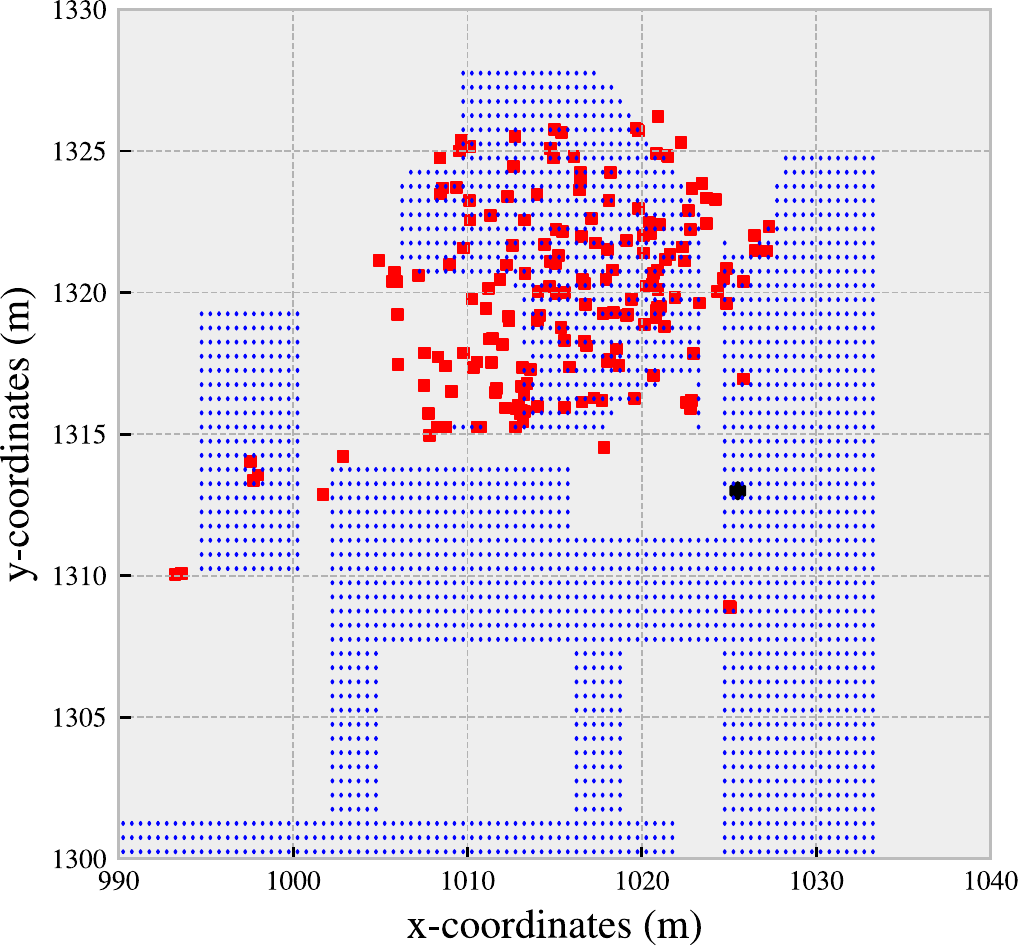}}\hspace{2ex}
	\subfloat[35\%]{
		\label{subfig:distribution_pt_35}
		\includegraphics[width=0.3\linewidth]{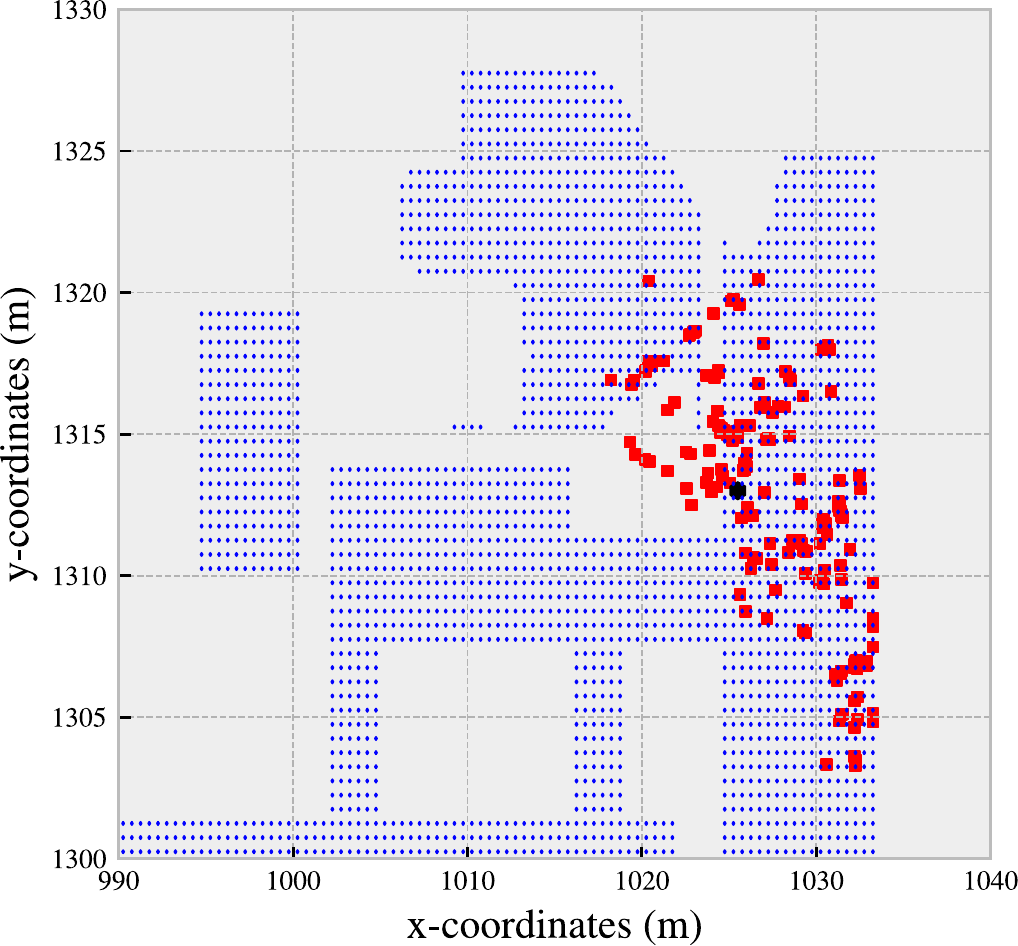}}\hspace{2ex}
	\subfloat[65\%]{
		\label{subfig:distribution_pt_65}
		\includegraphics[width=0.3\linewidth]{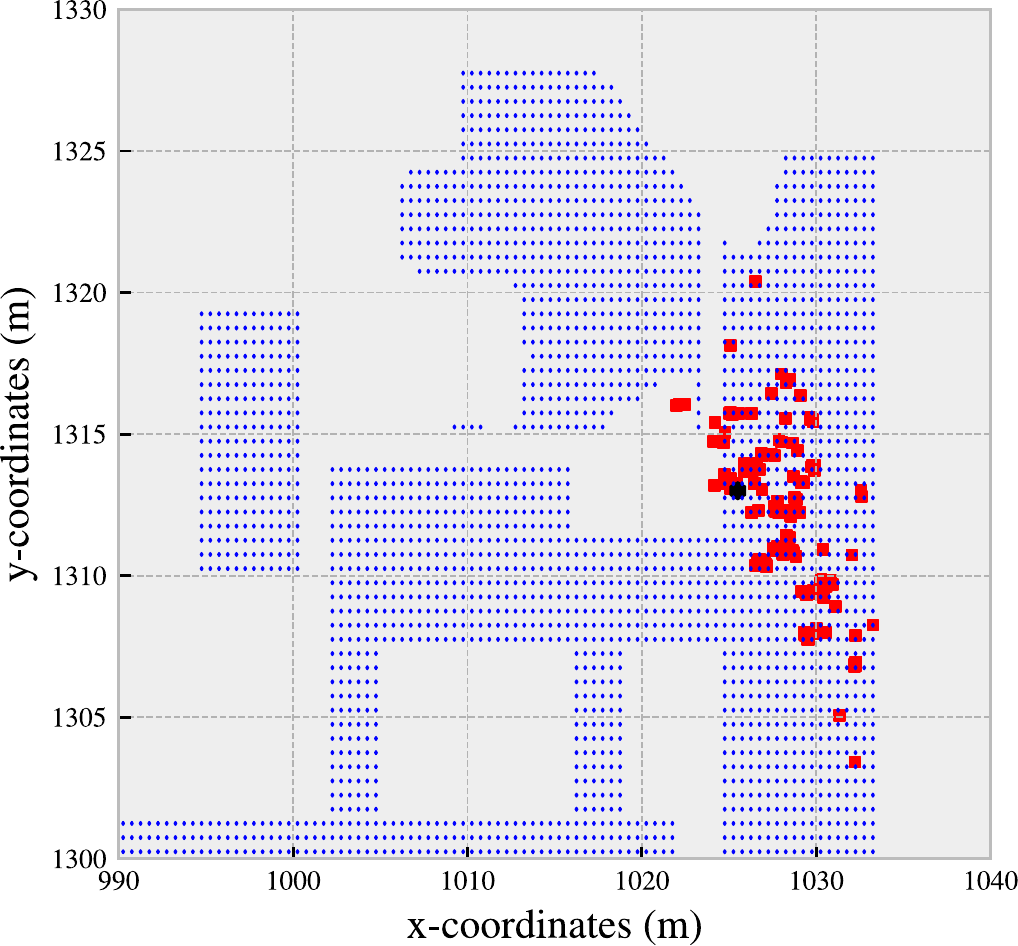}}
	\caption{An example of the dispersiveness of the locations estimated by \acs{knn} with \acs{cdm} using the sampled observations. The sampling ratio is 5\%, 35\% and 65\%,  respectively. The tiny blue dots denote the reference points of the interpolated \acs{rfm}. The black hexagon denotes the ground truth location. The red squares denote the estimated locations.} 
	\label{fig:cmp_distribution}
\end{figure}

The output of the fingerprint matching is the estimated location of the given positioning approach, \ie $ \vecSymScript{\hat{l}}{j}{res} =\funSym{f}(\setSymScript{O}{j}{res}), j\in\intSet{\norSymScript{N}{}{res}} $. The distribution of $ \norSymScript{N}{}{res} $ estimated locations of a given measurement $ \setSymScript{O}{i}{u} $ varies \acs{wrt} the different sampling ratio $ \norSymScript{\alpha}{}{res} $ as shown in \figPref\ref{fig:cmp_distribution}. In order to find an optimal range of the sampling ratio, we propose two metrics for quantifying and balancing the dispersiveness and the bias of the estimated locations.
\begin{itemize}
	\item \textbf{Dispersiveness}: defined as the area of the elliptic covering the estimated locations. It is calculated by the eigen-decomposition of the empirical variance-covariance matrix of those estimated locations.
	\item \textbf{Bias}: defined as the minimum value among the error distances between these estimated locations and the ground truth location.
\end{itemize}

The intuition of introducing the above two metrics is motivated by the definiteness of the fingerprinting-based positioning and the variability of the discriminativeness of the sampled observation. As increasing percentage of features is sampled for positioning, the estimated locations assemble together. I.e. the dispersiveness reduces as the increasing of the sampling ratio. However, the bias depends on the discriminativeness of the sampled features. In case that too few relevant features or too many bad features (\ie the ones that have been changed) are sampled, it likely yields large bias. We thus suppose that the curve of bias \acs{wrt} the sampling ratio is close to the U-shape. By performing the experimental analysis, we could discover a suitable range of the sampling ratio that balances the dispersiveness and bias (see \secPref\ref{subsec:disp_bias_ana}). 

\subsection{Candidate position identification}
The goal of this step is to retrieve the locations that closest to the ground truth (though unknown) from those $ \norSymScript{N}{}{res} $ locations estimated using the sampled observations. We thus can recover good positioning accuracy in case that partial of the measured features has changed. The first trial is to count the number features whose residual $ \norSymScript{\delta}{ij}{} $ is smaller than a given threshold $ \norSymScript{\lambda}{}{res} $ by comparing the user's measurement $ \setSymScript{O}{i}{u} $ with the expected measurement $\setSymScript{\tilde{\hat{O}}}{j}{res}  $ (where $\setSymScript{\tilde{\hat{O}}}{j}{res} := \funSym{F}(\vecSymScript{\hat{l}}{j}{res})  $) at location $ \vecSymScript{\hat{l}}{j}{res}  $ . $ \norSymScript{\delta}{ij}{} $ is represented as vector containing $ |\setSymScript{O}{i}{u}| $ values and is calculated:
\begin{equation}
	\label{eq:residual}
	 \norSymScript{\delta}{ij}{} = [|\norSymScript{v}{i1}{u} - \norSymScript{\tilde{\hat{v}}}{j1}{res}|, |\norSymScript{v}{i2}{u} - \norSymScript{\tilde{\hat{v}}}{j2}{res}|, \cdots, |\norSymScript{v}{i|\setSymScript{O}{i}{u}| }{u} - \norSymScript{\tilde{\hat{v}}}{j|\setSymScript{O}{i}{u}| }{res}|]
\end{equation}
where $ (\norSymScript{a}{ik}{u}, \norSymScript{v}{ik}{u})\in \setSymScript{O}{i}{u}$ and $ (\norSymScript{\tilde{\hat{a}}}{jk}{res}, \norSymScript{\tilde{\hat{v}}}{jk}{res}) \in \setSymScript{\tilde{\hat{O}}}{j}{res} (k\in\intSet{|\setSymScript{O}{i}{u}|})$. Herein we assume that the world model of the \acs{rfm}  is continuously and smoothly representable. I.e., the expect values of all measurable features contained in $ \setSymScript{O}{i}{u} $ at the estimated locations can be obtained. From the computing procedure, we can qualitatively identify three main sources of the residuals: value changes of the measured features, deviations caused by the difference between the estimated locations and the ground truth, and the modeling error of the \acs{rfm}. We assume that the value changes of the feature dominate among the residuals.

The indicating value $ \funSym{S}^{\mathrm{res}}: \convSetSym{R}^{|\setSymScript{O}{i}{u}|}\mapSym\convSetSym{R} $ is then defined as the number of features whose residuals are no larger than the given threshold $ \norSymScript{\lambda}{}{res} $:
\begin{equation}
	\label{eq:support_set}
	\funSym{S}^{\mathrm{res}}(\norSymScript{\delta}{ij}{}) = \sum_{k=1}^{|\setSymScript{O}{i}{u}|}\funSym{I}(|\norSymScript{v}{ik}{u} - \norSymScript{\tilde{\hat{v}}}{jk}{res}|\le\norSymScript{\lambda}{}{res})
\end{equation}
where $ \funSym{I}(\cdot) $ yields 1 if the condition is satisfied otherwise yields 0. Among $ \norSymScript{N}{}{res} $ sampled observations of $ \setSymScript{O}{i}{u} $, the one achieves the maximum indicating value is identified as the candidate position $ \vecSymScript{\hat{l}}{i}{u} $.

\begin{figure}[!h]
	\centering
	\subfloat[35\%, $ \norSymScript{\lambda}{}{res} = 5 \text{dBm}$]{
		\label{subfig:retrieving_35_5}
		\includegraphics[width=0.3\linewidth]{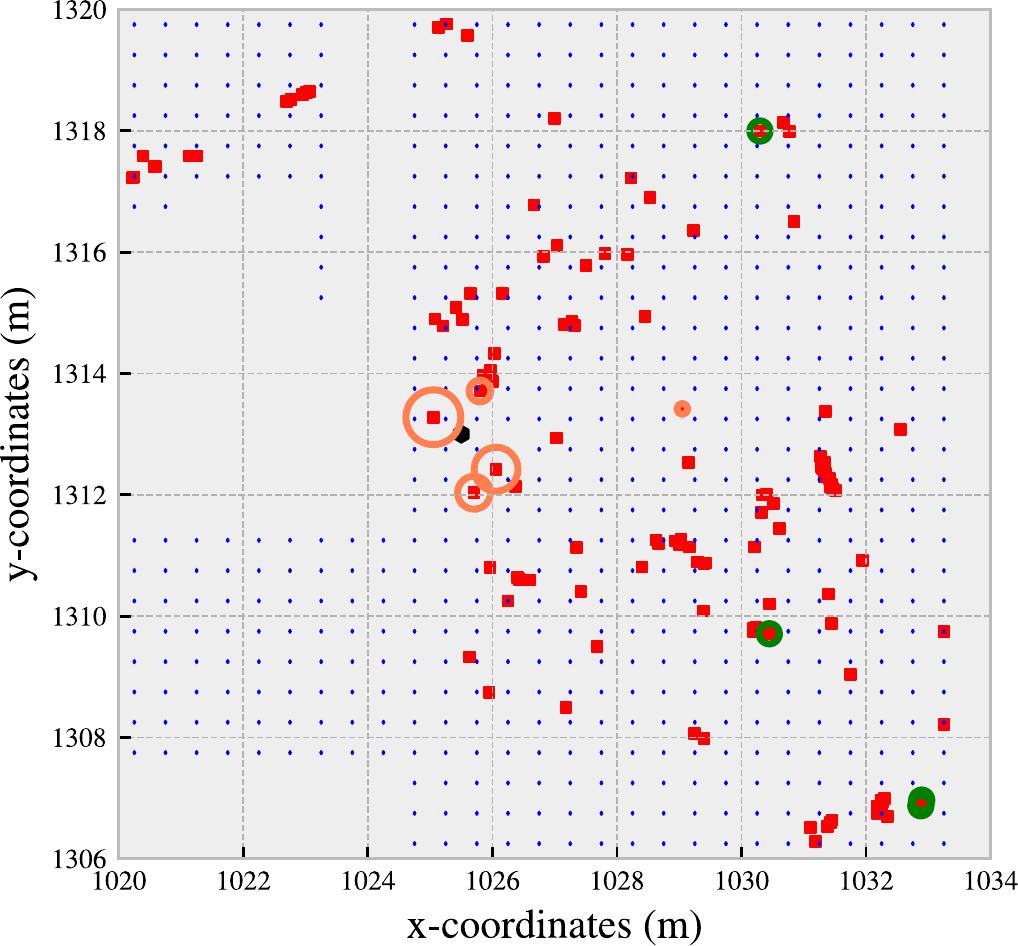}}\hspace{2ex}
	\subfloat[35\%, $ \norSymScript{\lambda}{}{res}  = 10 \text{dBm}$]{
		\label{subfig:retrieving_35_10}
		\includegraphics[width=0.3\linewidth]{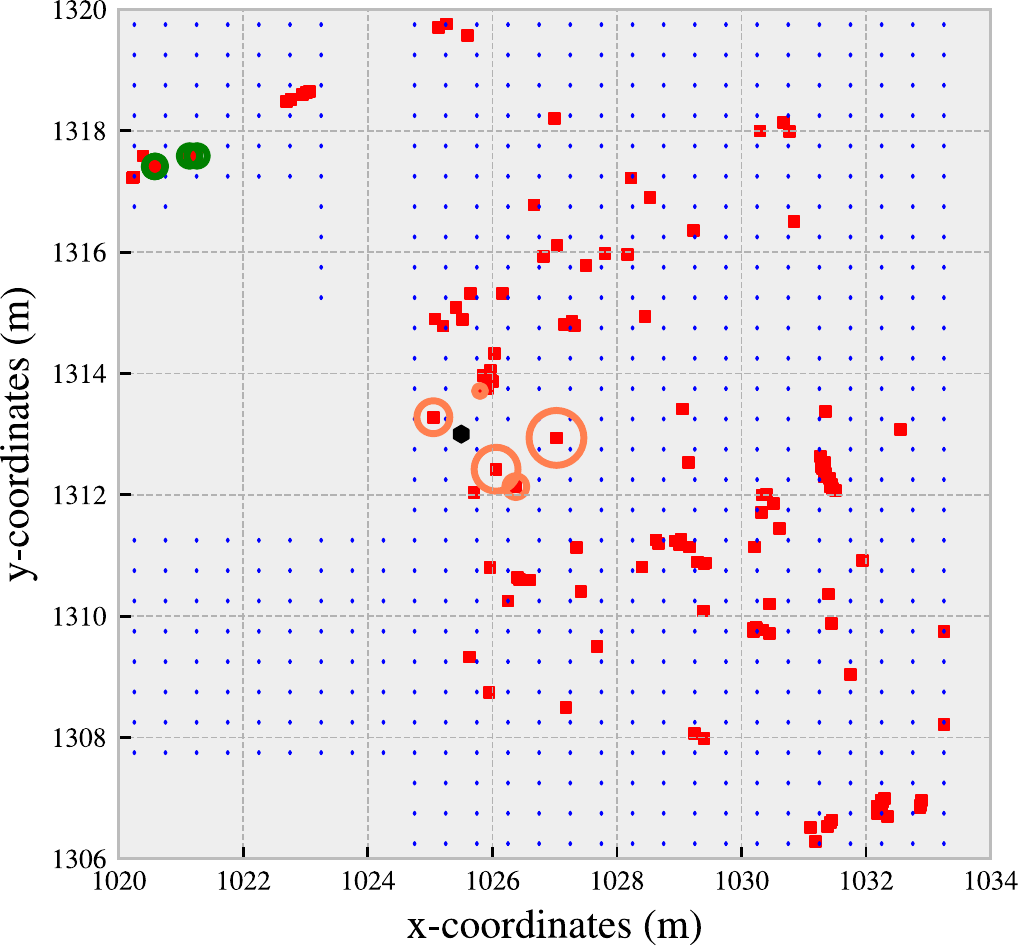}}\hspace{2ex}
	\subfloat[35\%, $ \norSymScript{\lambda}{}{res}  = 15 \text{dBm}$]{
		\label{subfig:retrieving_35_15}
		\includegraphics[width=0.3\linewidth]{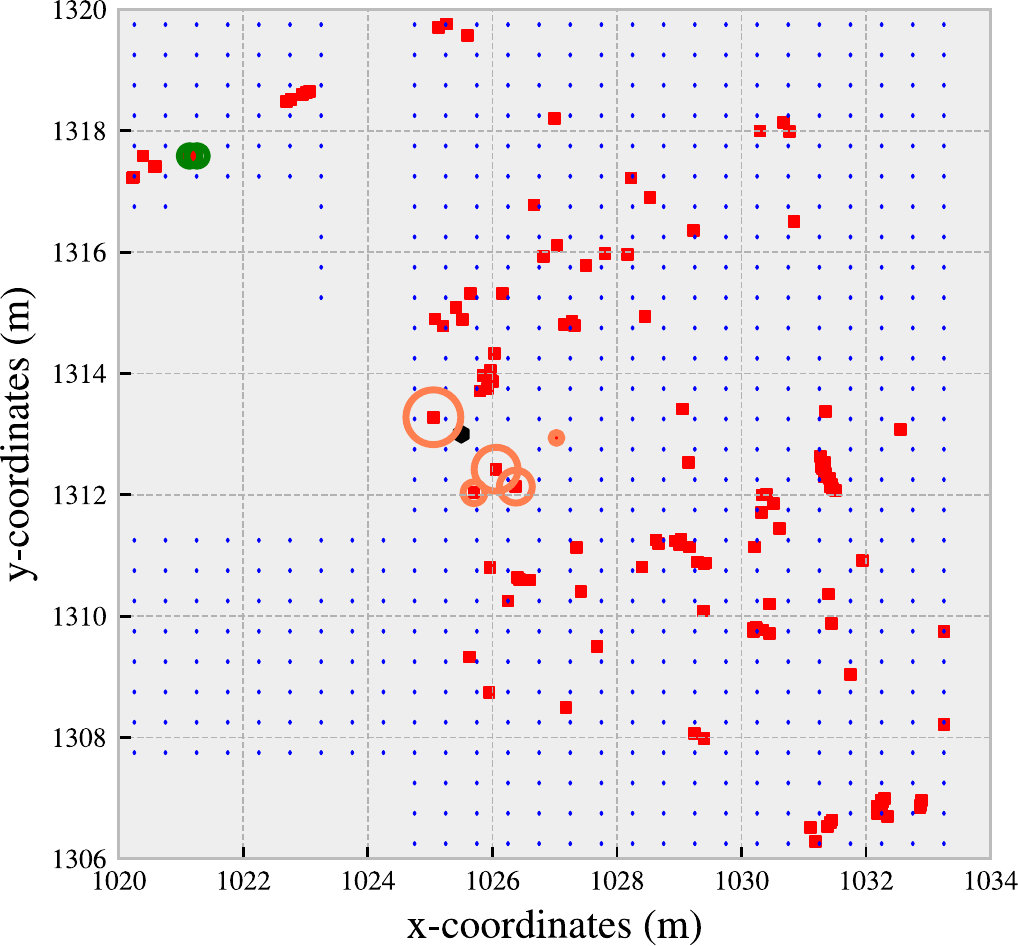}}\\
	\subfloat[65\%, $ \norSymScript{\lambda}{}{res}  = 5 \text{dBm}$]{
		\label{subfig:retrieving_65_5}
		\includegraphics[width=0.3\linewidth]{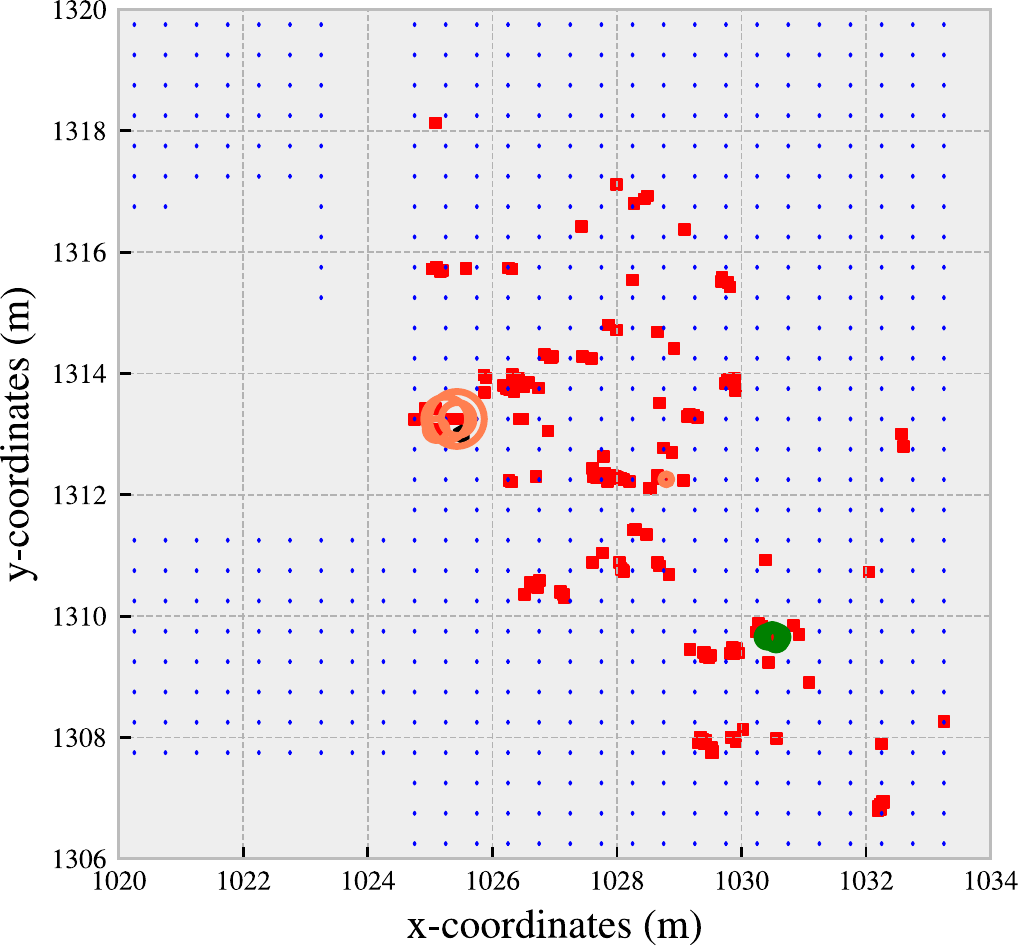}}\hspace{2ex}
	\subfloat[65\%, $ \norSymScript{\lambda}{}{res}  = 10 \text{dBm}$]{
		\label{subfig:retrieving_65_10}
		\includegraphics[width=0.3\linewidth]{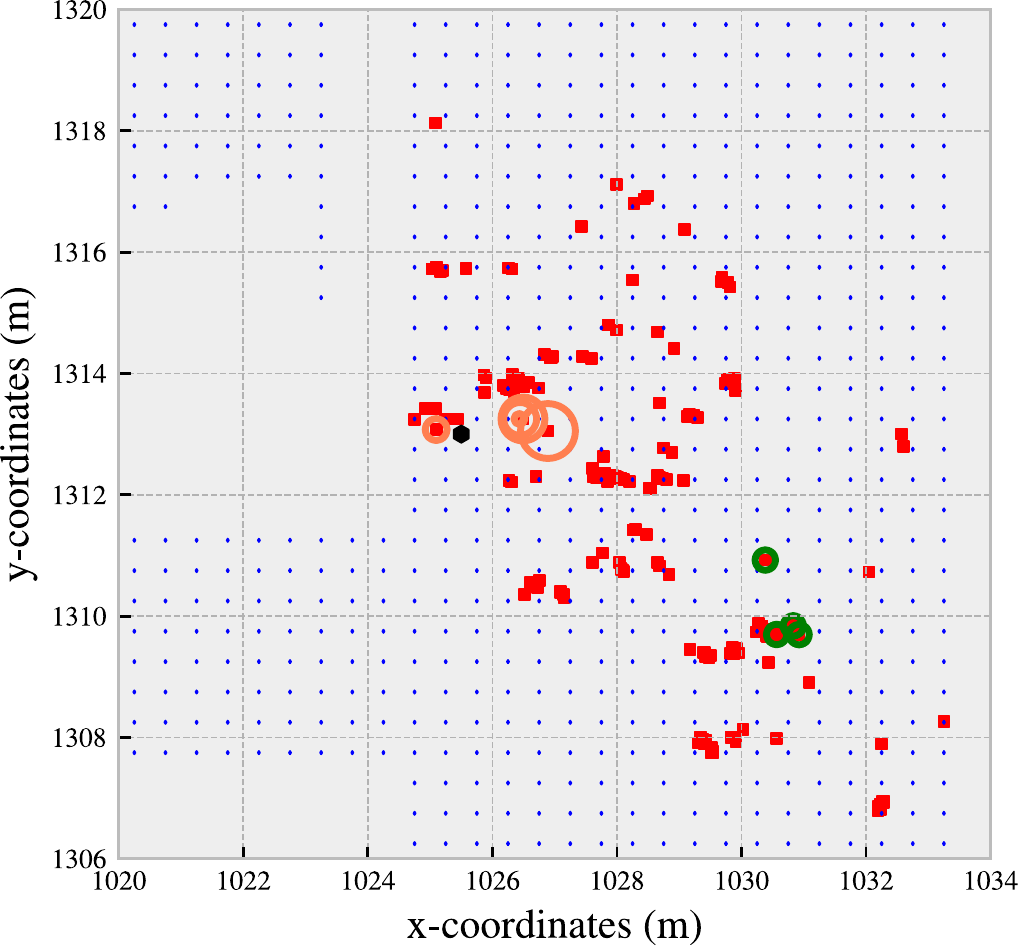}}\hspace{2ex}
	\subfloat[65\%, $ \norSymScript{\lambda}{}{res}  = 15 \text{dBm}$]{
		\label{subfig:retrieving_65_15}
		\includegraphics[width=0.3\linewidth]{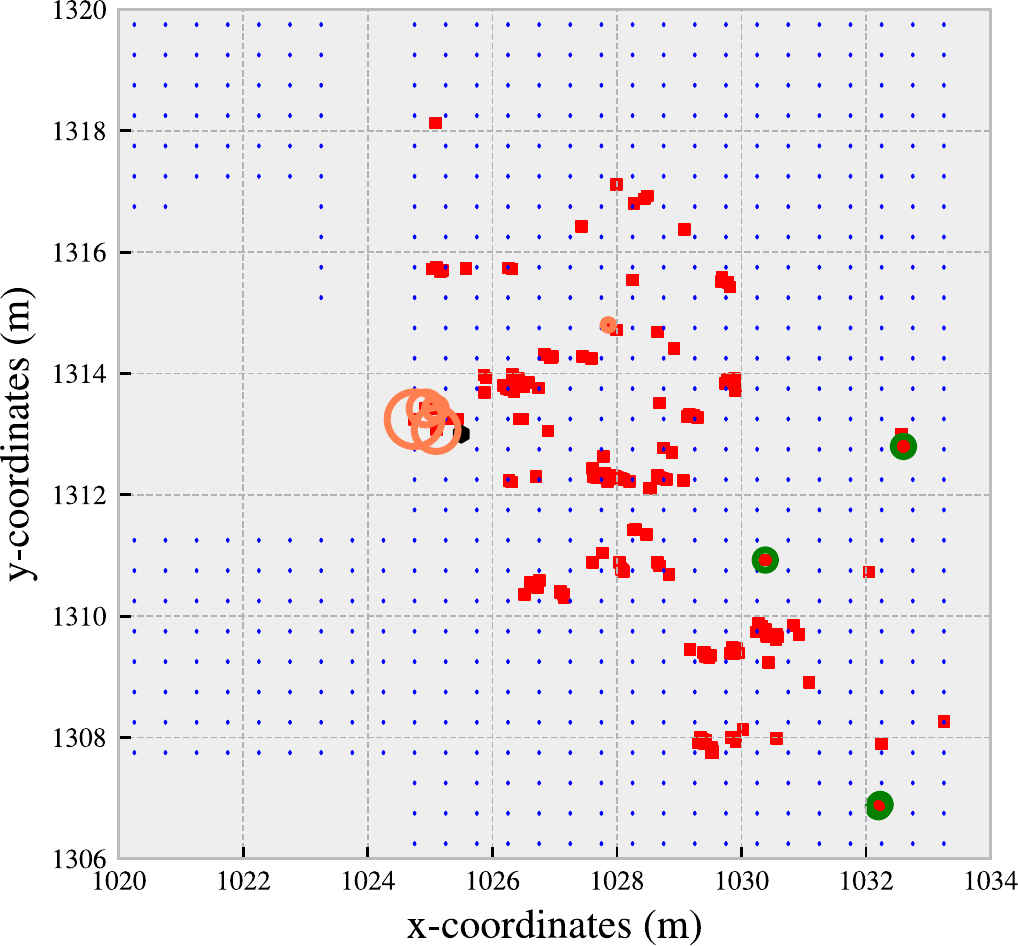}}\\
	\caption{An example of candidate identification result using variant values of threshold. The tiny blue dots denote the reference points of the interpolated \acs{rfm}. The black hexagon denotes the ground truth location. The red squares denote the estimated locations. The 5 locations whose indicating values rank at top 5 are denotes by a unfilled coral circle (the radius is proportional to the indicating value).  The 5 locations whose indicating values rank at bottom 5 are denotes by a unfilled green circle.}
	\label{fig:cmp_retrieving}
\end{figure}

\figPref\ref{fig:cmp_retrieving} gives the example of impact of threshold value $ \norSymScript{\lambda}{}{res} $ on regaining the candidate location. From this initial result, we find out that the candidate position identification is highly sensitive to the threshold though the retrieved candidate locations are quit close to each other in the given example. In this situation, it is difficult to find a threshold that is suitable for different observations with varies changes and sampling ratio. One extra information that unused in the thresholding way of identifying the candidate location is similarity/dissimilarity measure between $ \setSymScript{O}{i}{u} $ and $\setSymScript{\tilde{\hat{O}}}{j}{res}  $. Because the location is estimated using a subset of the measured features, the similarity between  two full measurements, \ie $ \setSymScript{O}{i}{u} $ and $\setSymScript{\tilde{\hat{O}}}{j}{res}  $, should be high if the estimated location is closed to the ground truth location. We thus use a threshold-free way for candidate position identification according the value of \acf{mji} \citep{Zhou2018}. It is capable of measuring the similarity according to measurability of the features. The \acs{mji} value $ \funSym{S}^{\mathrm{MJI}}: \setSym{O}\times\setSym{O}\mapSym\convSetSym{R} $:
\begin{equation}
	\label{eq:mji}
	\funSym{S}^{\mathrm{MJI}}(\setSymScript{O}{i}{u}, \setSymScript{\tilde{\hat{O}}}{j}{res}) = \frac{1}{2}\Big(\frac{|\setSymScript{A}{i}{u}\cap \setSymScript{\tilde{\hat{A}}}{j}{res}|}{|\setSymScript{A}{i}{u}\cup \setSymScript{\tilde{\hat{A}}}{j}{res}|} + \frac{|\setSymScript{A}{i}{u}\cap \setSymScript{\tilde{\hat{A}}}{j}{res}|}{|\setSymScript{A}{i}{u}|}\Big)
\end{equation}
where $ \setSymScript{A}{i}{u} $ and $ \setSymScript{\tilde{\hat{A}}}{j}{res} $ are the set of measured features of $  \setSymScript{O}{i}{u} $ and $ \setSymScript{\tilde{\hat{O}}}{j}{res} $, respectively. In short, we denote $ \funSym{S}^{\mathrm{MJI}}(\setSymScript{O}{i}{u}, \setSymScript{\tilde{\hat{O}}}{j}{res}) $ as $ \funSym{S}^{\mathrm{MJI}}_{ij} $.

\begin{figure}[!h]
	\centering
	\subfloat[\acs{mji}]{\label{subfig_mji_35}
		\includegraphics[width=0.3\linewidth]{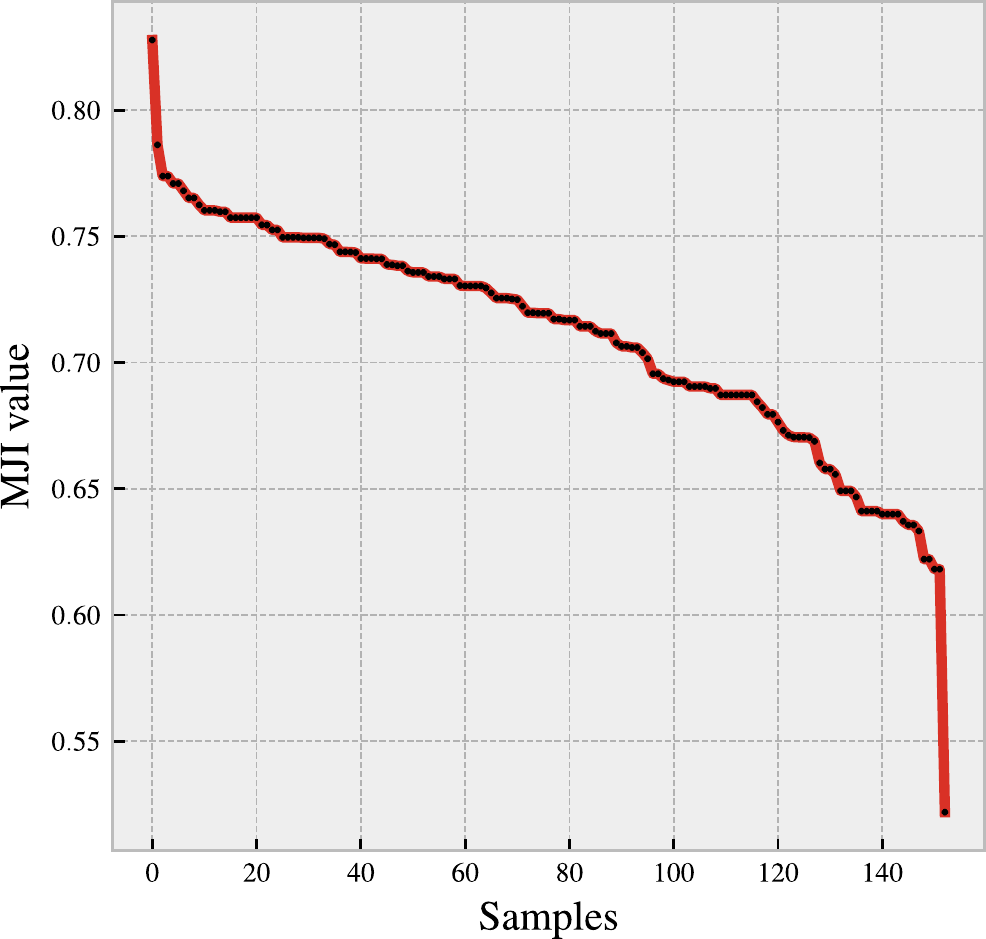}}\hspace{1ex}
	\subfloat[Top 5]{\label{subfig_mji_retrieve_5_35}
		\includegraphics[width=0.3\linewidth]{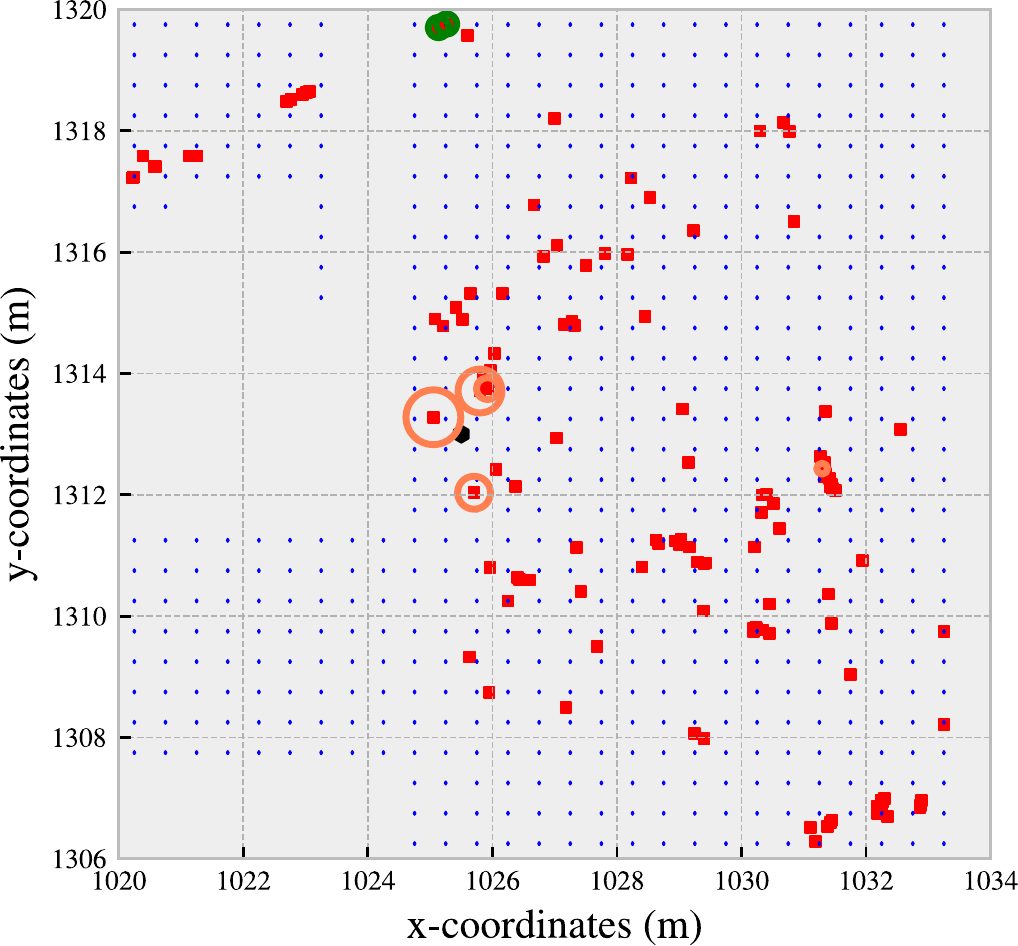}}\hspace{1ex}
	\subfloat[Adaptive (Top 1)]{\label{subfig_mji_retrieve_1_35}
		\includegraphics[width=0.3\linewidth]{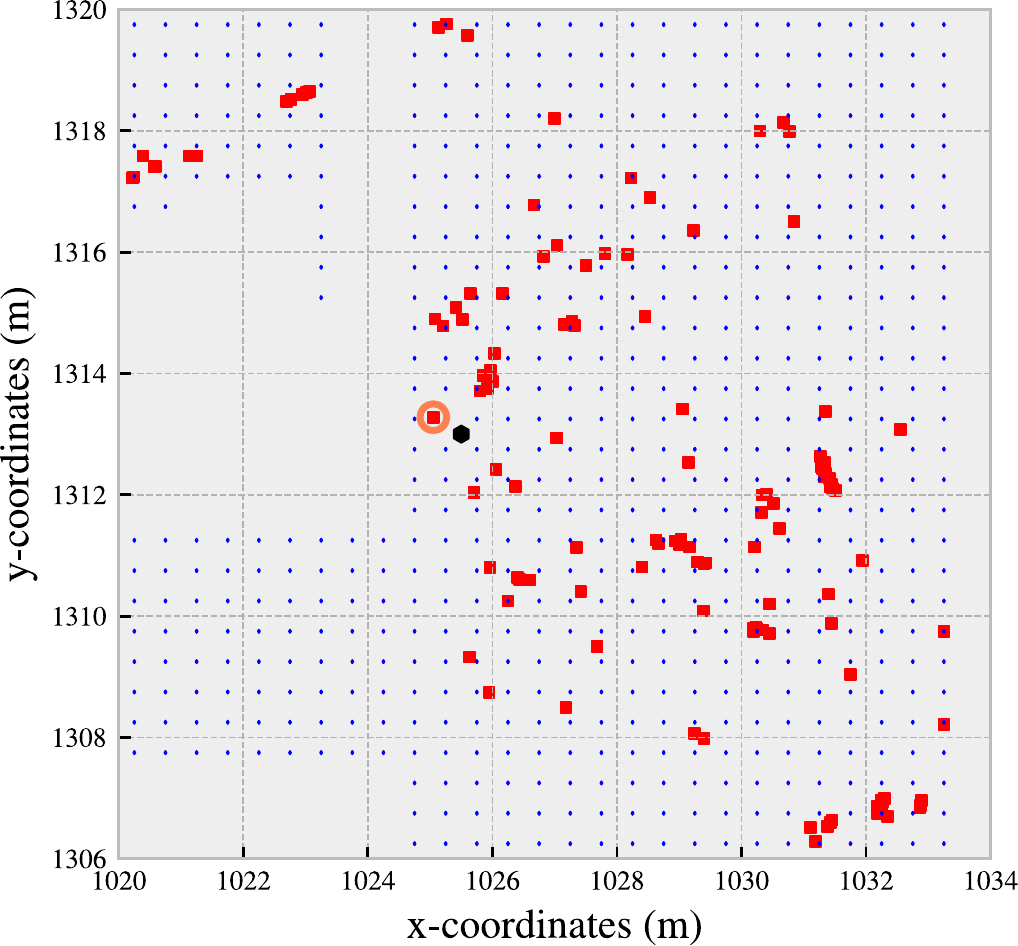}}\\
	\subfloat[\acs{mji}]{\label{subfig_mji_65}
		\includegraphics[width=0.3\linewidth]{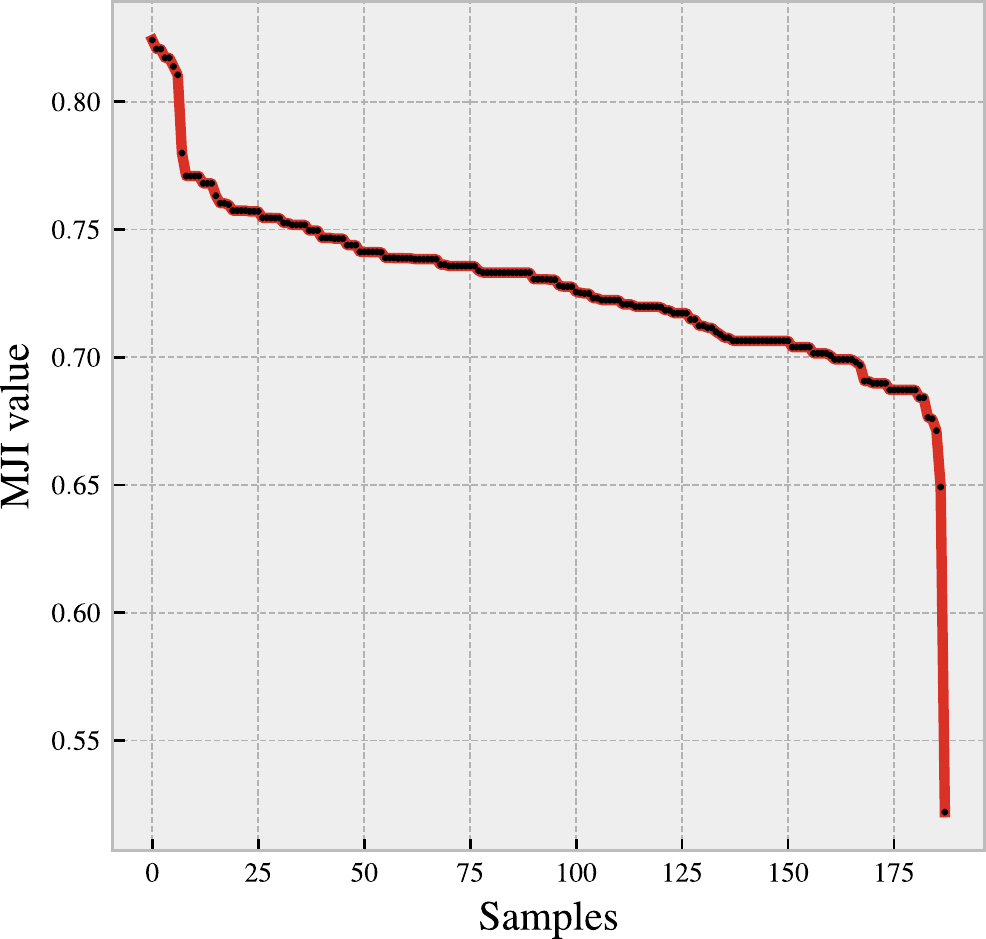}}\hspace{1ex}
	\subfloat[Top 5]{\label{subfig_mji_retrieve_5_65}
		\includegraphics[width=0.3\linewidth]{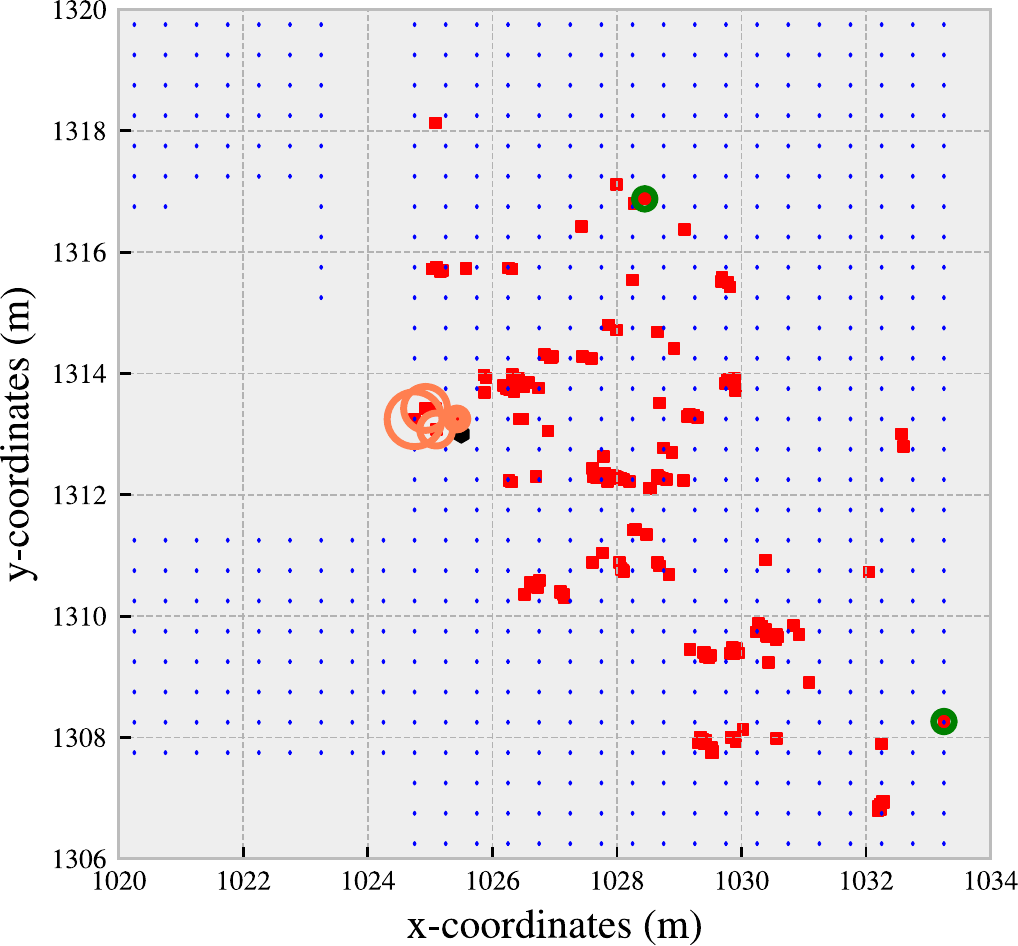}}\hspace{1ex}
	\subfloat[Adaptive (Top 7)]{\label{subfig_mji_retrieve_7_65}
		\includegraphics[width=0.3\linewidth]{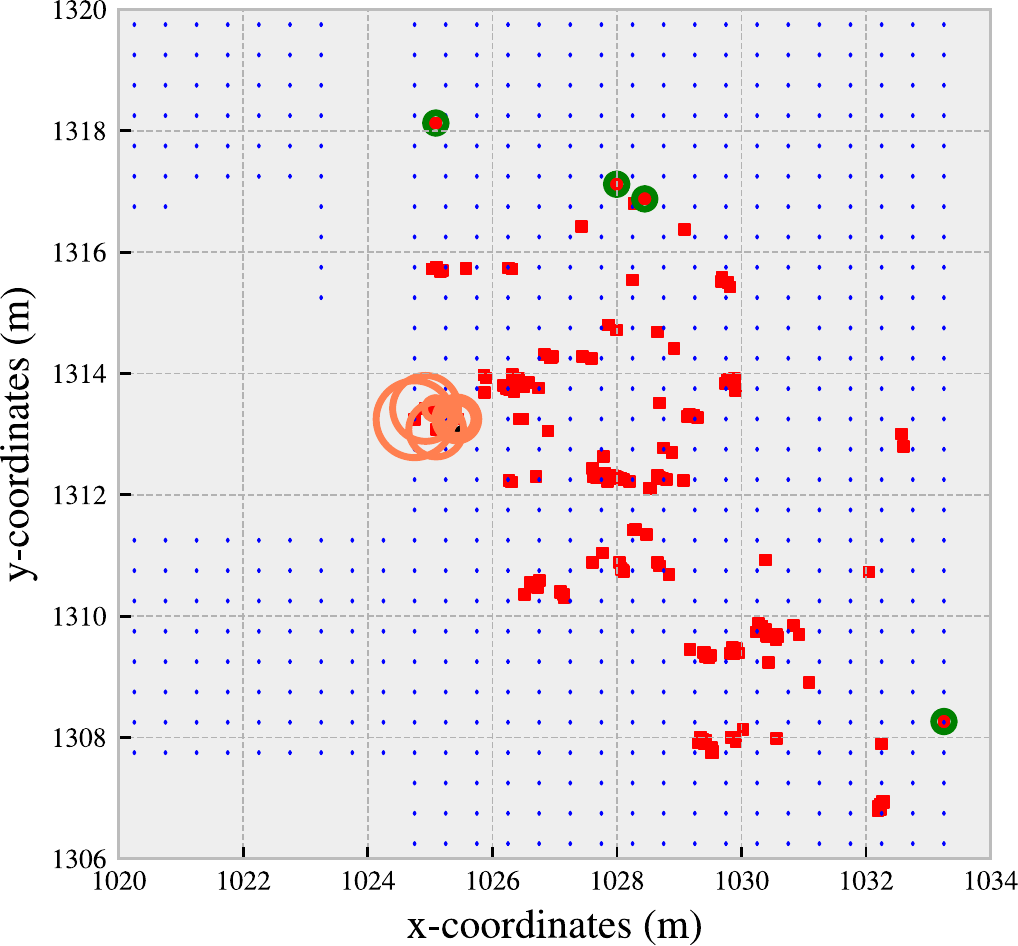}}
	\caption{An example of \acs{mji}-based candidate position identification in case that sampling ratio is 35\% (1st row) and 65\% (2nd row). For the adaptive selection of the candidate position, the value of decremental factor $ \norSymScript{\lambda}{}{MJI} $ equals to 0.97.}
	\label{fig:cmp_mji_retrieving}
\end{figure}

We compute the \acs{mji} value of each pair of them and select the candidate position corresponding to the one whose \acs{mji} is high. Motivated by the distribution of the estimated locations as shown in \figPref\ref{fig:cmp_distribution} and \figPref\ref{fig:cmp_retrieving}, we adaptively select the number of candidate position according to the ranking of \acs{mji} values \acs{wrt} samples (see \figPref\ref{fig:cmp_mji_retrieving}) because there might have multi-candidates which locate close to the ground truth location. This adaptively number $ \norSymScript{k}{}{MJI} $ is determined by bounding the decrement of the \acs{mji}:
\begin{equation}
	\norSymScript{k}{}{MJI} = \sum_{j=1}^{\norSymScript{N}{}{res}}\funSym{I}(\funSym{S}^{\mathrm{MJI}}_{ij} \ge \norSymScript{\lambda}{}{MJI}\cdot\operatorname{max}(\funSym{S}^{\mathrm{MJI}}_{ij}))
\end{equation}
where $ \norSymScript{\lambda}{}{MJI} $ is the decremental factor that controls the selected number of candidate positions.  It should be close to one, \ie only the ones whose \acs{mji}value are very close  the maximum value are chosen as candidates. Among these $ \norSymScript{k}{}{MJI} $ selected candidate positions, we estimate the user's location $ \vecSymScript{\hat{l}}{i}{u} $ by weighting them according to their \acs{mji} values.

\subsection{Change belief approximation}
\begin{figure}[!h]
	\centering
	\subfloat[ \acs{ap} 1, sampling ratio 65\%]{\label{subfig:distr_vap_13}
		\includegraphics[width=0.35\linewidth]{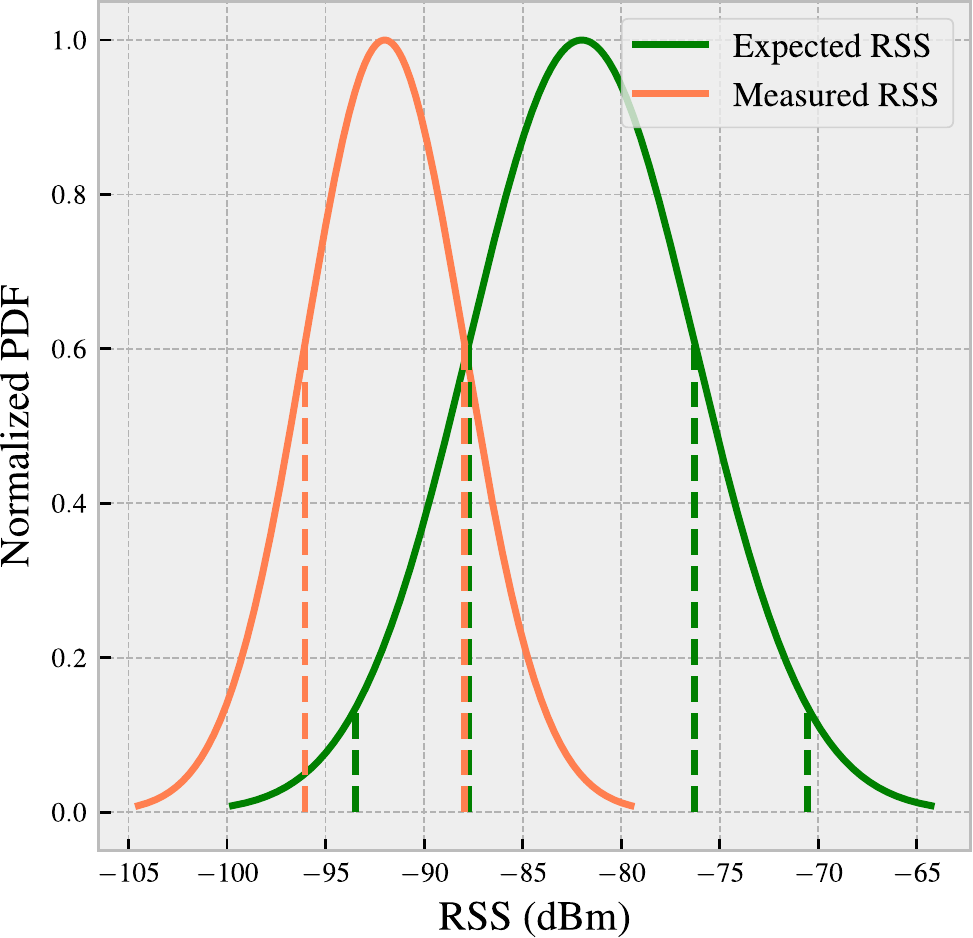}}\hspace{4ex}
	\subfloat[\acs{ap}  2, sampling ratio 25\%]{\label{subfig:distr_vap_17}
		\includegraphics[width=0.363\linewidth]{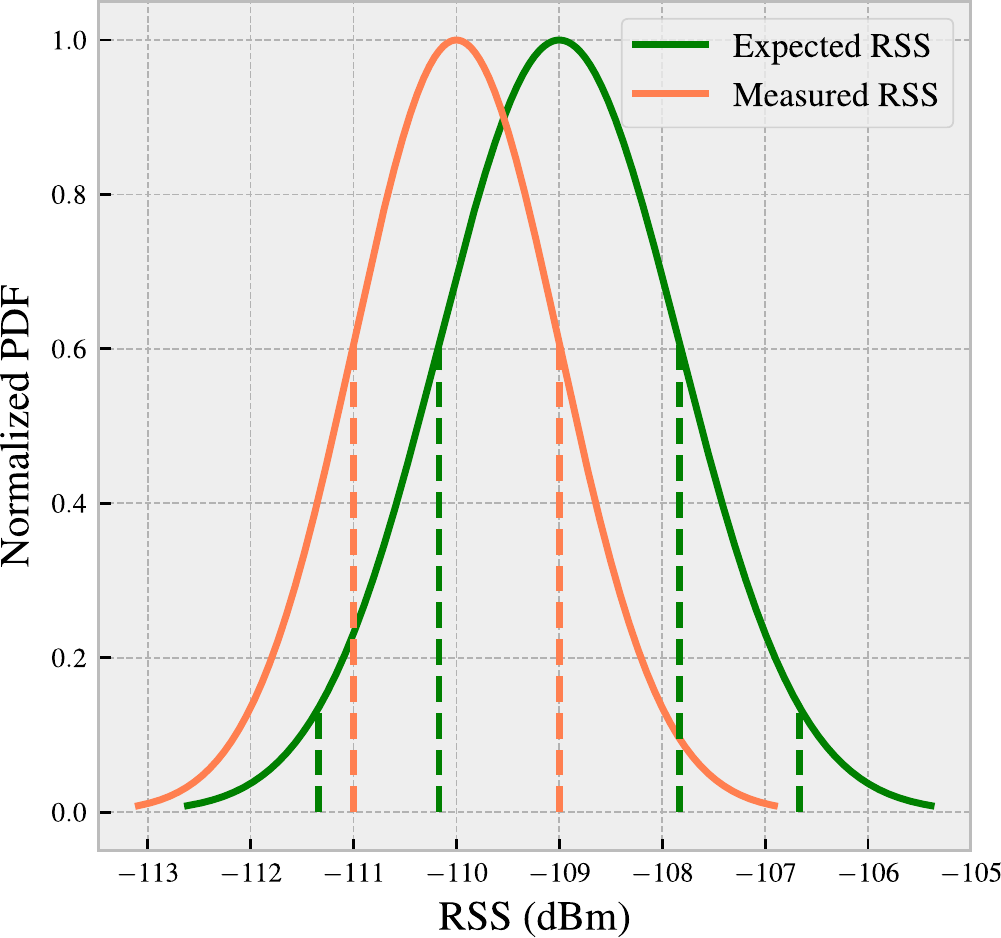}}
	\caption{Examples of the approximated Gaussian distributions of the measured and expected \acs{rss}. Both \acsp{ap} are unchanged and we can infer the change belief based on the approximated distributions. The dash lines are indicating the 1- or 2-$ \sigma $ confidence intervals. The MAC of two \acsp{ap} are (4c:fa:ca:fe:ec:82) and (4c:fa:ca:fe:ec:82), respectively.}
	\label{fig:distribution_examples}
\end{figure}

Once the users' position $ \vecSymScript{\hat{l}}{i}{u} $ is estimated, we flag whether one measured feature has changed according to the absolute residuals between the expected measurement $ \setSymScript{\tilde{\hat{O}}}{i}{u} := \funSym{F}(\vecSymScript{\hat{l}}{i}{u}) $ at $ \vecSymScript{\hat{l}}{i}{u} $ and the one measured by the user. We assume that the variability of the feature value is known or can be estimated (see \figPref\ref{fig:variability_ls}), \ie given a  measured value $ v $ of the feature $ a $, the \acf{std} $ \sigma $ of it equals to $ \funSym{g}(v) $. We use a normal distribution centered at a the measured/expected feature value and with the estimated \acs{std} for approximating the distribution of the feature value (see \figPref\ref{fig:distribution_examples}). Though a more sophisticated assumption of the distribution might be more suitable, especially when using the \acs{rss} as the feature (\eg Rayleigh distribution), it is difficult to decide which kind of distribution when few measurement is available.  For $ \norSymScript{k}{}{th} $ feature $ \norSymScript{a}{ik}{} \in \setSymScript{A}{i}{}$ (where $ \setSymScript{A}{i}{}=\setSymScript{A}{i}{u}\cup\setSymScript{\tilde{\hat{A}}}{i}{u} $) , its corresponding measured value $ \norSymScript{a}{ik}{u} $ and the expected one $ \norSymScript{\tilde{\hat{a}}}{ik}{u} $ are Gaussian distributed:
\begin{equation}
	\label{eq:norm_distr}
	\begin{split}
		&\norSymScript{v}{ik}{u}\sim \mathcal{N}(\norSymScript{v}{ik}{u}, \norSymScript{\sigma}{ik}{u})\\
		&\norSymScript{\tilde{\hat{v}}}{ik}{u} \sim \mathcal{N}(\norSymScript{\tilde{\hat{v}}}{ik}{u}, \norSymScript{\tilde{\hat{\sigma}}}{ik}{u})\\
		&\mbox{where } \norSymScript{\sigma}{ik}{u} = \funSym{g}(\norSymScript{v}{ik}{u}),\, \norSymScript{\tilde{\hat{\sigma}}}{ik}{u} = \funSym{g}(\norSymScript{\tilde{\hat{v}}}{ik}{u})
	\end{split}
\end{equation}

In case that we have multi-observations at one location, z-test is applicable for identifying the change status \citep{scott2015multivariate}. However, we are focusing on one observation-based change detection, we compute the intersection points $ v_1, v_2 $ between two approximated distributions by solving the quadratic equation. The change belief $ \norSymScript{\hat{c}}{ik}{u} $ of $ \norSymScript{k}{}{th} $ feature is computed:
\begin{equation}
	\label{eq:change_belief}
	\norSymScript{\hat{c}}{ik}{u} = 1 - {\operatorname{max}}\{\funSym{p}(v_1), \funSym{p}(v_2)\}
\end{equation}
where $ \funSym{p}(v) := \mathcal{N}(v|\norSymScript{\tilde{\hat{v}}}{ik}{u}, \norSymScript{\tilde{\hat{\sigma}}}{ik}{u})$. The change detection result of $ \setSymScript{O}{i}{u} $ is collected in $ \setSymScript{\hat{C}}{i}{u}:=\{(\norSymScript{a}{ik}{u}, \norSymScript{\hat{c}}{ik}{u})| \norSymScript{a}{ik}{u}\in \setSymScript{A}{i}{u}\} $. These detected statuses of feature can be employed in two ways: i) cumulated in the background for \acs{rfm} update, and ii) actively used online. The former is for achieving the \acs{rfm} update based on the collected feature change information from mutli-users/devices over a period. Features frequently reported as changed over the period would be updated with newly measured values according to the world model of the \acs{rfm} or perform a on-site data collection covering the part of \acs{roi} where the changed features has been measured. It thus can reduce the time and labor cost of keeping the \acs{rfm} up-to-date because the update of the \acs{rfm} is only carried out for the specific region and features. The latter helps to carry out the fingerprinting matching more specifically, \ie only using those unchanged features or removing the changed features for positioning during the later positioning process. Regarding the \acs{rfm} update procedure, it is an inverse process of positioning, \ie estimating the measurements at a given location with the combination of newly measured ones. Approaches used for \acs{rfm} update have been widely studied, specifically \eg Kriging \citep{Peng2018,jan2015received}, \acs{svr} \citep{8533825,10.1007/978-3-319-71470-7_1}, \acs{gpr} \citep{atia2013dynamic,Zou2017,7565565,8445663}, \acs{plsr} \citep{8003484} or neural network-based regression \citep{Sun2018}. In this paper, we focus on the change detection for controlling the quality of the adapted \acs{rfm} and accelerating the \acs{rfm} update process. 

\subsection{Summary of the parameters}\label{subsec:sumHyp}
We summarize its parameters of each step included in our proposed approach for achieving joint robust positioning and feature-wise change detection as follows:
\begin{itemize}
	\item Resampling trick: the number of samples $ \norSymScript{N}{}{res} $, and the sampling ratio $ \norSymScript{\alpha}{}{res} $;
	\item Fingerprint matching: the nested of parameters of \acl{fbp} methods (\eg nearest neighbors of \acs{knn} or probability functions of \acs{map}), and nested parameters of the world model of the \acs{rfm};
	\item Candidate position identification:  the decremental factor $ \norSymScript{\lambda}{}{MJI} $ for adaptively determining the number of candidate positions.
\end{itemize}

We experimentally and empirically determine the value of parameters for the analysis. The work for searching the optimal values (\eg using grid search \citep{Murphy:2012:MLP:2380985}, or \acl{bo} \citep{Snoek:2012}) of the parameters is left for future.

\section{Experimental results and discussion}\label{sec:experimentalResults}
In this section, we first describe the settings of the experiments including the algorithms for modeling the \acs{rfm} and for estimating the position, the datasets for validation, and metrics for evaluating the performance. Primary results regarding the influences of the sampling ratio on distribution the estimated locations are discussed in the following subsection. We then present the results of change detection and positioning on the simulated dataset. In the last subsection, we validate the generalization performance of our proposed approach using the long-term collected dataset \cite{mendoza2018long}.

\subsection{Configurations of the experimental analysis}\label{subsec:config_exp}
\subsubsection{A short description of the world model of the \acs{rfm} and \acl{fbp} methods}
As shown in the flowchart of the proposed approach (\figPref\ref{fig:system_ransac}), our proposal relies on the \acl{fbp} method and the world model of the \acs{rfm}. We employ \acf{ks} as the continuous representation of the \acs{rfm}, and \acs{knn} and its variant for \acl{fbp} without weakening the generosity. For further details about the world models of \acs{rfm} and \acl{fbp} can be found \eg in \citep{he2016wi,Zhou2018} and \citep{xia2017indoor}, respectively.

\begin{itemize}
	\item \acs{ks} is applied as the function for mapping a given location $ \vecSym{l} $ to the measurement $ \setSym{O} $, which is formulated as a kernelized regression problem. It can filter out the noise in the raw measurements and smoothing them (\figPref\ref{subfig:raw_rss_r1} and \figPref\ref{subfig:raw_rss_r2}). However, a trade-off has to be found for balancing the discontinuity and smoothness of the spatial distribution of the values of features \cite{10.1007/978-3-642-32645-5_18}. Because the discontinuity is part of the inherent characteristics of the \acs{rfm}, which is contributed from the indoor environments. The length scale $ \norSymScript{\lambda}{}{LS} $ of the spatial kernel is used for harmonizing these two aspects. In addition, we use the variant version of \acs{ks} for reducing the computational complexity by introducing a parameter named querying radius $\norSymScript{r}{}{KS}:= \norSymScript{\lambda}{}{KS} \cdot \norSymScript{\lambda}{}{LS}$ ($ \norSymScript{\lambda}{}{KS} > 1 $ is the scale factor).  A brief introduction to \acs{ks} and its variant are given in \appRef\ref{subsec:ks_variant} and \appRef\ref{subsec:matern}. We have analyzed the impact of \acs{ks} querying radius on the \acs{ks} quality and the computational complexity in the appendix. Herein we fixed the length scale $ \norSymScript{\lambda}{}{LS}=1.0 $ and the scale factor $ \norSymScript{\lambda}{}{KS}=5 $ for the experimental analysis.
	
	\item We use \acs{knn} and its variant for \acl{fbp}. The variant is for adaptively handling the varying measurability of features by using a non-vector-based dissimilarity measure: \acf{cdm} \cite{Zhou2018ipin}. \acs{cdm} can be used to measure the dissimilarity between two measurements without priorly knowing the total number of measurable features within the \acs{roi}, \ie it does not have to format all the measurement into vectors or a matrix. We focus on analyzing the results of \acs{knn} with \acs{cdm}. Two relevant parameters: i) the number of nearest neighbor $ k $ and ii) regularization factor $ \norSymScript{\lambda}{}{CDM} $ of \acs{cdm}, are taken into account. In \cite{Zhou2018ipin}, the detailed analysis of \acs{cdm} and its applications to \acl{fbp} are given. Following from \cite{Zhou2018ipin}, we set the parameters, \ie $ k $ and $ \norSymScript{\lambda}{}{CDM} $ of \acs{knn} with \acs{cdm} to 3 and 3.0, respectively.
\end{itemize}

\subsubsection{Raw datasets}
Two datasets are used for experimental analysis in this paper (\tabPref\ref{tab:summaryRFMs}). They are collected by Zhou \etal~using the same hardware (Nexus 6P) at the same office building \cite{Zhou2018} and \acs{rss} from all opportunistically measurable \acs{wlan} \acsp{ap} were recorded as the features. However, \textit{HIL-R2} was collected three and half months later than \textit{HIL-R1}. Because these two datasets were collected independently, the \acsp{ap} were  neither registered in the same sequence nor merged. The summarized characteristics of the raw datasets are given in \tabPref \ref{tab:summaryRFMs}.

\renewcommand{\arraystretch}{1.15}
\begin{savenotes}
	\begin{table}[!htb]
		\centering
		\caption{The characteristics of the datasets}
		\small
		\label{tab:summaryRFMs}
		\begin{tabular}{cp{0.1\columnwidth}p{0.1\columnwidth}p{0.1\columnwidth}cc}
			\noalign{\smallskip}\hline\noalign{\smallskip}
			\hspace{1.5ex}
			\multirow{2}{*}{Dataset} & \multirow{2}{*}{\shortstack{$ \mathrm{\sharp} $ Buildings}} &\multirow{2}{*}{$ \mathrm{\sharp} $ Floors} &\multirow{2}{*}{$ \mathrm{\sharp} $\acsp{ap}}  & \multirow{2}{*}{\shortstack{$ \mathrm{\sharp} $ Training\\ samples}} & \multirow{2}{*}{\shortstack{$ \mathrm{\sharp} $ Validation \\samples\tablefootnote{In case there is no provided validation samples, we randomly split the training samples into two datasets. 75\% of them are used for training and the remaining ones are used for validation.}}} \tabularnewline\tabularnewline
			\noalign{\smallskip}\hline\noalign{\smallskip}
			\textit{HIL-R1}&1&1&490&1525&509\tabularnewline
			\textit{HIL-R2}&1&1&434&3714&0\tabularnewline
			\noalign{\smallskip}\hline\noalign{\smallskip}
		\end{tabular}
	\end{table}
\end{savenotes}

\subsubsection{Preprocessing of the raw measurements}
\begin{figure*}[!h]
	\centering
	\subfloat[Raw \acs{rss}]{
		\label{subfig:raw_rss_r1}
		\includegraphics[width=0.3\linewidth]{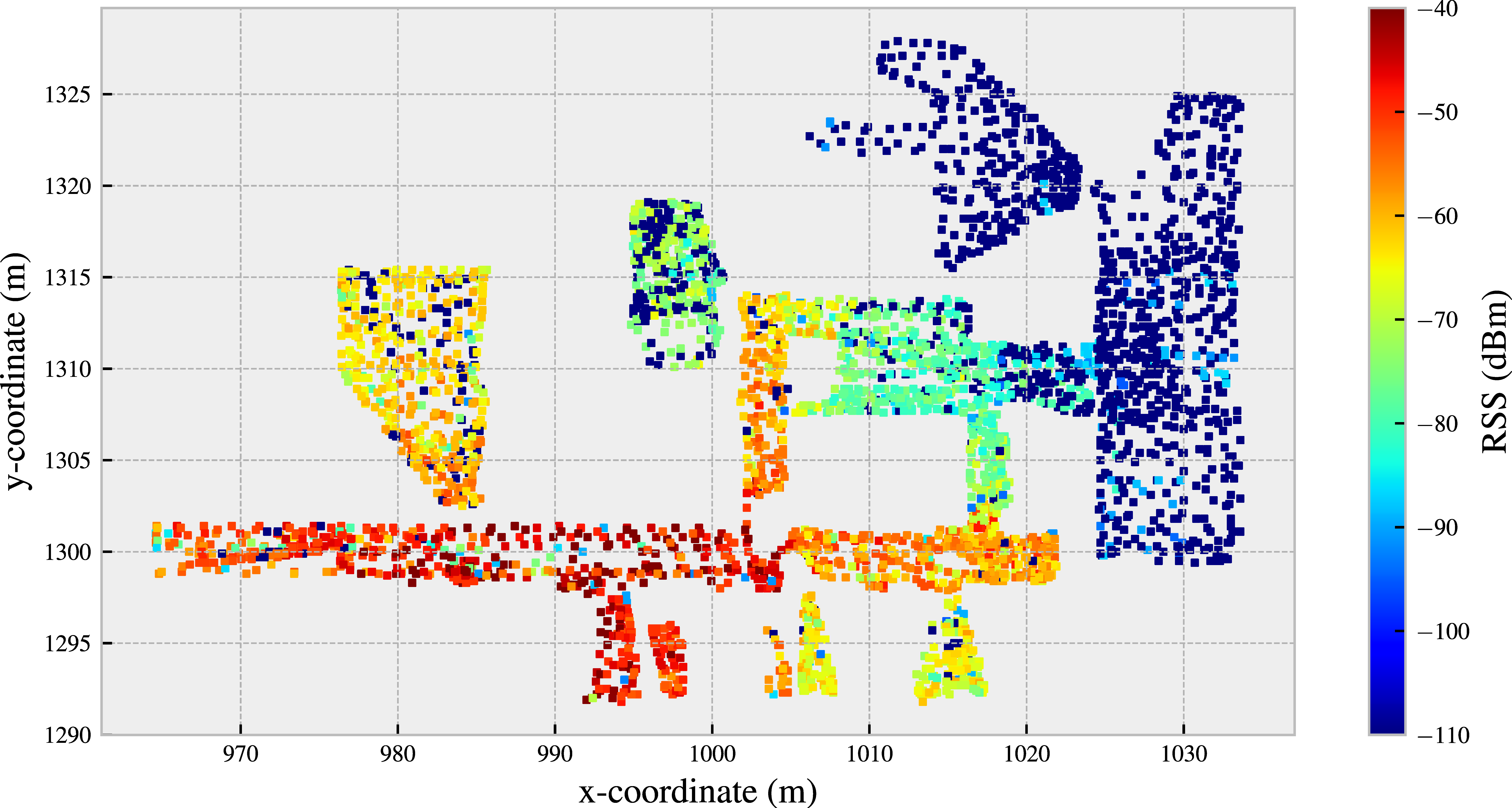}}\hspace{0.5ex}
	\subfloat[Smoothed \acs{rss}]{
		\label{subfig:ks_rss_r1}
		\includegraphics[width=0.3\linewidth]{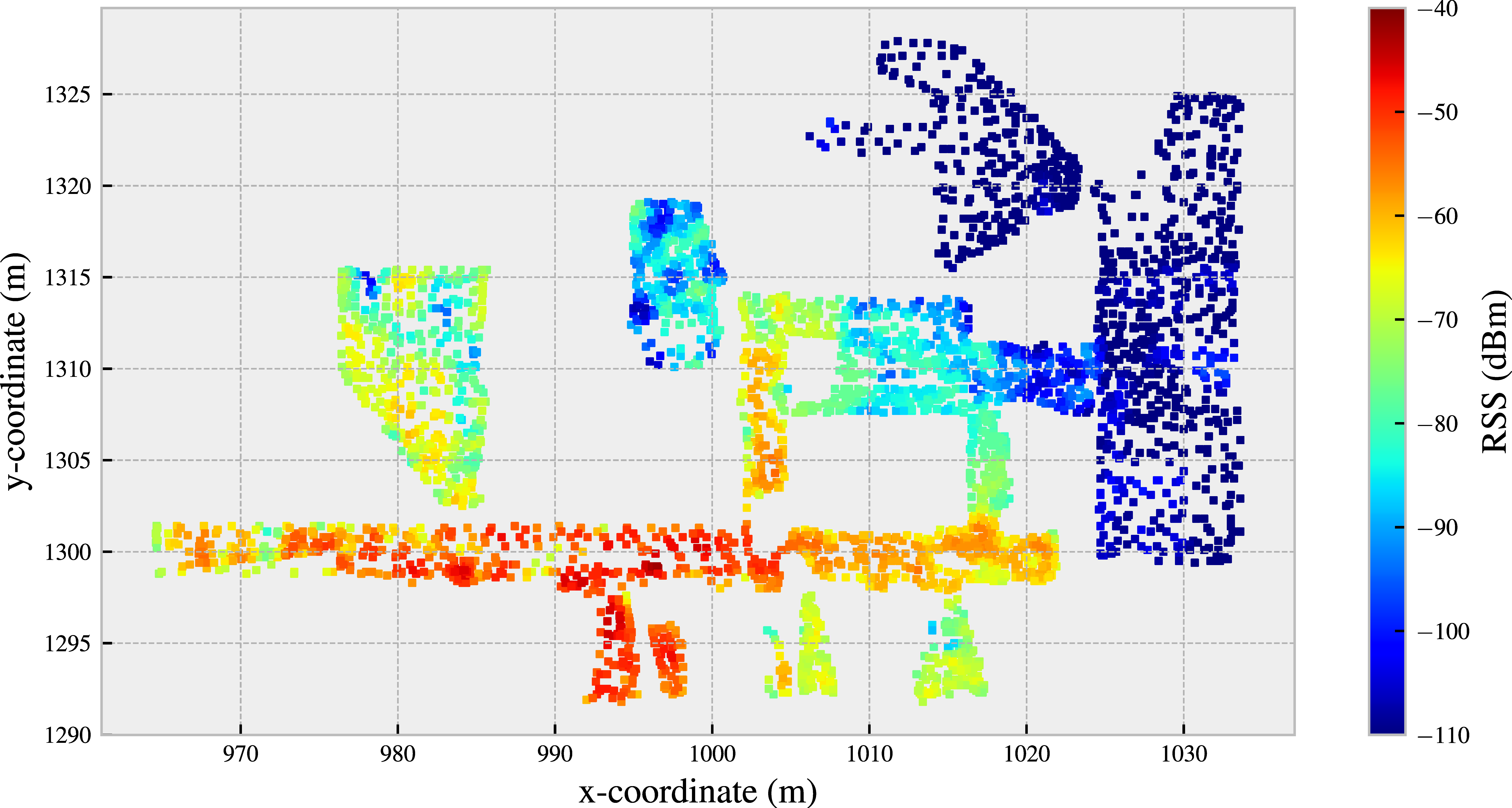}}\hspace{0.5ex}
	\subfloat[Interpolated \acs{rss}]{
		\label{subfig:qks_interp_raw_rss_r1}
		\includegraphics[width=0.3\linewidth]{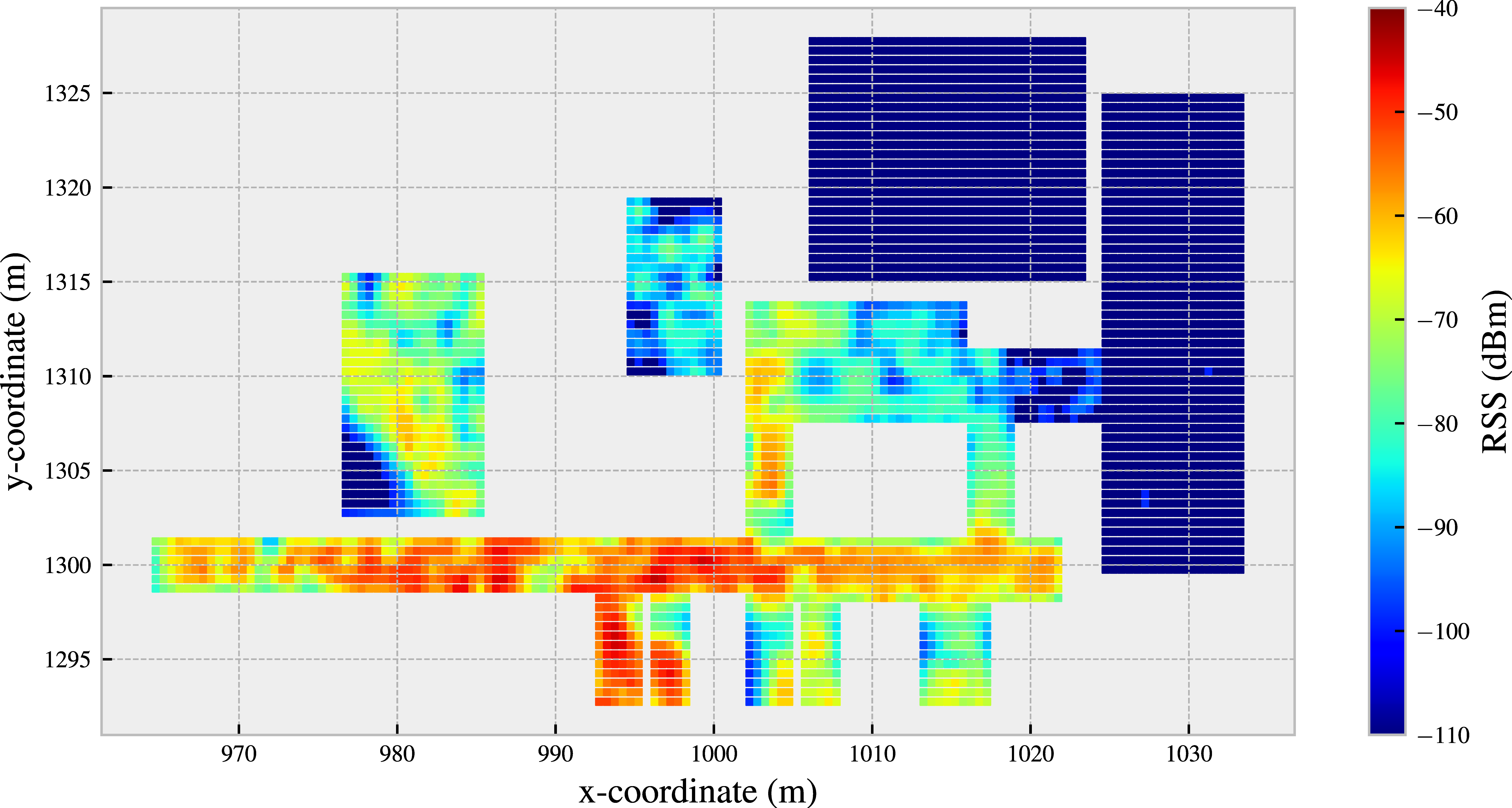}}\\
	\subfloat[Raw \acs{rss}]{
		\label{subfig:raw_rss_r2}
		\includegraphics[width=0.3\linewidth]{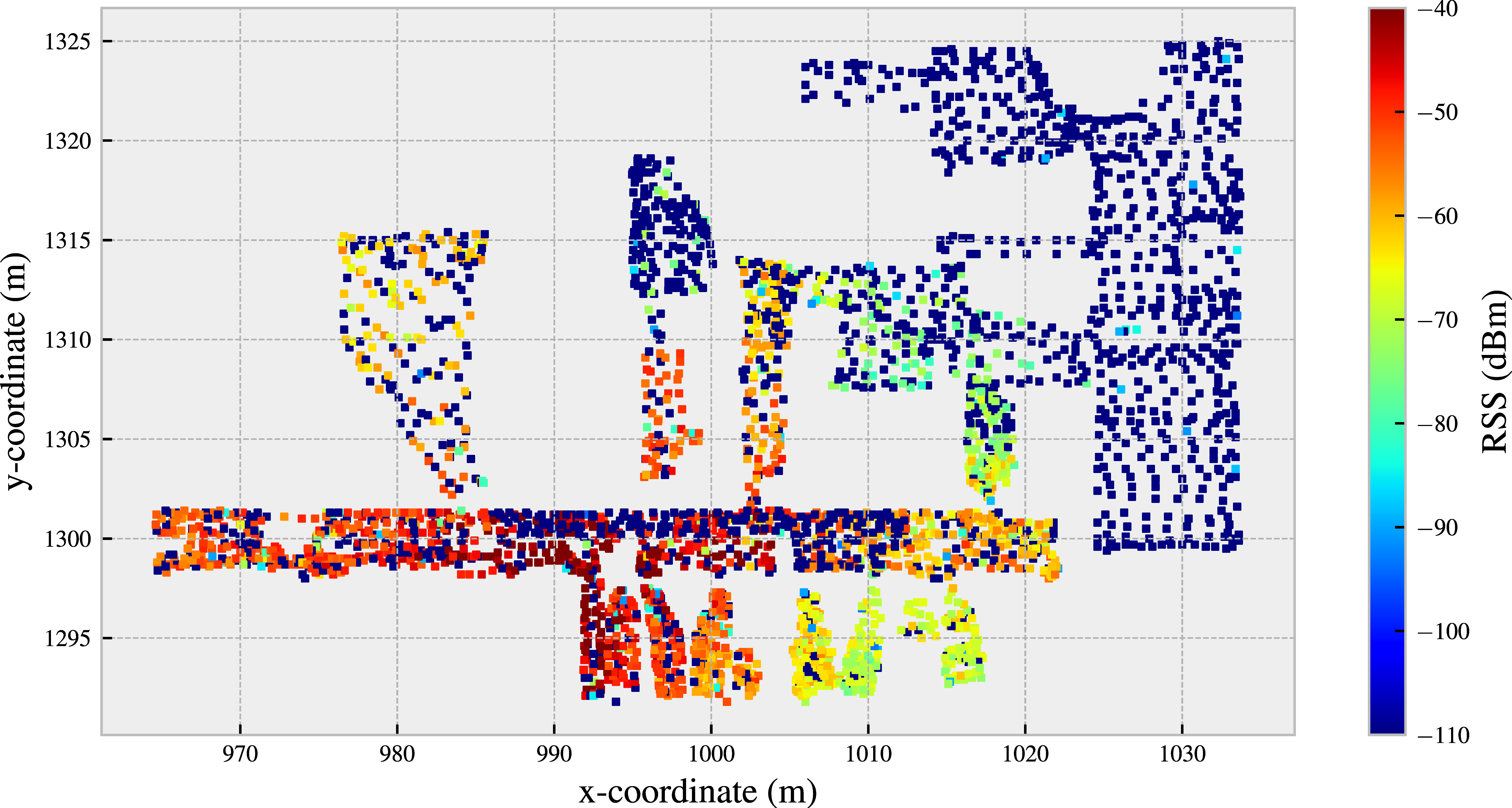}}\hspace{0.5ex}
	\subfloat[Smoothed \acs{rss}]{
		\label{subfig:ks_rss_r2}
		\includegraphics[width=0.3\linewidth]{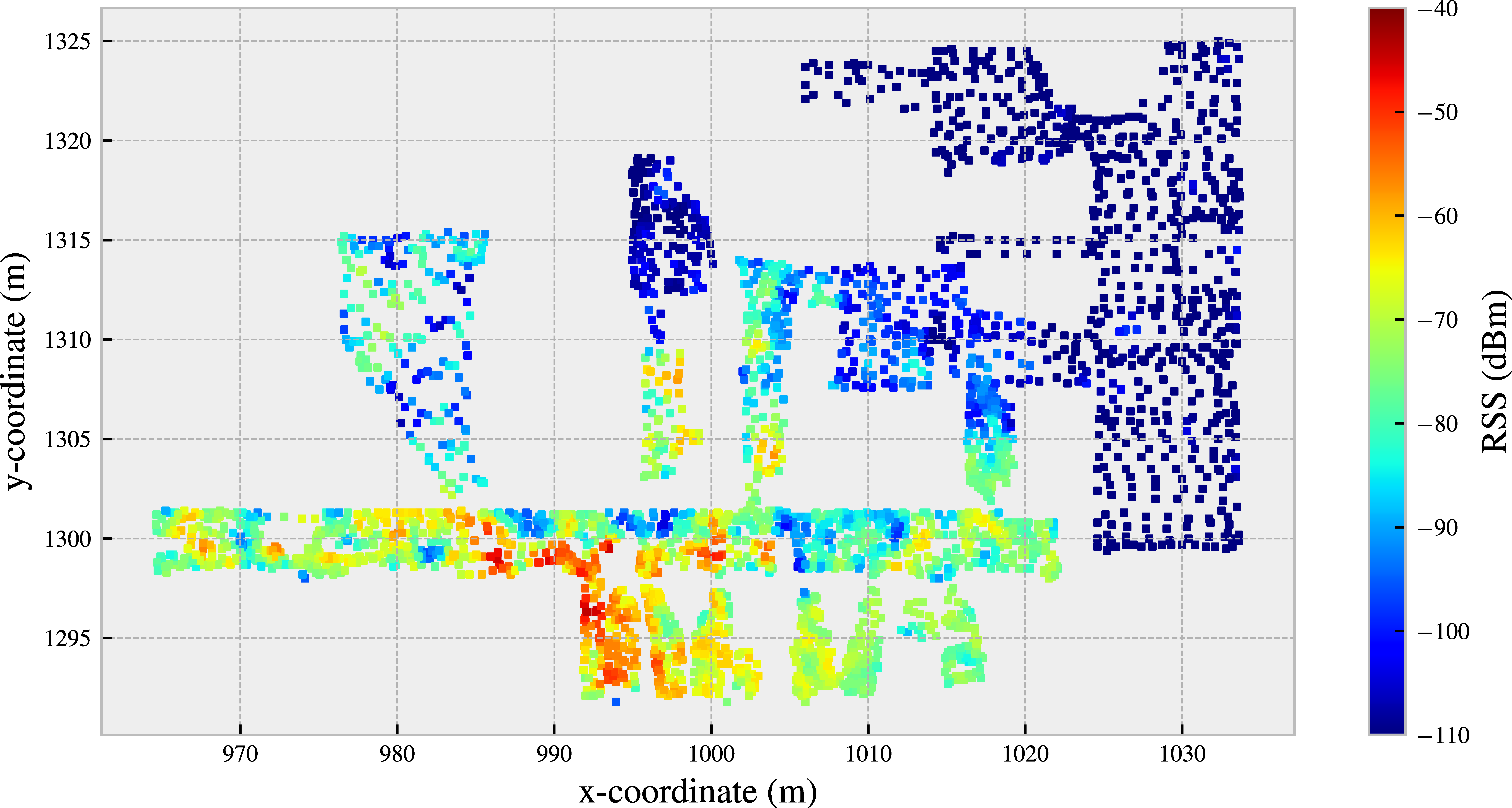}}\hspace{0.5ex}
	\subfloat[Interpolated \acs{rss}]{
		\label{subfig:qks_interp_raw_rss_r2}
		\includegraphics[width=0.3\linewidth]{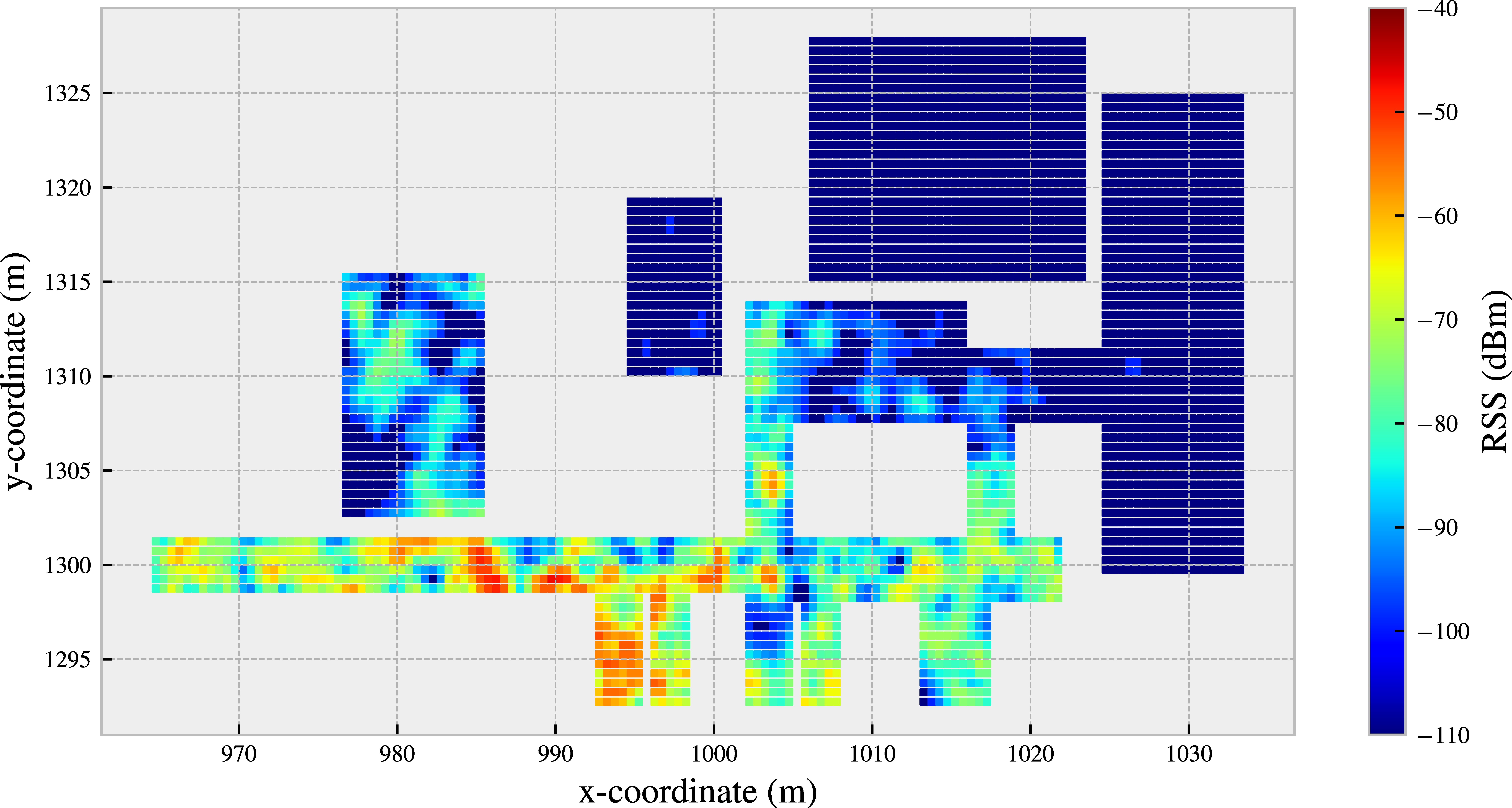}}\\
	\caption{Examples of raw, smoothed and interpolated \acs{rss} of an \acs{ap} (\acs{mac} is \textit{9c:50:ee:09:61:c4}) contained in \textit{HIL-R1} (The first row) and \textit{HIL-R2}. \acs{ks} extrapolates the data for the uncovered part of the \acs{roi} (\eg the upright part) but mostly fills with -110~dBm, which is the missing indicator value that used herein.}
	\label{fig:raw_ks_interp_r12}
\end{figure*}

We perform the preprocessing to the raw measurements for i) denoising and interpolating the \acs{rfm}; and ii) generating/simulating the dataset without changes. Both of them are achieved by \acs{ks}. The former is applied to the training samples of both datasets \footnote{For the sake that we need split \textit{HIL-R2} dataset into training and validation by ourself, we only apply \acs{ks} on the part that used for training.}. As shown in \figPref\ref{fig:raw_ks_interp_r12}, \acs{ks} is capable of filtering out the noise from the raw measurements. We generate the \acs{rfm} by interpolating a discrete grid size of 0.5-\text{by}-0.5~$\mathrm{m^2} $ over the \acs{roi} based in the raw measurements.

The latter is only applied to dataset \textit{HIL-R1}, which is used for generating the validation dataset for change detection evaluation. In order to generate it, we need a dataset which does not contain any changes. It is difficult to ensure this in reality when performing the data collection. The compromise that we made is smoothing the validation data using \acs{ks} according to the training data. In this way, we at least ensure that the pattern of the smoothed validation data is the same as the training data. We thus can assume that no changes has occurred in the smoothed validation dataset.

\subsubsection{Generation of the data for performance evaluation}
Using the smoothed validation dataset from \textit{HIL-R1}, we generate the data used for evaluating the performance of change detection. In order to cover varying types of possible changes might happen to the measurements, we simulate the changes covering three ranges: i) the percentage of  the measurable \acsp{ap} is set to be missed and is from 0 to 50\% in the interval of 10\%; ii) the percentage of the values of the measurable \acsp{ap} is shifted and is from 0\% to 50\% in the interval of 10\%; and iii) the shift value of the randomly selected visible \acsp{ap} is in the range of -15~dBm to -5~dBm and 5~dBm to 15~dBm in the interval of 5~dBm. But we ensure that the sum of missing ratio and shift ratio is no larger than 50\%, \ie the changed features are not dominated. We denote each combination of the simulated changes as a tuple containing three values (\eg (20\%, 10\%, -10~dBm)) for denoting the missing ratio, shift ratio, and shift value, respectively. Given a combination of the simulated changes, we arbitrarily introduce the corresponding changes to each sample stored in smoothed validated data of \textit{HIL-R1}.

\subsubsection{The usage of dataset \textit{HIL-R2}}
Dataset \textit{HIL-R2} was collected in the same \acs{roi} as \textit{HIL-R1}, and it can be used as the new \acs{rfm} for \acl{fbp}. We use it for two purposes:
\begin{itemize}
	\item Positioning performance evaluation: we estimate the locations of \textit{HIL-R2} and compare the performance with our proposed robust positioning scheme.
	\item Change detection evaluation: Though the changes that occurred in the \textit{HIL-R2} are unknown, but we expect that the positioning performance can be improved by using only the features detected as unchanged.
\end{itemize}

\subsubsection{The metrics for performance evaluation}
In order to evaluate the performance of the proposal, we use two types metrics, which are widely applied in positioning and change detection (\ie classification), respectively:
\begin{itemize}
	\item Positioning performance:  Empirical cumulative distribution function (\acs{ecdf}) \acs{wrt} error distances between the estimated locations and the ground truth is employed herein.
	\item Change detection: The confusion matrix is a table containing the information of comparing the detected changes to the actual ones. Four terms, i) \acf{tp}: correctly detected cases where there are changes features; ii) \acf{fn}: incorrect decisions that changes actually happened; iii) \acf{tn}: unchanged cases which are correctly detected as negative; and iv)\acf{fp}: changes which are incorrectly detected,  are associated to it and are used for defining following metrics:
	\begin{enumerate}
		\item True positive rate (\acs{tpr}, \ie sensitivity): The portion of true positives which are correctly detected as changed, \ie $ \frac{TP}{TP + FN} $;\vspace{0.5ex}
		\item True negative rate (\acs{tnr}, \ie specificity): The portion of true negative which are correctly detected as unchanged, \ie $ \frac{TN}{TN + FP} $;\vspace{0.5ex}
	\end{enumerate}
	In order to mitigate the influence of the change belief threshold on detecting the changes, we use the area under the \acf{roc} curve (\acs{auc} in short) for representing the achievable performance of change detection \citep{Murphy:2012:MLP:2380985}. The \acs{roc} curve is obtained by computing each pair of \acs{tpr} and \acs{tnr} \acs{wrt} varying values of the change belief threshold in the range of $ \in[0, 1]$. The curve is plotted in the way of \acs{tnr} versus \acs{tpr} (see \figPref\ref{subfig:roc_auc}).
\end{itemize}
\subsubsection{The implementation}
All related algorithms are implemented in Python. Specifically, we implement joint robust positioning and feature-wise change detection based on scikit-learn, a Python package for machine learning\cite{scikit-learn}. The computational complexity evaluation (\ie the processing time) is obtained using the time package \footnote{{https://docs.python.org/3/library/time.html}} on a Windows 64-bit PC with the configuration of 32G RAM and Xeon CPU (12 cores, 3.5GHz).

\subsection{The influence of sampling trick on dispersiveness and bias} \label{subsec:disp_bias_ana}
\begin{figure}[!h]
	\centering
	\subfloat[]{\label{subfig:mr_50_sr_0_m_15}
	\includegraphics[width=0.22\linewidth]{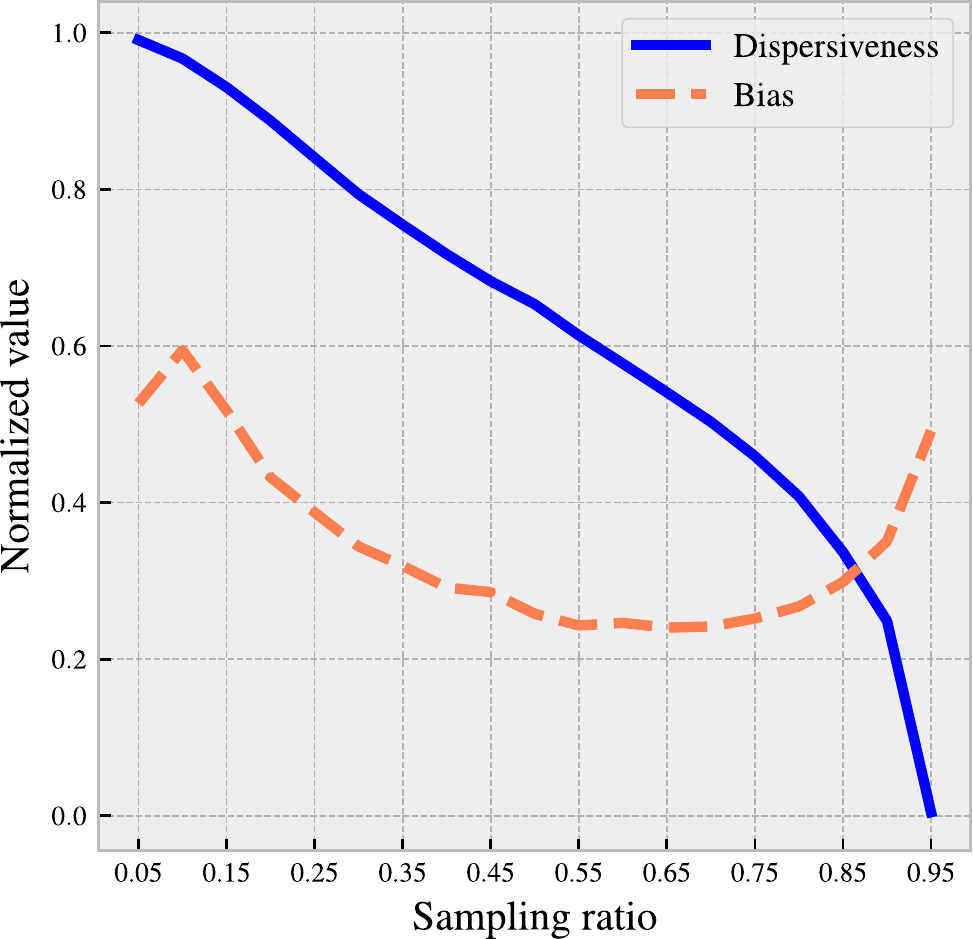}}\hspace{1ex}
	\subfloat[]{\label{subfig:mr_0_sr_50_m_15}
		\includegraphics[width=0.22\linewidth]{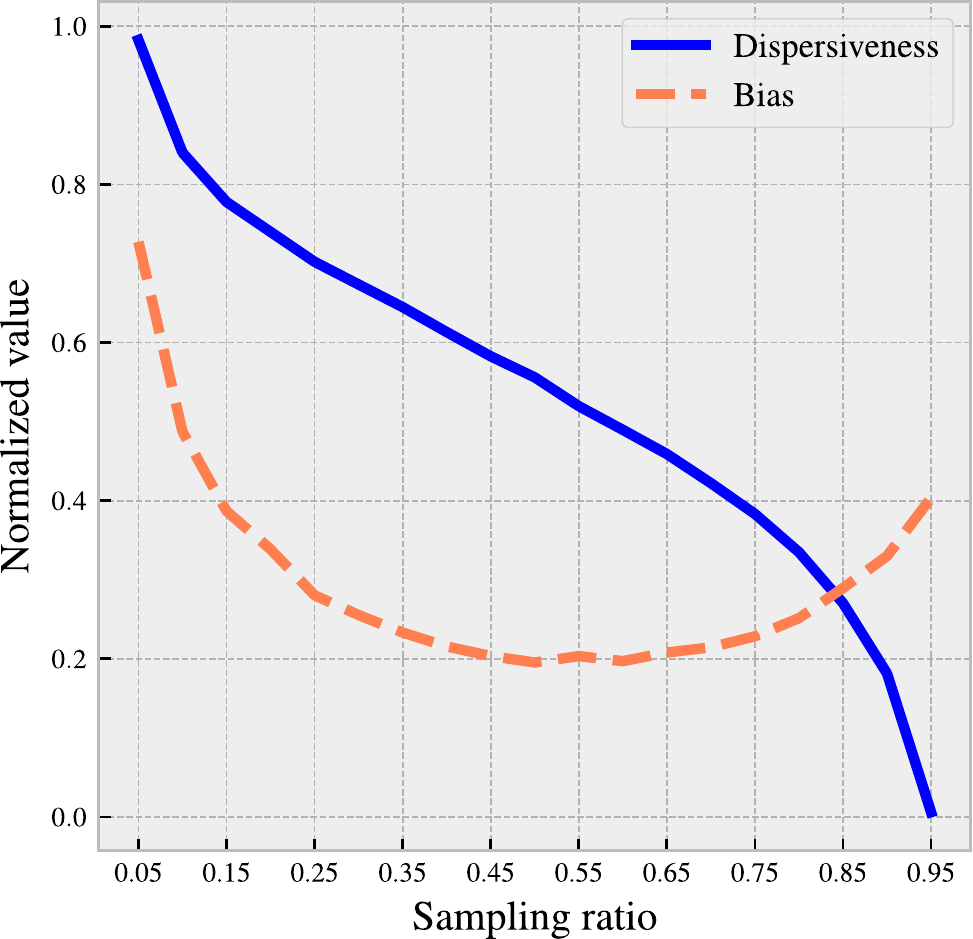}}\hspace{1ex}
	\subfloat[]{\label{subfig:mr_30_sr_20_m_15}
		\includegraphics[width=0.22\linewidth]{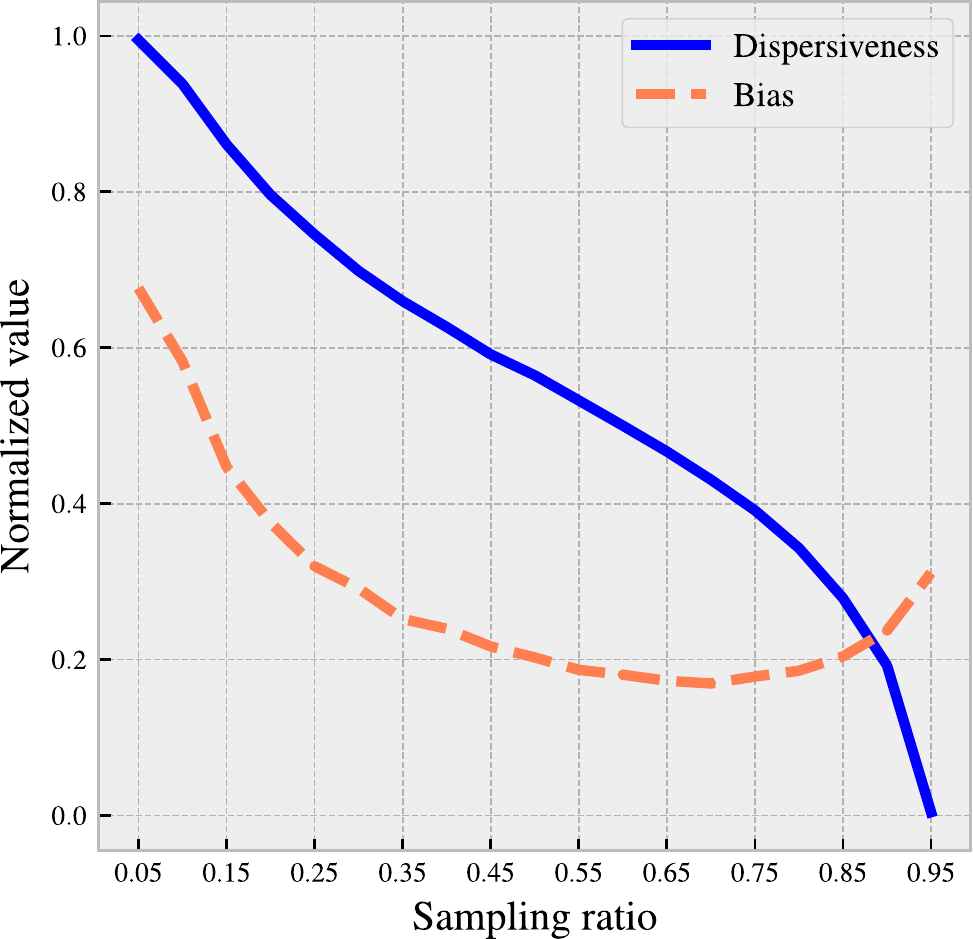}}\hspace{1ex}
	\subfloat[]{\label{subfig:mr_20_sr_30_m_15}
		\includegraphics[width=0.22\linewidth]{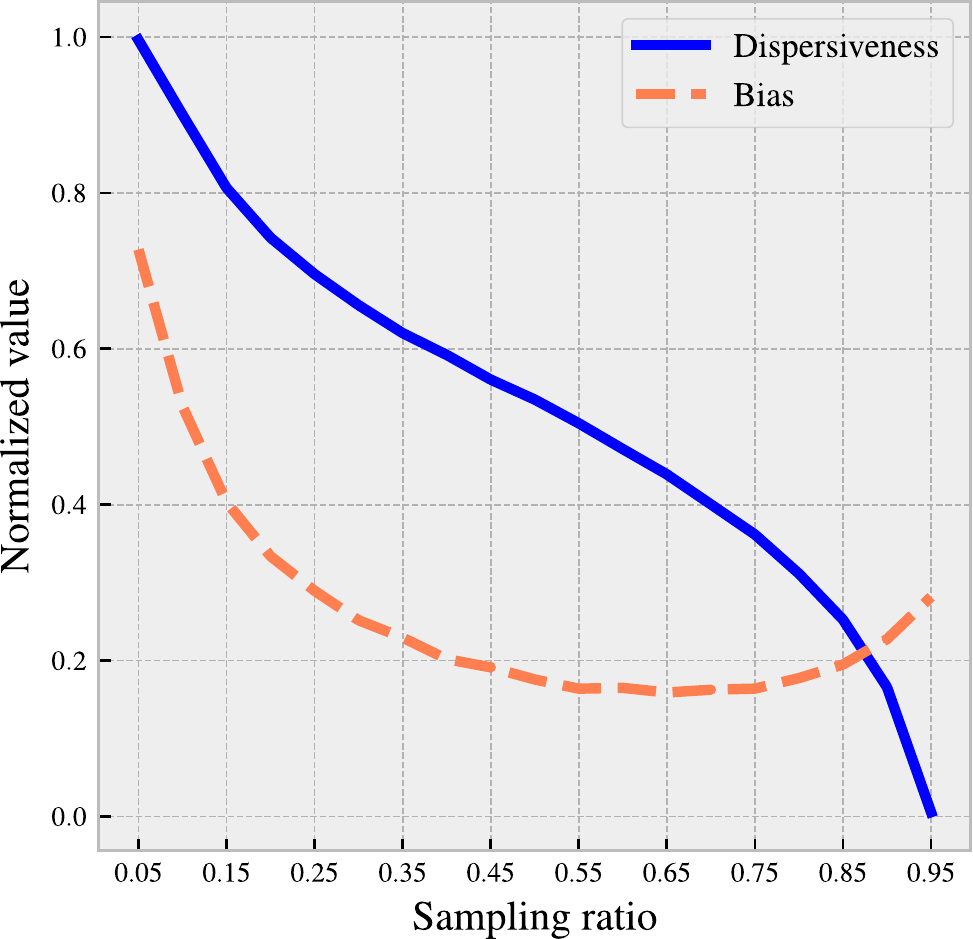}}
	\caption{Examples of dispersiveness and bias \acs{wrt} the sampling ratio. The four examples are illustrated the simulated changes combination of (a) (50\%, 0, -15 dBm), (b) (0, 50\%, -15 dBm), (c) (30\%, 20\%, -15 dBm), and (d) (20\%, 30\%, -15 dBm).}
	\label{fig:disp_bias_examples}
\end{figure}

Two parameters, sampling ratio $ \norSymScript{\alpha}{}{res} $ and number of resamples $ \norSymScript{N}{}{res} $, are introduced into the sampling step. We fix the value of $ \norSymScript{N}{}{res} $ to 200 and focus on carrying out a throughout analysis of the influence of $  \norSymScript{\alpha}{}{res}  $ on the dispersiveness and bias. For each example in the validation dataset, we sample 200 resamples in which each of them consists of a subset of arbitrarily selected measured features. The sampling ratio is from 5\% to 95\% in the interval of 5\%. In order to ensure that the sampled observation can be treated as a reasonable one, we sample at least three measured features if the given percentage of them is less than three. We compute the dispersiveness and bias as defined in \secPref\ref{subsec:fin_match} based on 200 locations estimated using the resamples. In \figPref\ref{fig:disp_bias_examples} shows the mean of the normalized dispersiveness and bias \acs{wrt} the sampling ratio. As the increasing of the sampling ratio, the dispersiveness reduces monotonically but the bias is obvious in the U-shape. We thus can find a proper range of sampling ratio such that it achieves small bias but keeps relative high dispersiveness. It means that there would be locations which are close to the ground truth location owing to the low bias and the chance to locate close to the ground truth is high because of the relative high dispersiveness. Intuitively, it is not reasonable that the value of dispersiveness is too high because it requires the higher number of resamples to ensure that at least one of the estimated position located closed enough to the ground truth.

\begin{figure}[!h]
	\centering
	\subfloat[-15 dBm]{\label{subfig:sr_left_m_15}
	\includegraphics[width=0.23\linewidth]{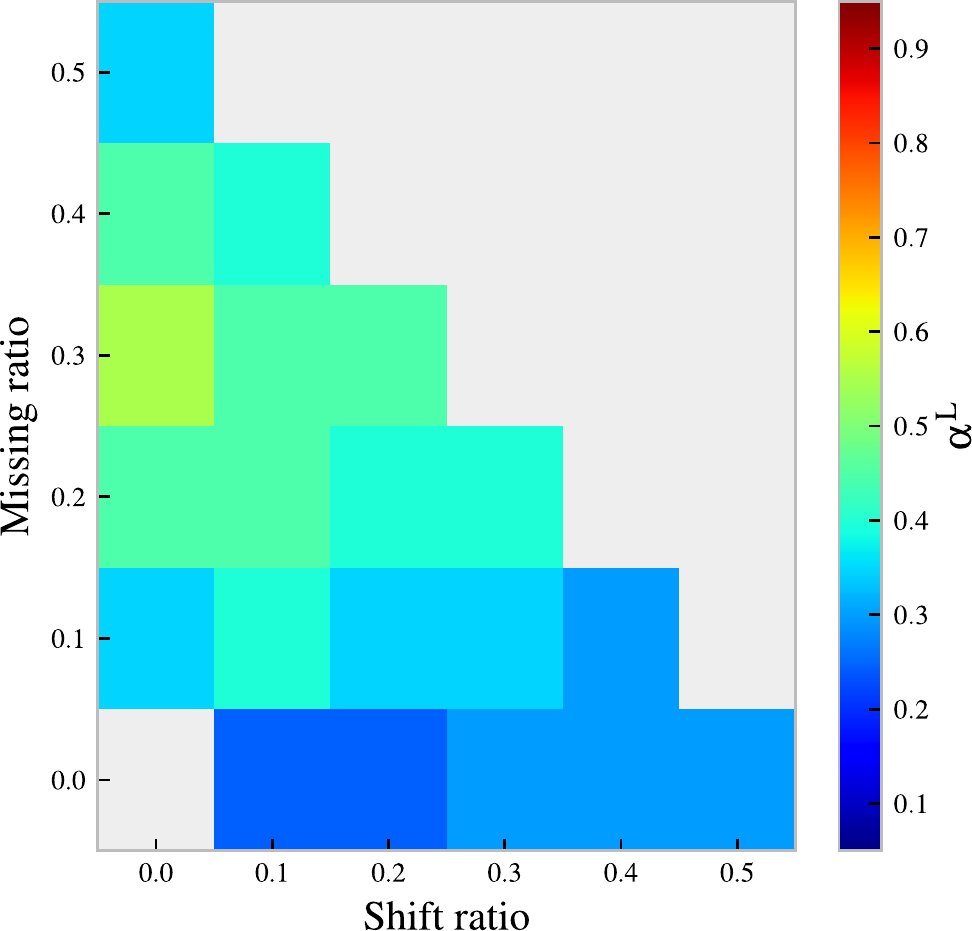}}\hspace{1ex}
	\subfloat[-10 dBm]{\label{subfig:sr_left_m_10}
	\includegraphics[width=0.23\linewidth]{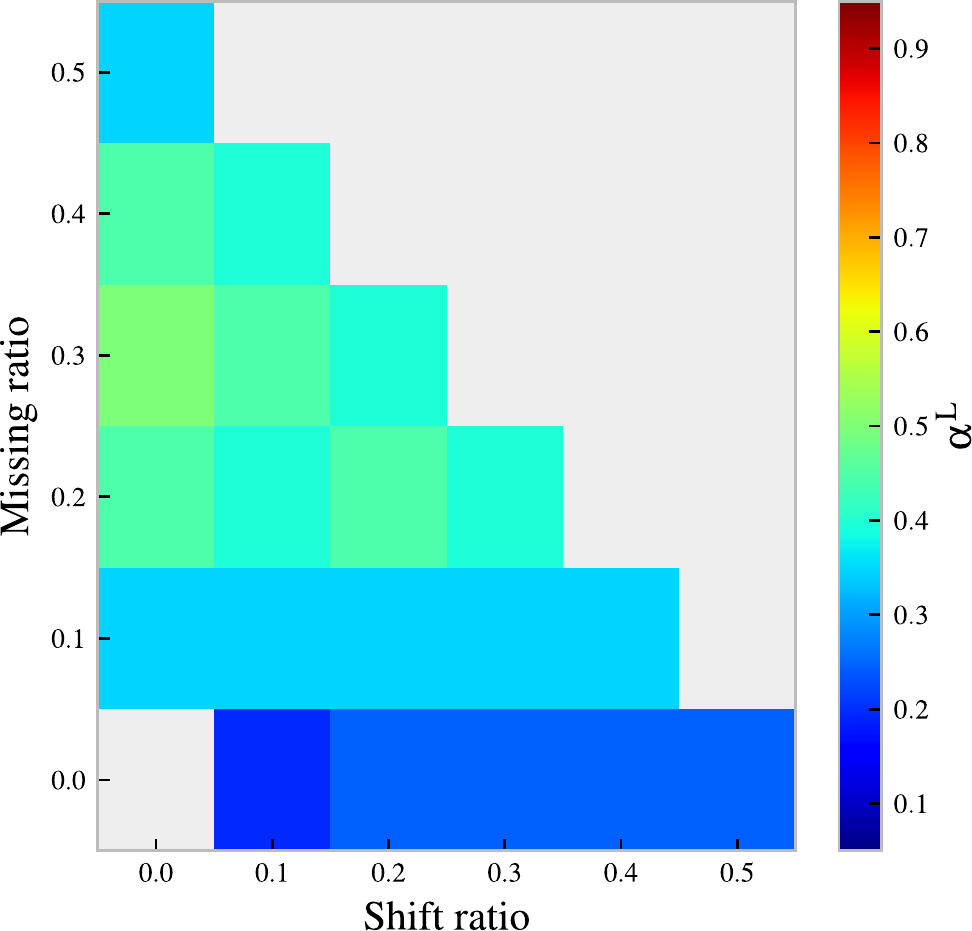}}\hspace{1ex}
	\subfloat[10 dBm]{\label{subfig:sr_left_p_10}
	\includegraphics[width=0.23\linewidth]{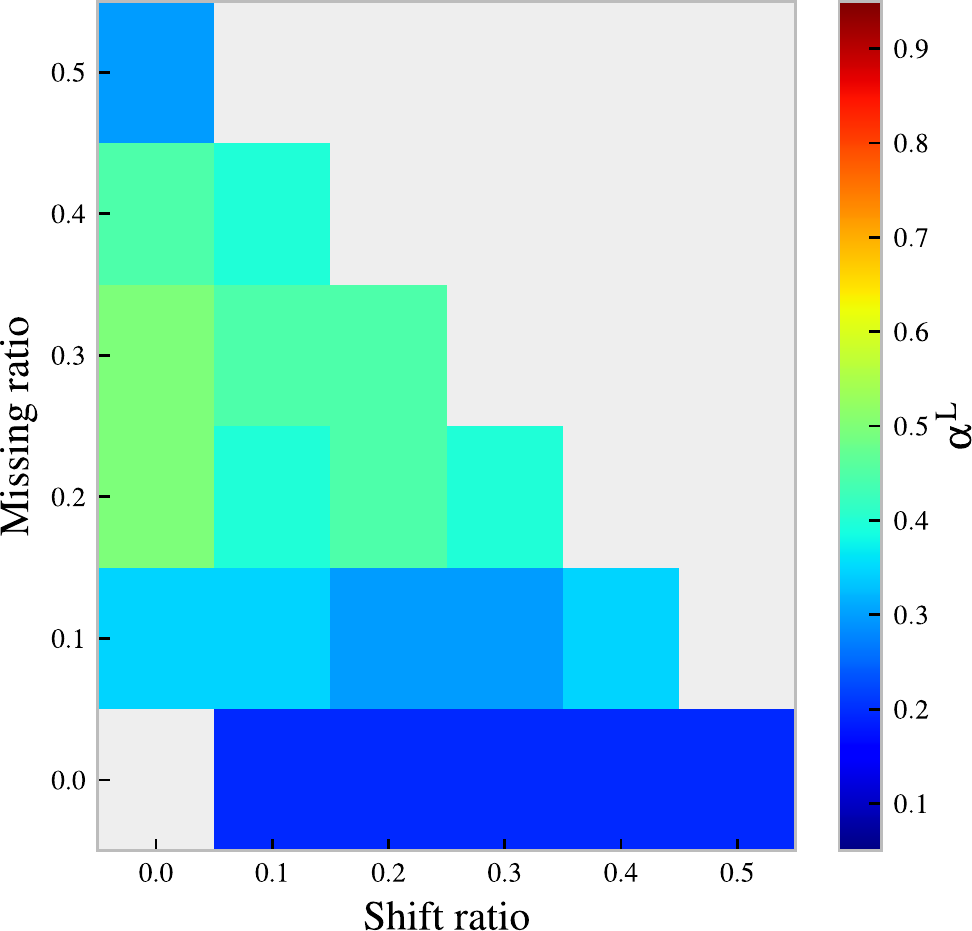}}\hspace{1ex}
	\subfloat[15 dBm]{\label{subfig:sr_left_p_15}
	\includegraphics[width=0.23\linewidth]{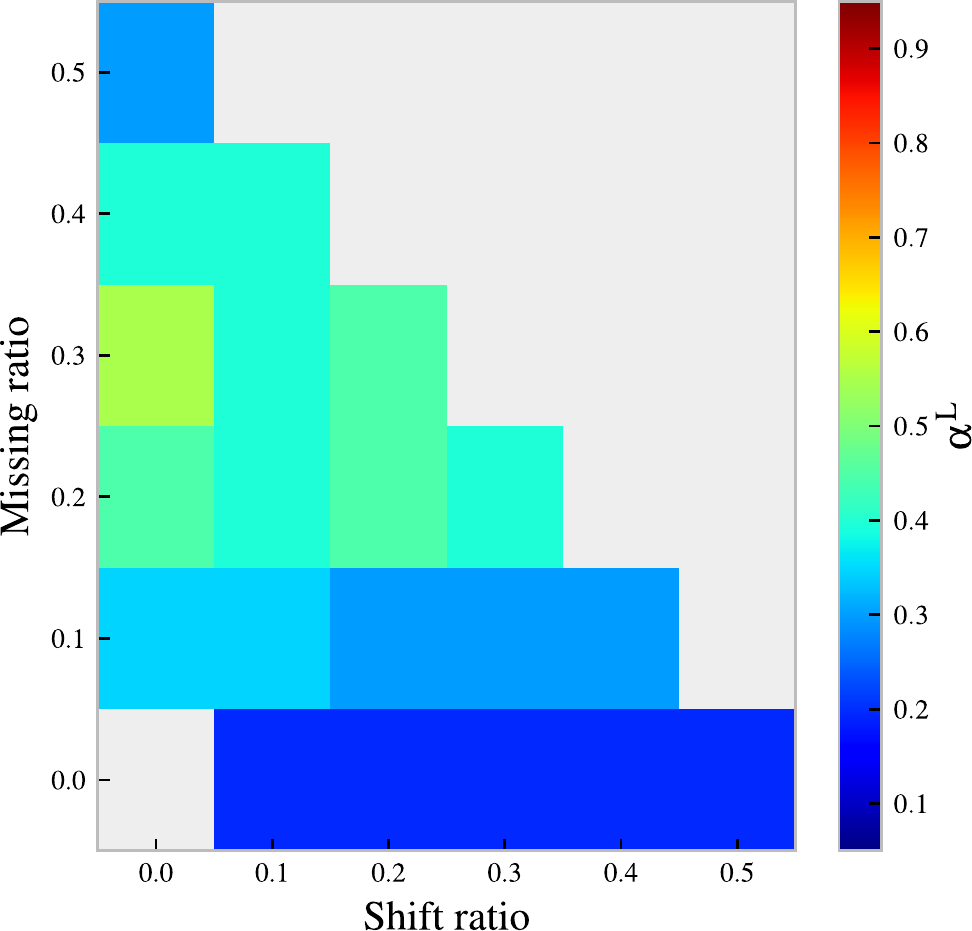}}\\
	\subfloat[-15 dBm]{\label{subfig:sr_min_m_15}
		\includegraphics[width=0.23\linewidth]{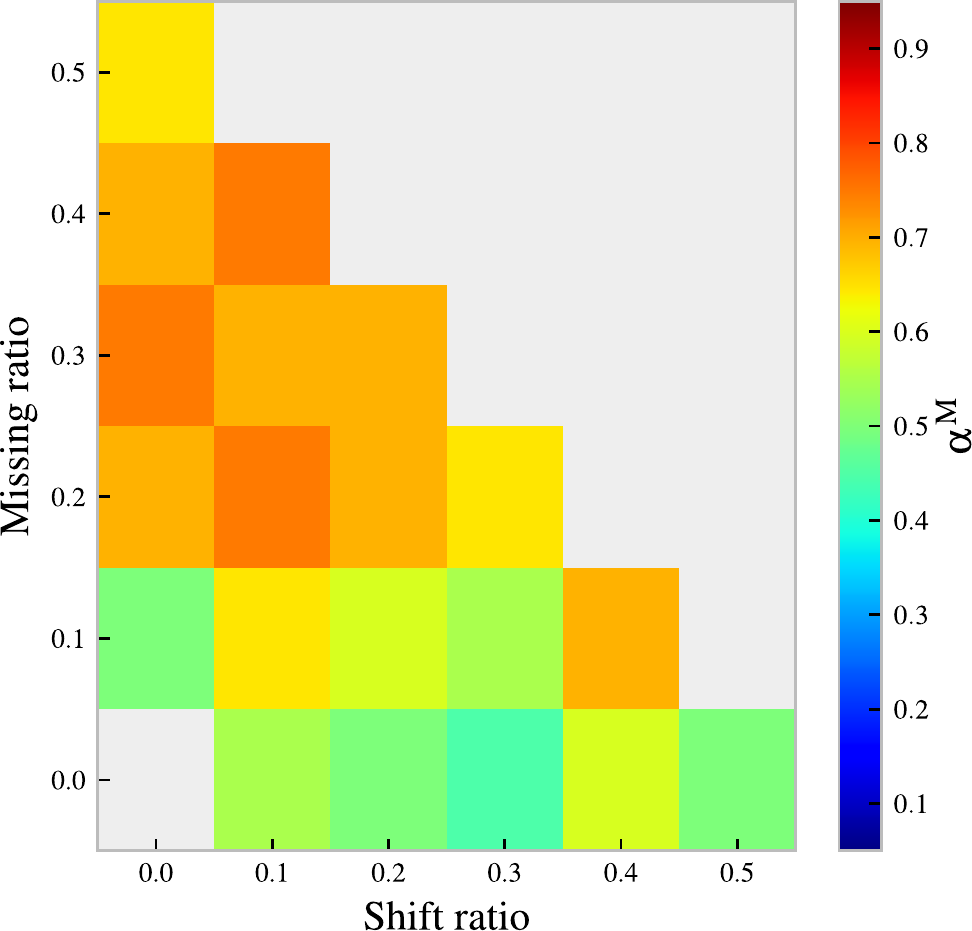}}\hspace{1ex}
	\subfloat[-10 dBm]{\label{subfig:sr_min_m_10}
		\includegraphics[width=0.23\linewidth]{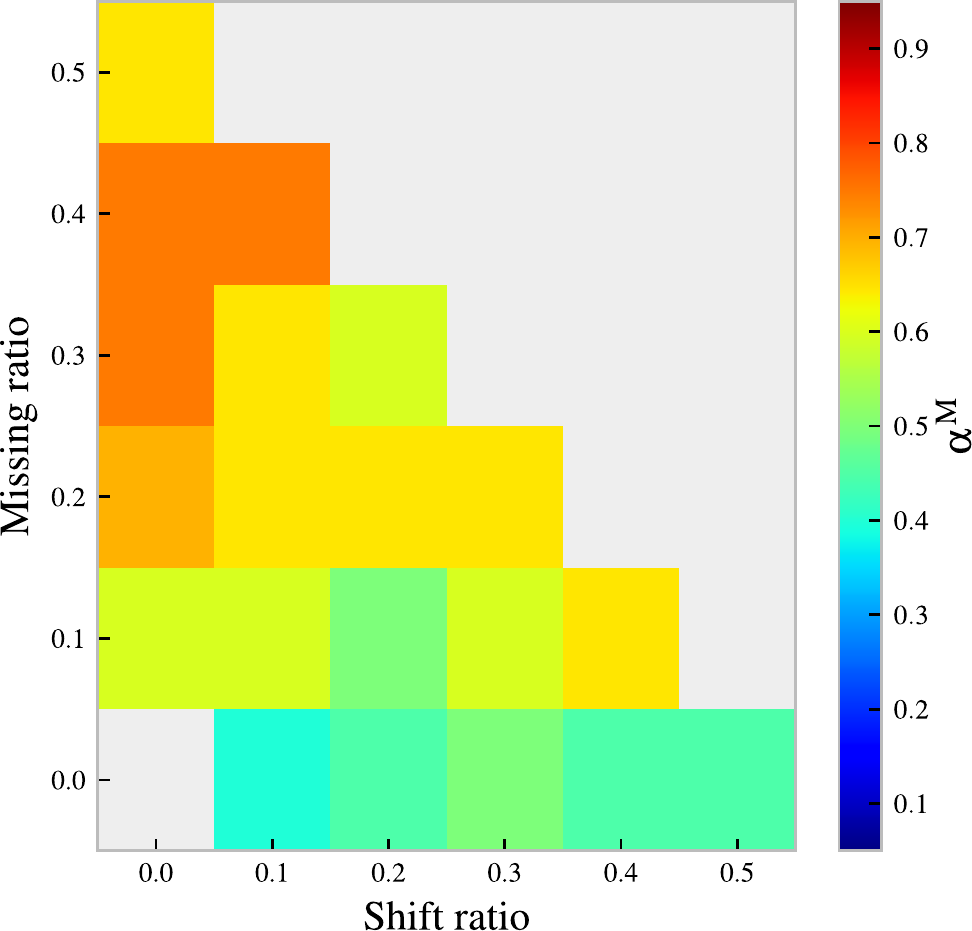}}\hspace{1ex}
	\subfloat[10 dBm]{\label{subfig:sr_min_p_10}
		\includegraphics[width=0.23\linewidth]{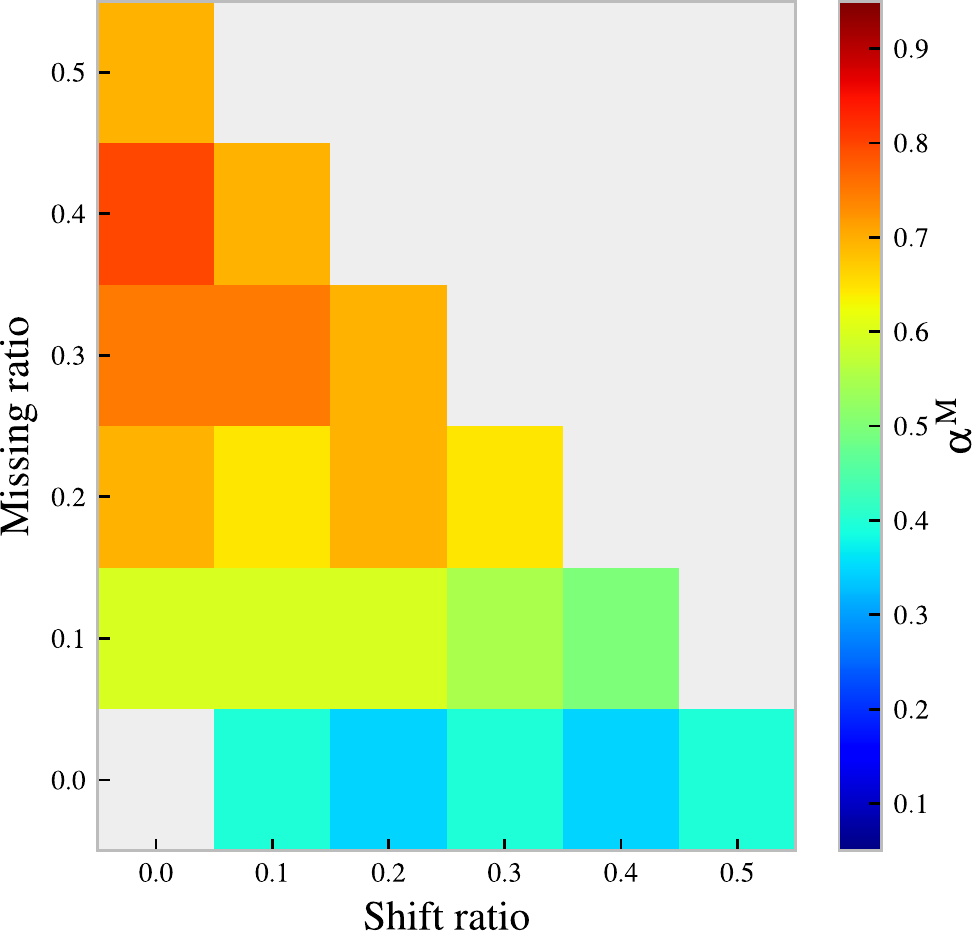}}\hspace{1ex}
	\subfloat[15 dBm]{\label{subfig:sr_min_p_15}
	\includegraphics[width=0.23\linewidth]{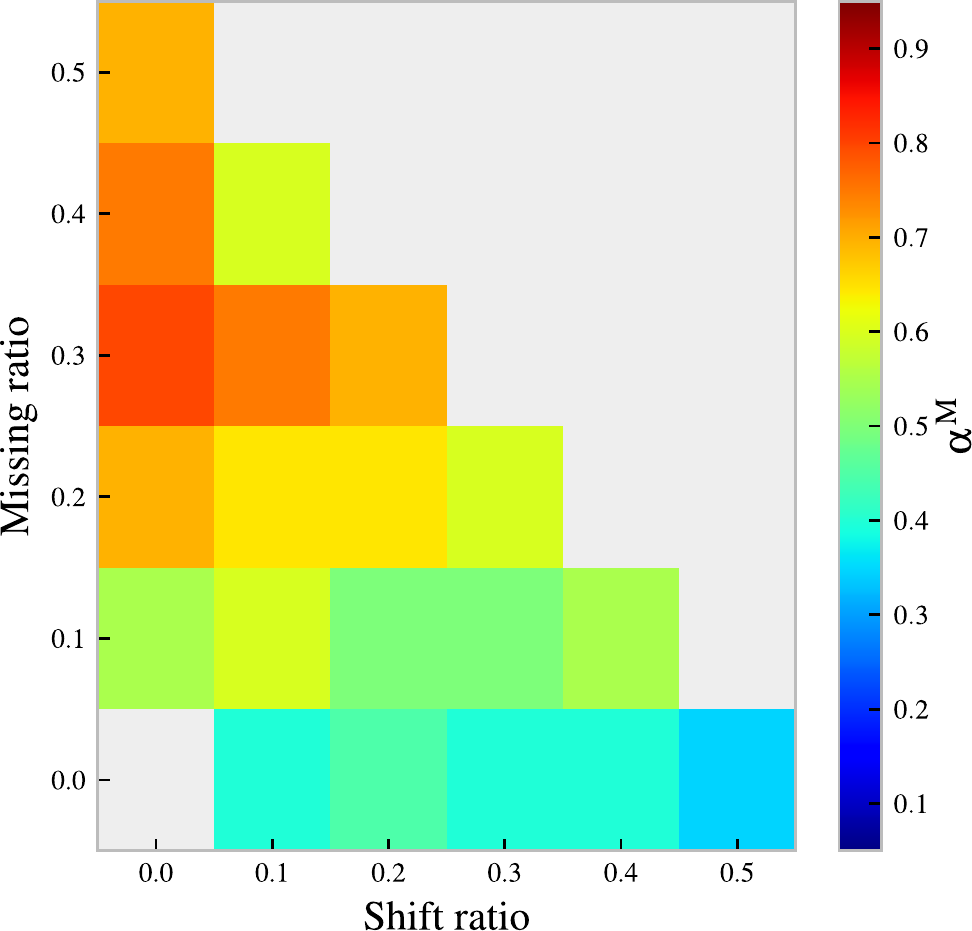}}
	\caption{The  sampling ratio $ \norSymScript{\alpha}{}{L} $ and $ \norSymScript{\alpha}{}{M} $  \acs{wrt} the varying changes. The gray area indicates that the corresponding changes have no been simulated.}
	\label{fig:sr_left}
\end{figure}

We also analyze the range of sampling ratio, which defined as similar to bandwidth, an established terminology in the signal processing field. The 3~dB bandwidth of the sampling ratio regarding the bias is defined as the sampling ratio interval $ [\norSymScript{\alpha}{}{L}, \,\norSymScript{\alpha}{}{R}] $ at which the bias is higher than $ \sqrt{2} $-times of the minimum bias \citep{Oppenheim:1996}. The sampling ratio that achieves the minimum bias is denoted as $ \norSymScript{\alpha}{}{M} $. According to the initial results, we find out that the values of $ \norSymScript{\alpha}{}{R}$ are not informative because they are saturated to the maximum sampling ratio 95\%. We thus focus on finding the lower bound of  $ \norSymScript{\alpha}{}{L} $ and the upper bound of $ \norSymScript{\alpha}{}{M} $, which are suitable for different types of simulated changes. We present the value of $ \norSymScript{\alpha}{}{L} $ \acs{wrt} varies of simulated changes  in \figPref\ref{fig:sr_left} (the 1st row). Regarding different changes, the value of $ \norSymScript{\alpha}{}{L} $ is from 25\% to 55\%. Combing with \figPref\ref{fig:disp_bias_examples}, the normalized dispersiveness is also relative high (close to 0.5) in the same range. As for $ \norSymScript{\alpha}{}{M} $, we can summarize from \figPref\ref{fig:sr_left} (the 2nd row) that its upper bound is up-to 75\%.  We choose three representative values of the sampling ratio, 45\%, 55\% and 65\% from this range for the analysis in the following sections and leave the work for finding the optimal value of the sampling ratio and the number samples for future.

\subsection{The performance of joint positioning and change detection}\label{subsec:perf_pos_change}
\subsubsection{Recovering the positioning performance}
\begin{figure}[!h]
	\centering
	\subfloat[45\%]{\label{subfig:ecdf_mr_45}
		\includegraphics[width=0.3\linewidth]{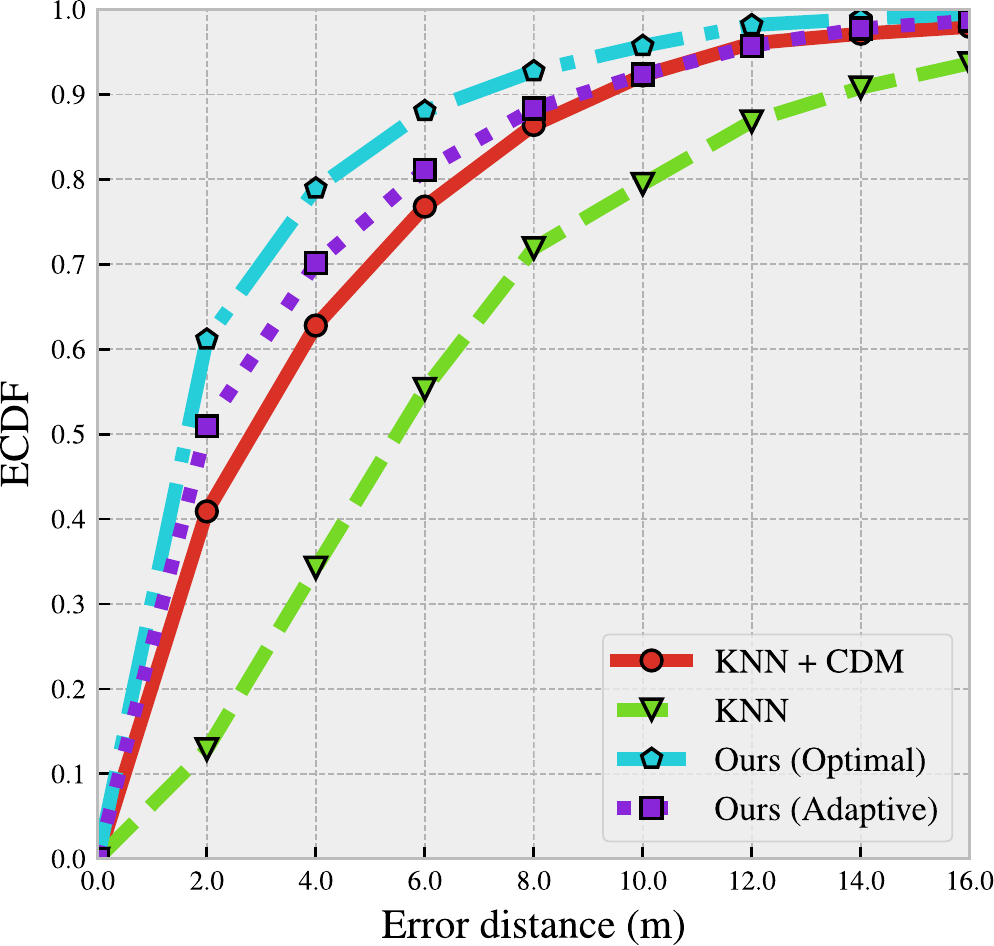}}\hspace{2ex}
	\subfloat[55\%]{\label{subfig:ecdf_mr_55}
		\includegraphics[width=0.3\linewidth]{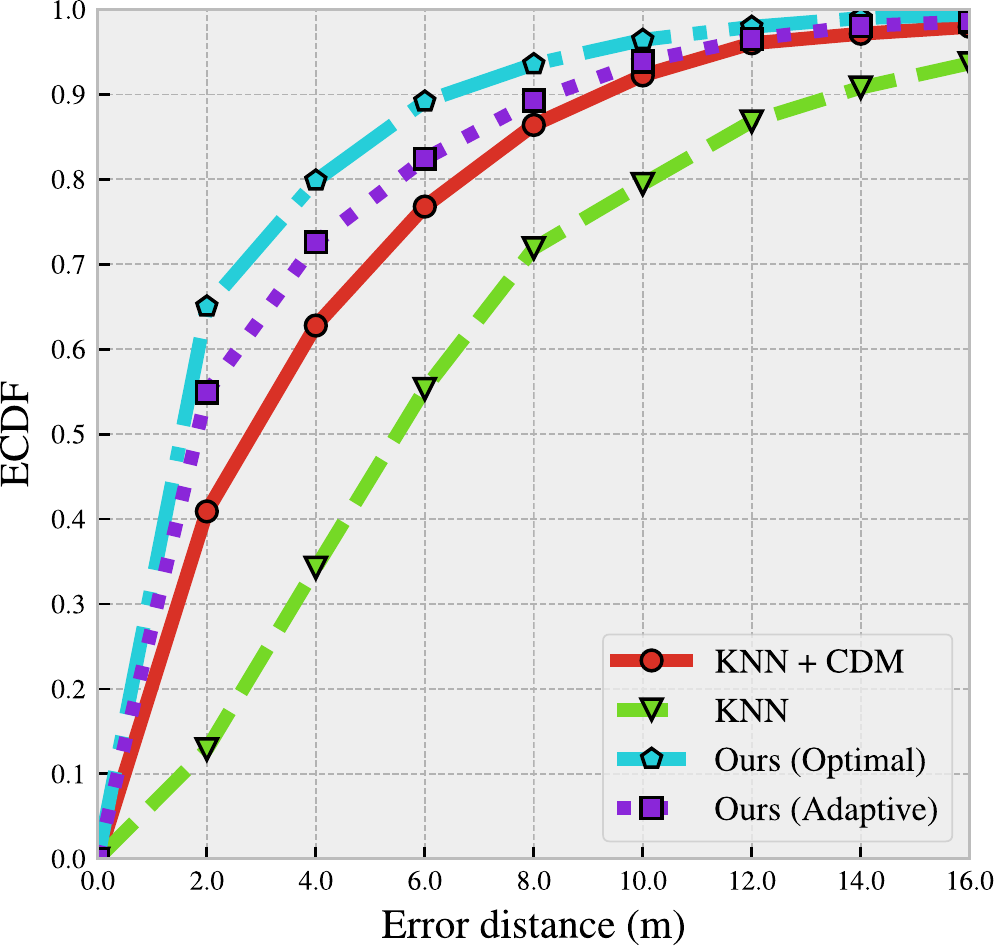}}\hspace{2ex}
	\subfloat[65\%]{\label{subfig:ecdf_mr_65}
		\includegraphics[width=0.3\linewidth]{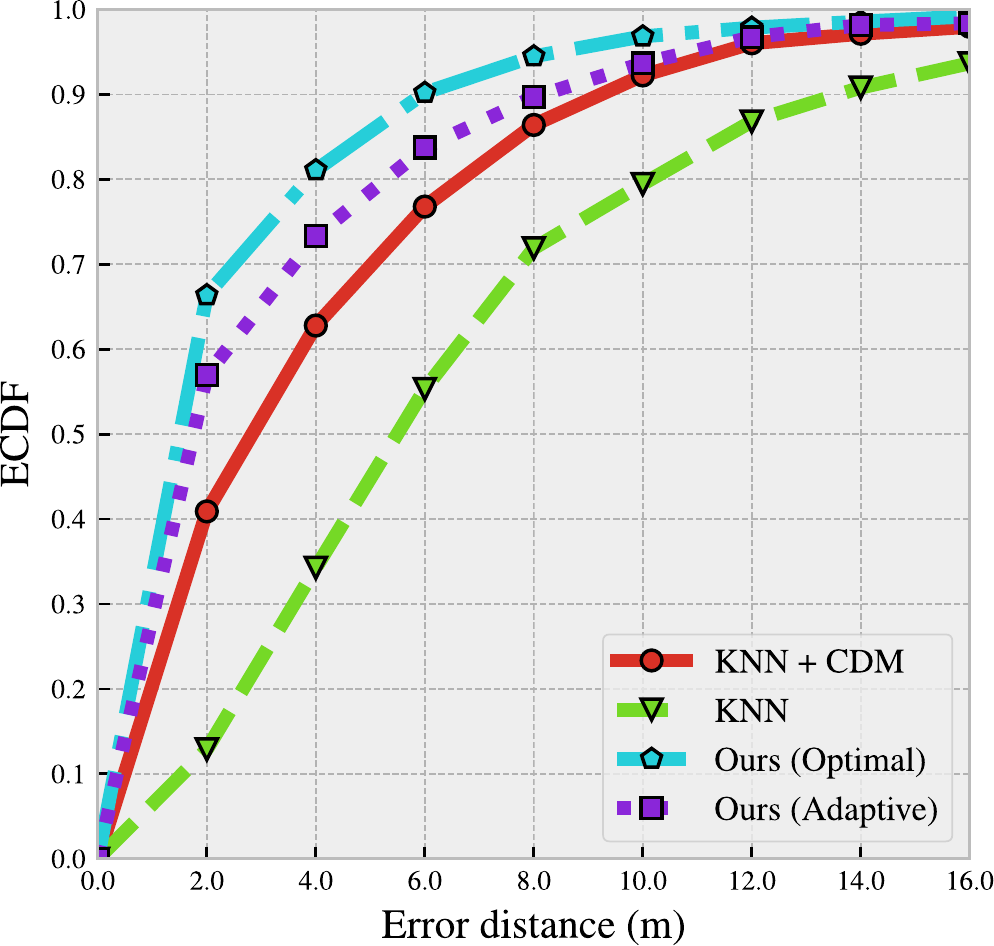}}
	\caption{The positioning accuracy of the simulated changes (50\%, 0, -15~dBm) using three sampling ratio. Fingerprinting-based positioning approaches, \acs{knn} and \acs{knn} with \acs{cdm}, are taken as the baseline. The result of the optimal (light blue-colored) is computed by assuming that the one closest to the ground truth among \acs{mji}-selected candidate positions is selected as the estimated location.}
	\label{fig:ecdf_mr_50}
\end{figure}

\figPref\ref{fig:ecdf_mr_50} shows the \acs{ecdf} \acs{wrt} the error distance for the simulated changes. The estimated location is computed by a weighted average of the candidate positions which are selected by ranking the values of \acs{mji}. We can conclude from the results that i) our proposed approach can recover and improve the positioning accuracy for against the changes. Comparing to \acs{knn} with \acs{cdm}, the improvement of the positioning accuracy within 2~m and 4~m are up-to 17\% and 11\% , respectively; ii) the proposed method is not sensitive to the sampling ratio. The positioning performances of different values of sampling ratio are better than that of two baseline approaches (\acs{knn} and \acs{knn} with \acs{cdm}). The variations of the positioning accuracy of different sampling ratio are small;  and iii) the potential improvement of the proposed approach is high if a scheme can ideally retrieve the closest one as the estimated location from the selected candidate positions (blue curves in \figPref\ref{fig:ecdf_mr_50}).

\subsubsection{Identifying feature-wise changes}\label{subsubsec:iden_feat}
\begin{figure}[!h]
	\centering
	\subfloat[Distribution of \acs{std}]{\label{subfig:std_hist}
		\includegraphics[width=0.341\linewidth]{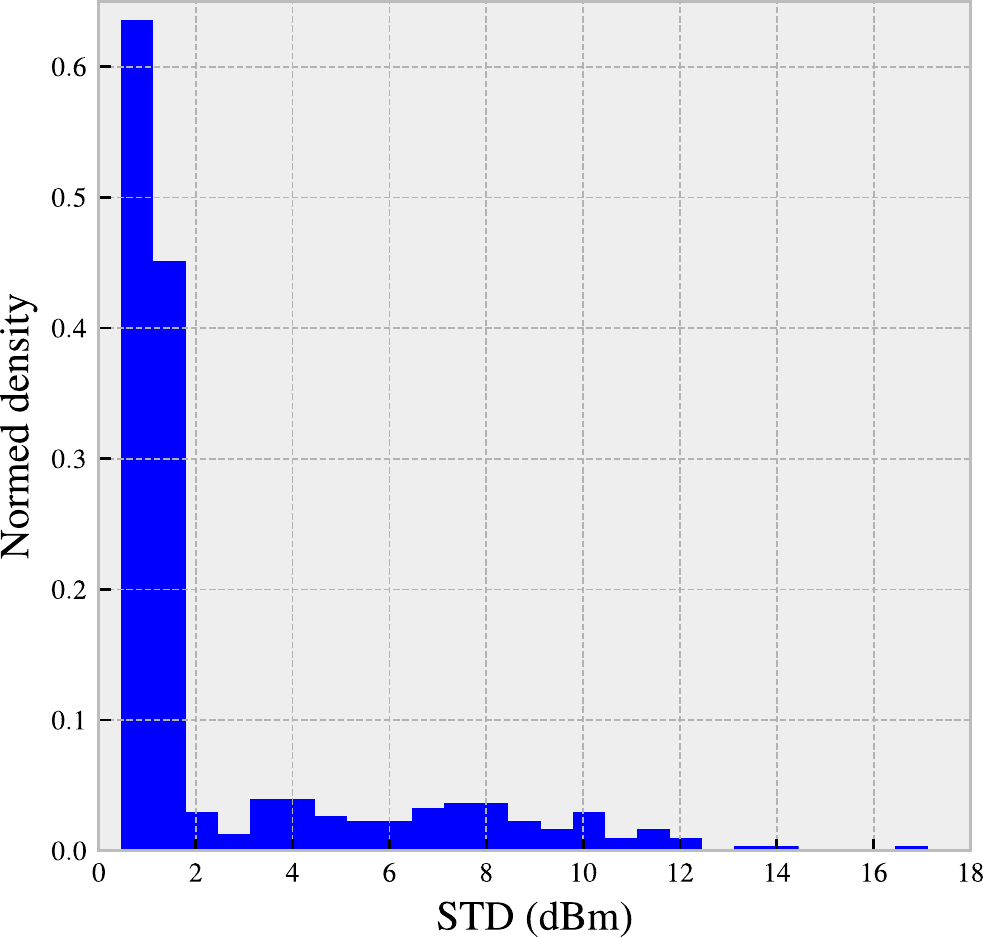}}\hspace{2ex}
	\subfloat[Least square (LS) fitted model]{\label{subfig:rss_std_ls}
		\includegraphics[width=0.35\linewidth]{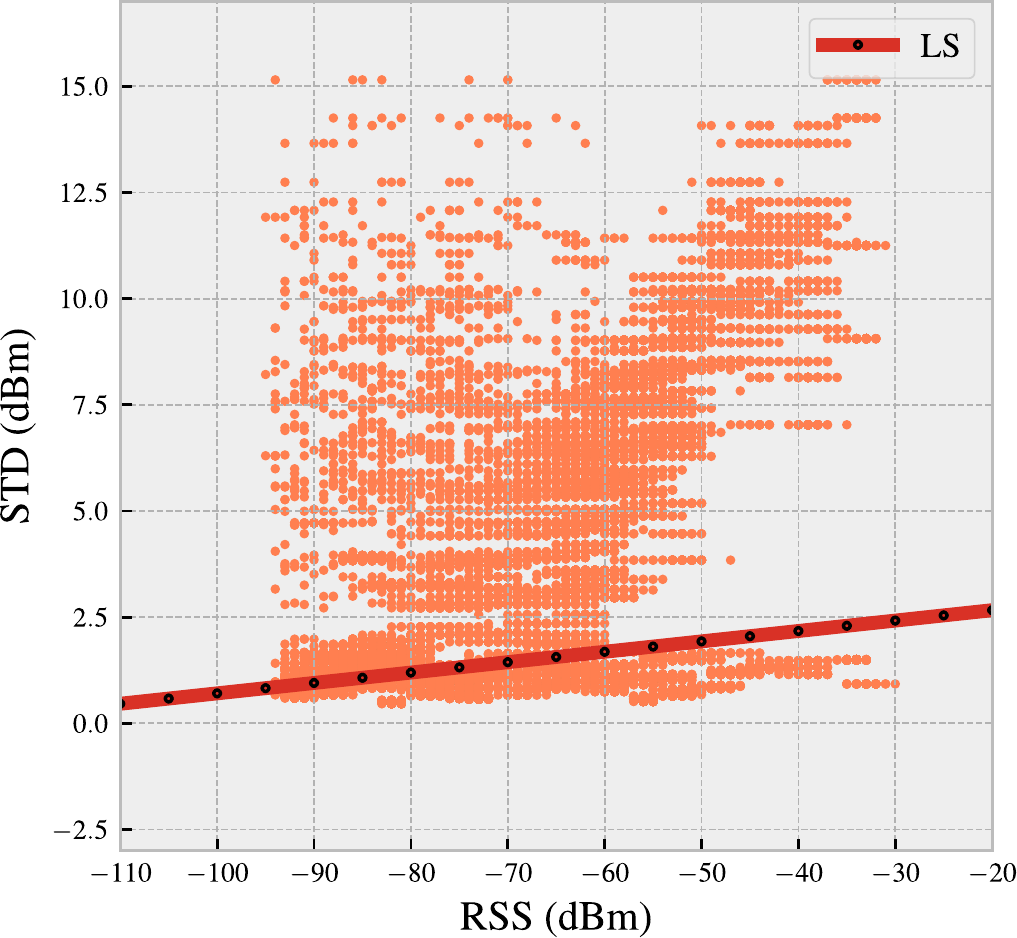}}
	\caption{The empirical estimated of the variability of the \acs{rss}}
	\label{fig:variability_ls}
\end{figure}

Before describing the change detection, we shortly present the way of empirically estimating the variability in case of using \acs{rss} as the feature value. In \citep{Sen:2013}, Sent \etal~have argued that the variability of the \acs{rss} is positively correlated to the signal strength. Inspired by their primary result, we have carried an independent collection of \acs{rss} at varying locations throughout the \acs{roi}. The mobile device is statically put on a table at the height about 1~m and around 100 samples are collected at each location. The aim of performing this collection is to find the model between the deviation of the \acs{rss} and the measured value. We thus first analyze the distribution of the \acs{std} (\figPref\ref{subfig:std_hist}) and it shows that the values of \acs{std} are mostly lower than 2.0~dBm and distribute in a long tail pattern. This pattern of the distribution helps us to interpret the linearly fitted model for estimating te \acs{std} of a given measured value (\figPref\ref{subfig:rss_std_ls}). The estimated model is slightly inclined and yields a slightly larger \acs{std} when giving a high value of \acs{rss} that of giving a lower \acs{rss} value. We use this fitted model as the empirical variability model of the feature value and construct the approximated Gaussian distribution used for change detection.

\begin{figure}[!h]
	\centering
	\subfloat[\acs{roc} curve]{\label{subfig:roc_auc}
		\includegraphics[width=0.35\linewidth]{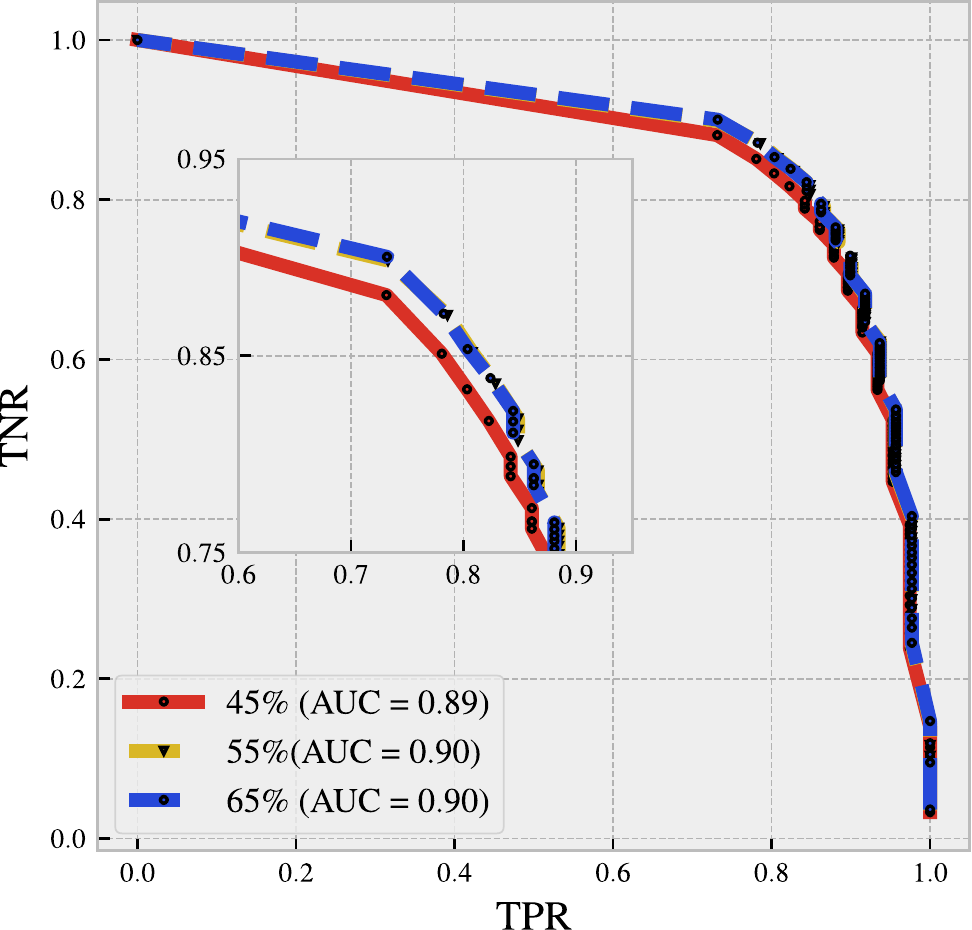}}\hspace{4ex}
	\subfloat[\acs{tpr} and \acs{fpr} curves]{\label{subfig:tnr_tpr}
		\includegraphics[width=0.35\linewidth]{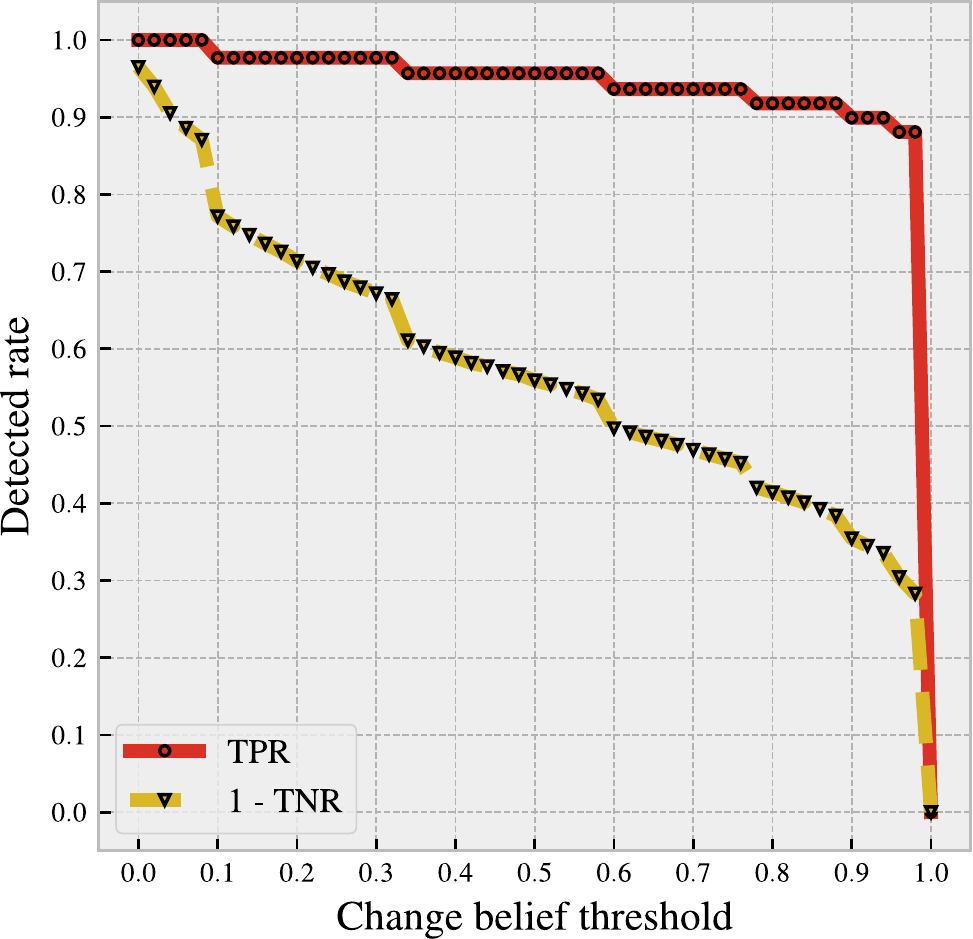}}
	\caption{The influence of change belief threshold on change detection performance of simulated changes (50\%, 0, -15~dBm).}
	\label{fig:roc_auc}
\end{figure}

\figPref\ref{subfig:roc_auc} illustrates the \acs{roc} curves for various values of the sampling ratio for all validation samples that contain the simulated changes (50\%, 0, -15~dBm). It indicates that their change detection performances are comparable when the sampling ratio is set to 55\% and 65\%, respectively. In \figPref\ref{subfig:tnr_tpr}, it illustrates the \acs{tpr} and \acs{fpr} (\ie 1 - \acs{tnr}) curve \acs{wrt} different change belief threshold. The \acs{tpr} remains high when increasing the change belief threshold but the reduction of \acs{fpr} is slow. I.e. it has the higher chance that falsely identifying a stable feature as changed than that of  incorrectly classifying a changed feature as stable if the change belief threshold is set close to 0.5. This tendency does not harm too much to the accumulative scheme of updating the \acs{rfm} because the change status of the features is collected over a period and is reported by multi-users/devices. But it might have the negative impact on the online positioning phase in case of dropping out the features that detected as changed when carrying out fingerprint matching. One hypothesis for explaining this trend is that the change belief approximation assumption does not hold. It suggests that the variability of the \acs{rss} is underestimated. In addition, the tendency suggests that the change belief threshold should be set more close to 1 than to 0.5. 

\begin{figure}[!h]
	\centering
	\subfloat[-15 dBm]{\label{subfig:med_auc_sr_45_m_15}
		\includegraphics[width=0.23\linewidth]{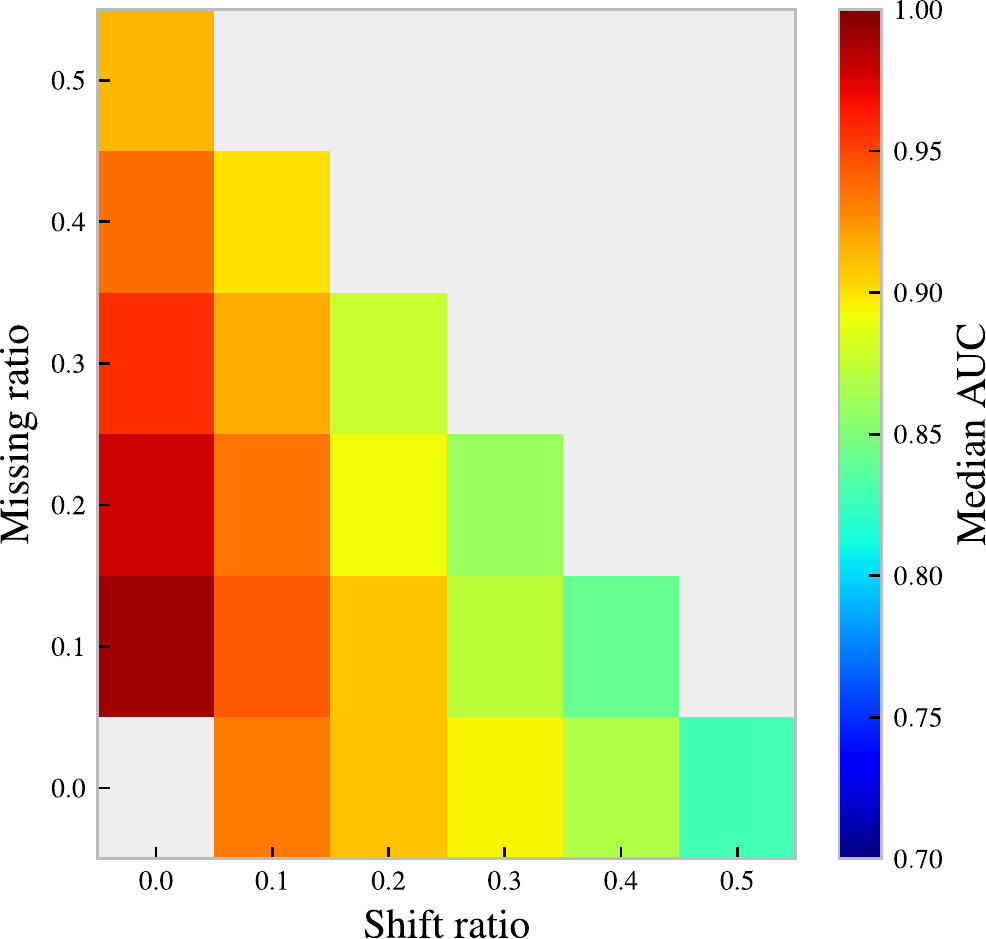}}\hspace{1ex}
	\subfloat[-10 dBm]{\label{subfig:med_auc_sr_45_m_10}
		\includegraphics[width=0.23\linewidth]{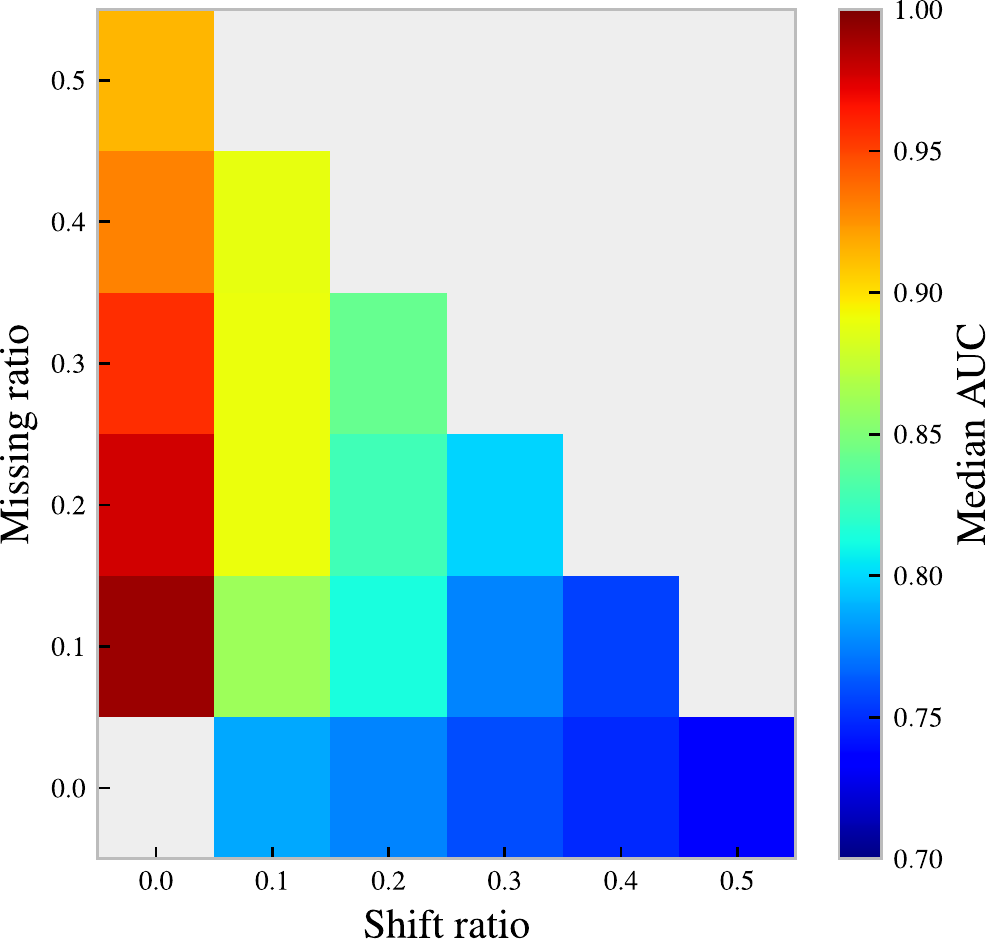}}\hspace{1ex}
	\subfloat[10 dBm]{\label{subfig:med_auc_sr_45_p_10}
		\includegraphics[width=0.23\linewidth]{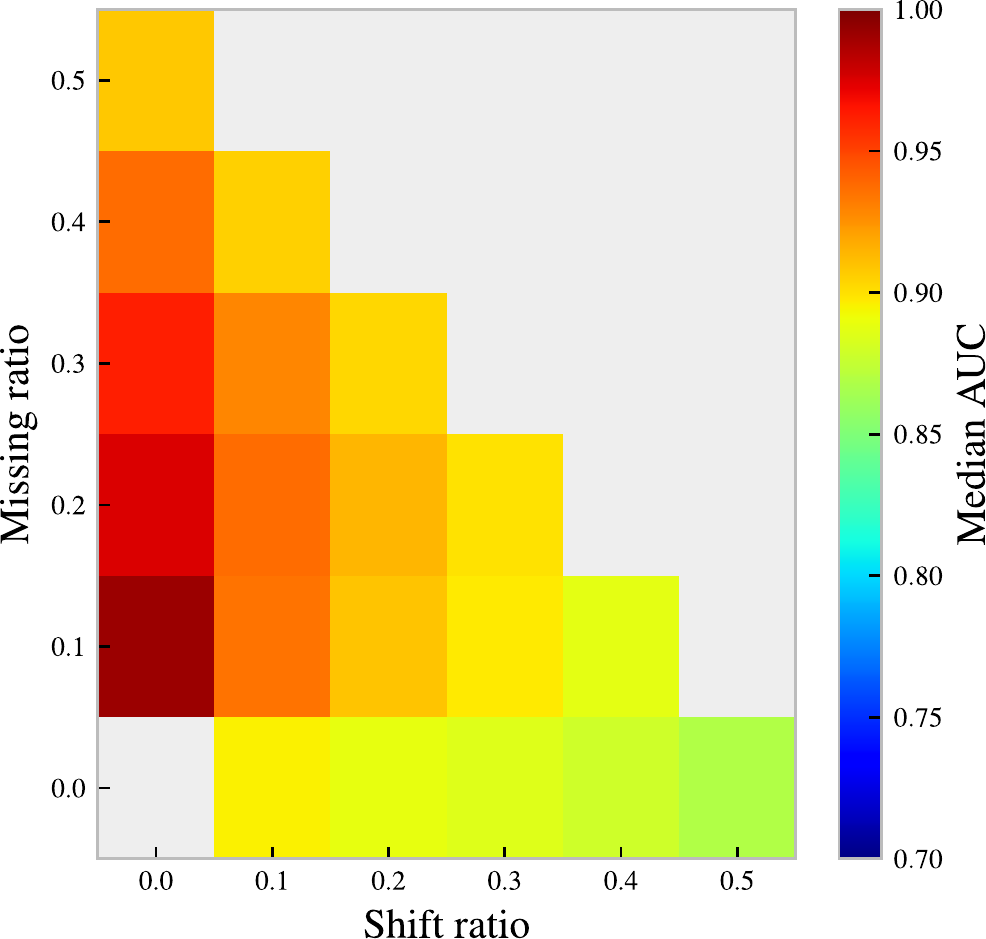}}\hspace{1ex}
	\subfloat[15 dBm]{\label{subfig:med_auc_sr_45_p_15}
		\includegraphics[width=0.23\linewidth]{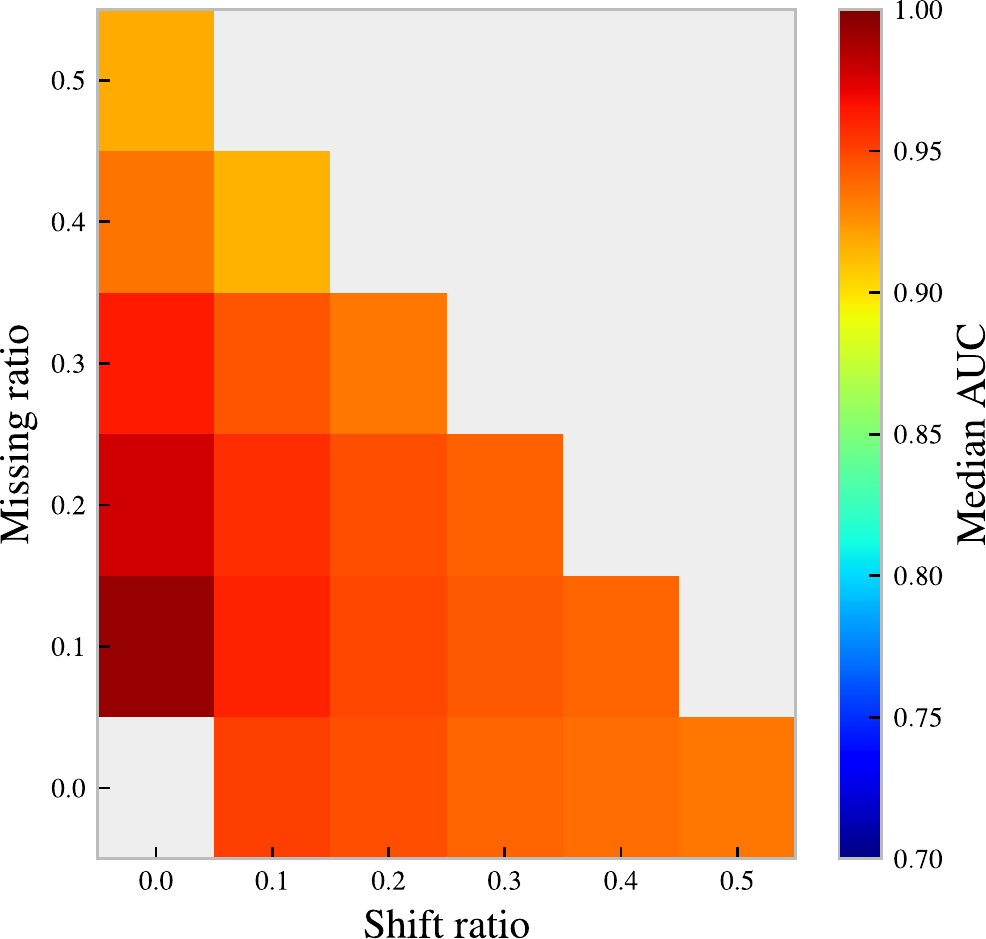}}\\
	\subfloat[-15 dBm]{\label{subfig:med_auc_sr_55_m_15}
		\includegraphics[width=0.23\linewidth]{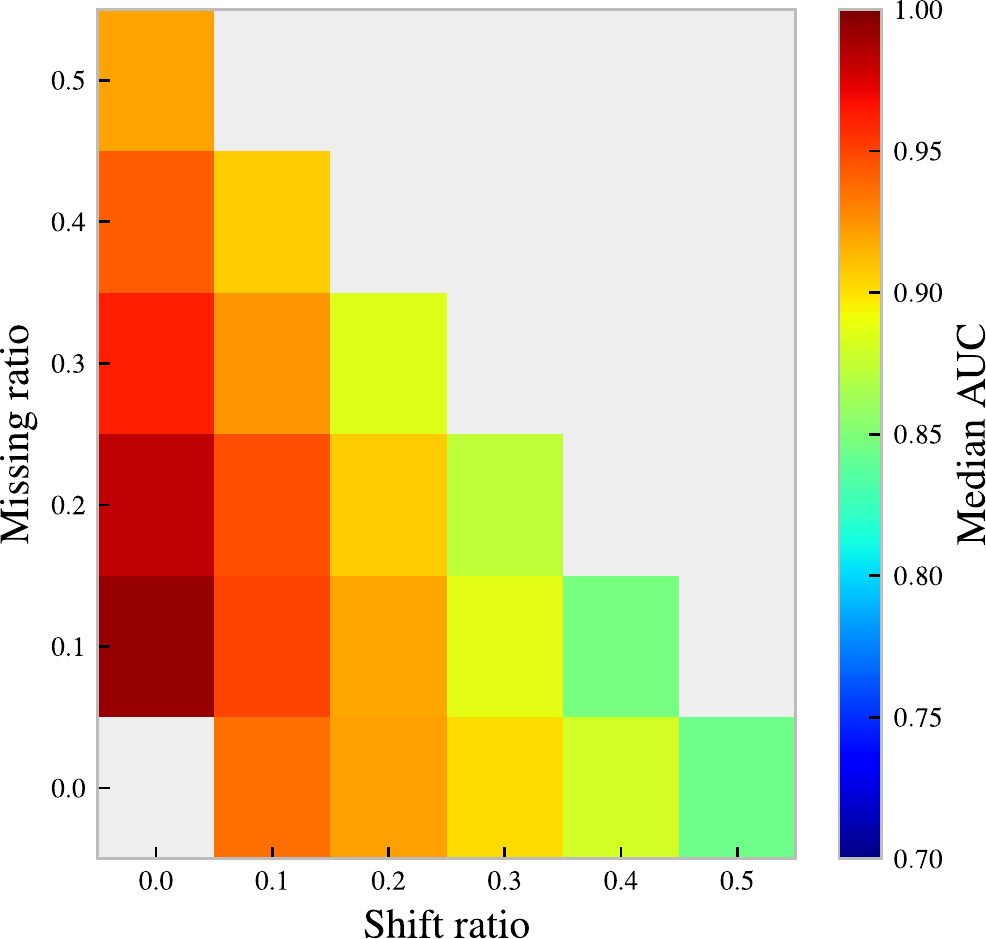}}\hspace{1ex}
	\subfloat[-10 dBm]{\label{subfig:med_auc_sr_55_m_10}
		\includegraphics[width=0.23\linewidth]{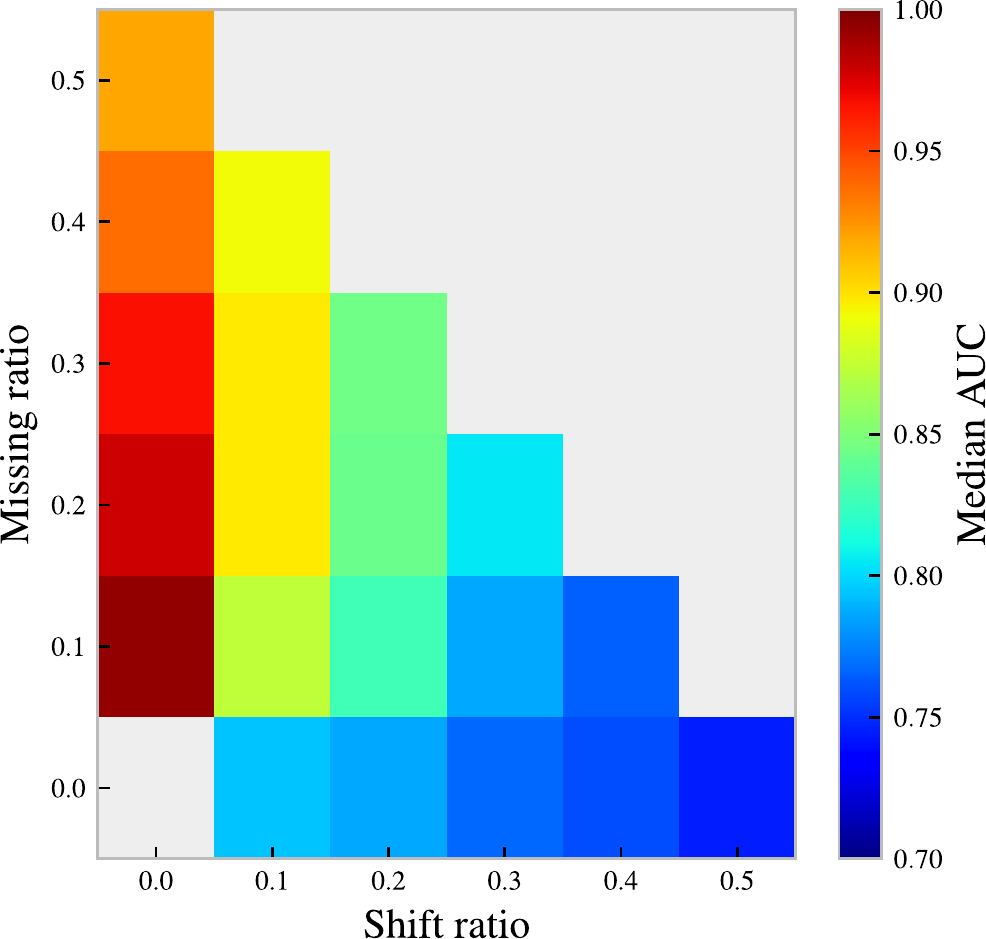}}\hspace{1ex}
	\subfloat[10 dBm]{\label{subfig:med_auc_sr_55_p_10}
		\includegraphics[width=0.23\linewidth]{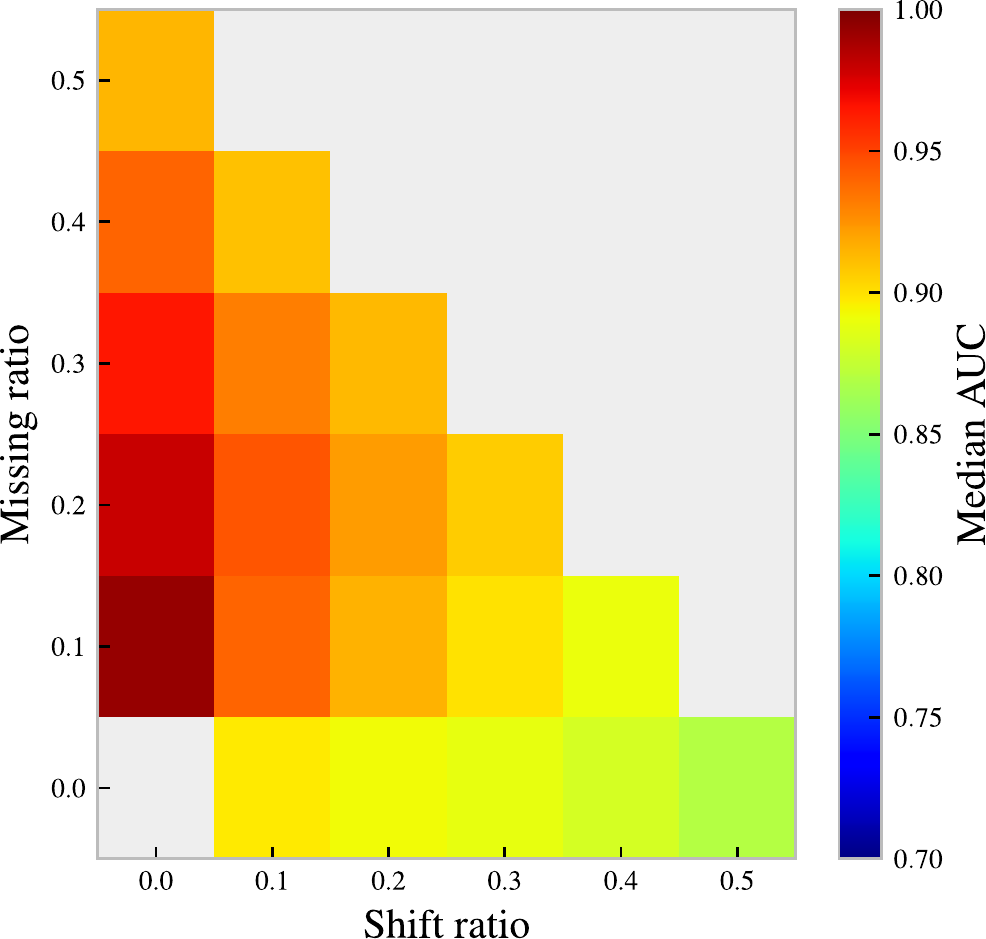}}\hspace{1ex}
	\subfloat[15 dBm]{\label{subfig:med_auc_sr_55_p_15}
		\includegraphics[width=0.23\linewidth]{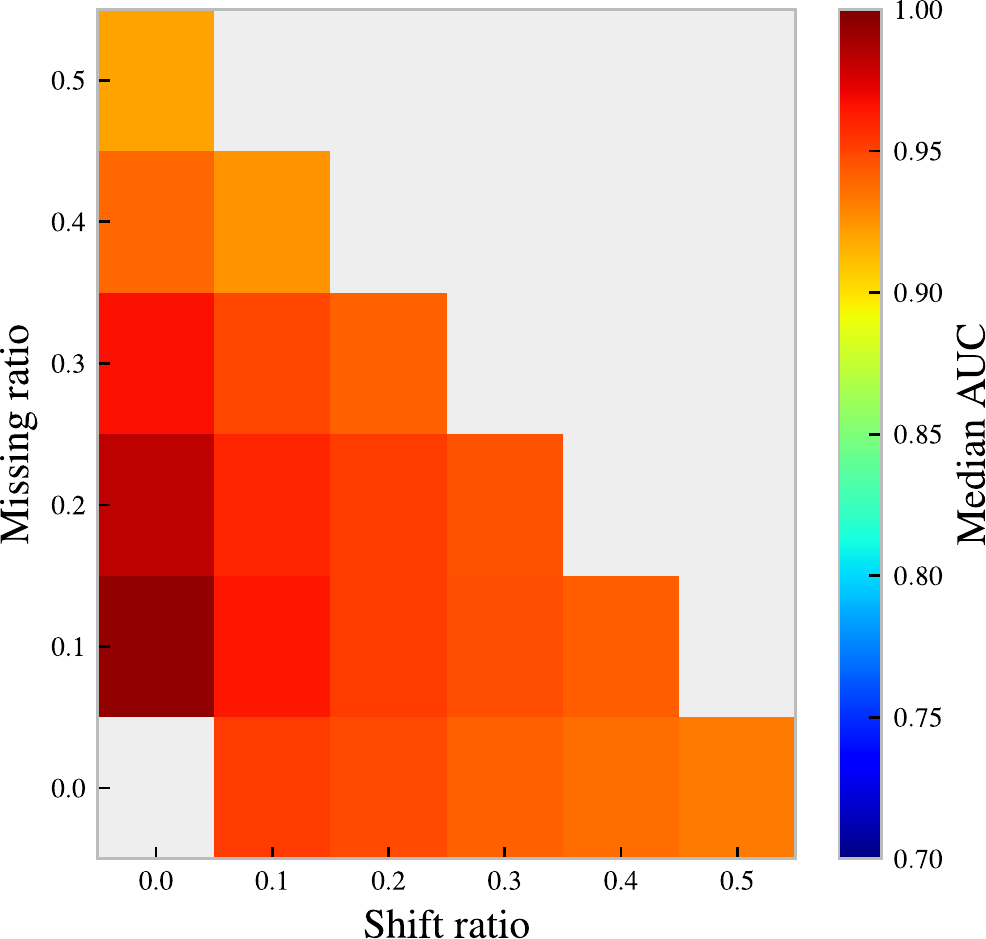}}\\
	\subfloat[-15 dBm]{\label{subfig:med_auc_sr_65_m_15}
		\includegraphics[width=0.23\linewidth]{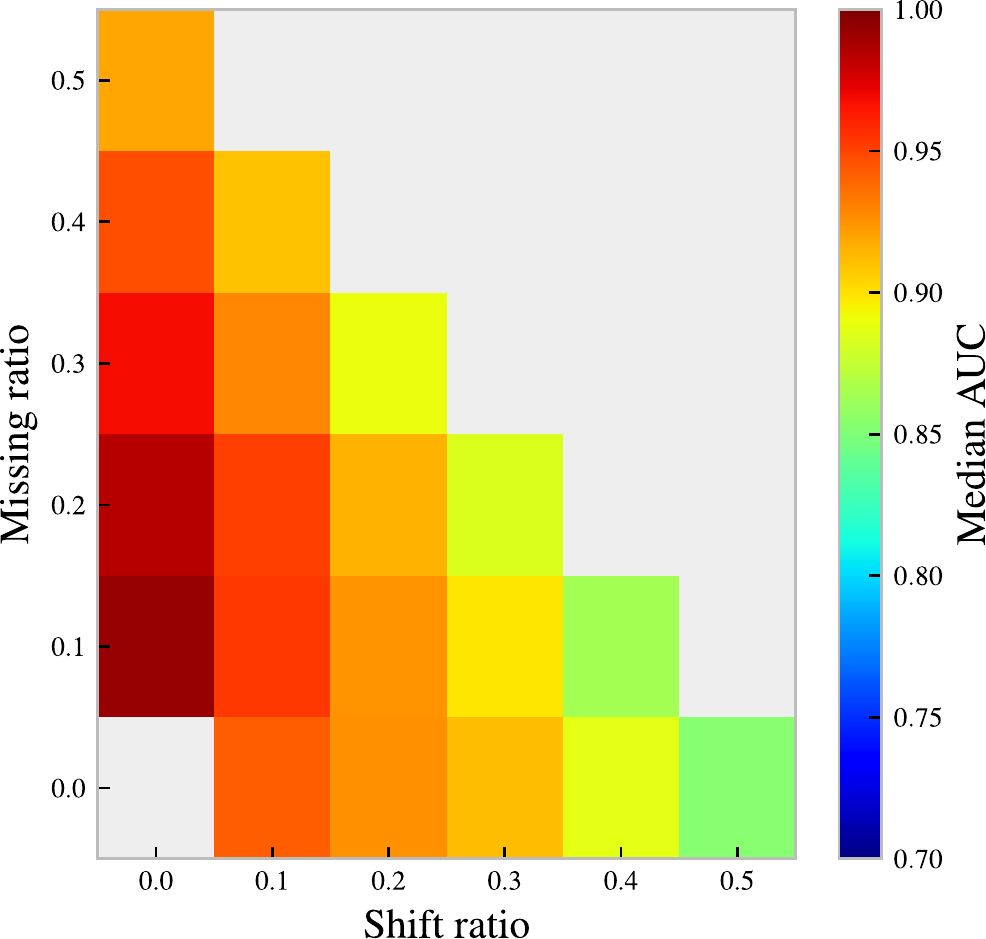}}\hspace{1ex}
	\subfloat[-10 dBm]{\label{subfig:med_auc_sr_65_m_10}
		\includegraphics[width=0.23\linewidth]{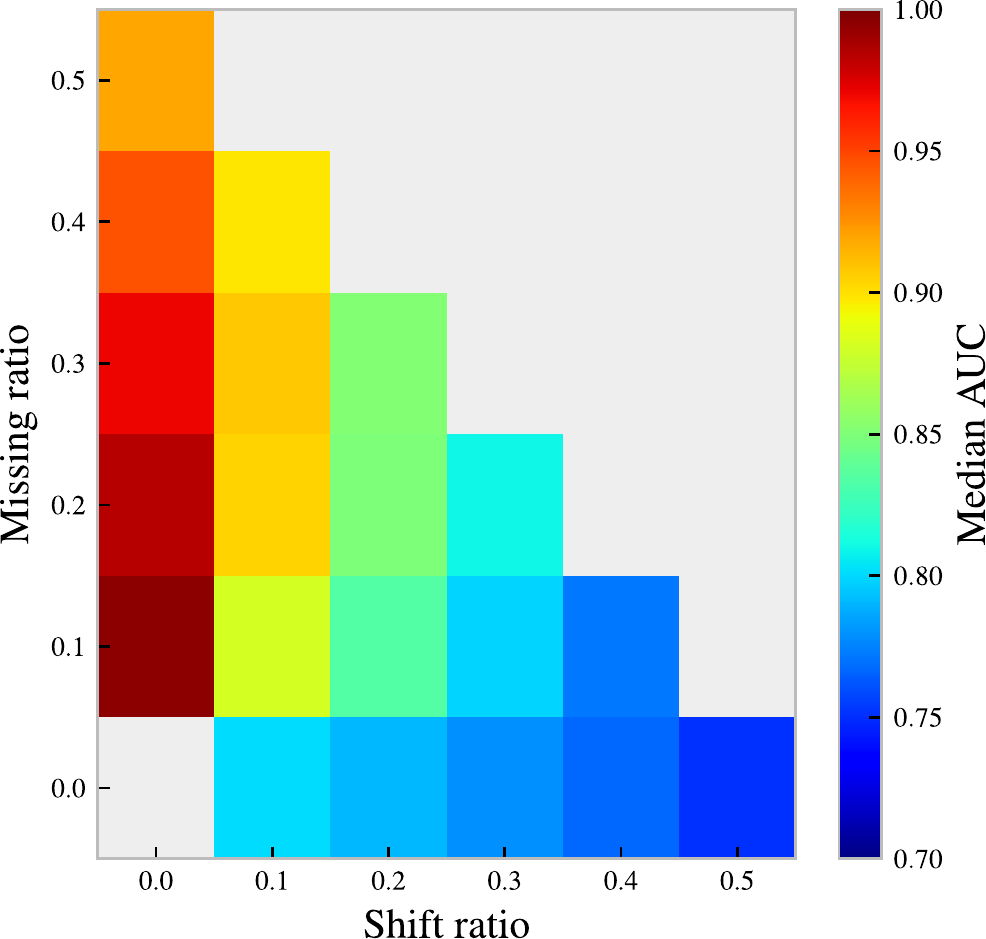}}\hspace{1ex}
	\subfloat[10 dBm]{\label{subfig:med_auc_sr_65_p_10}
		\includegraphics[width=0.23\linewidth]{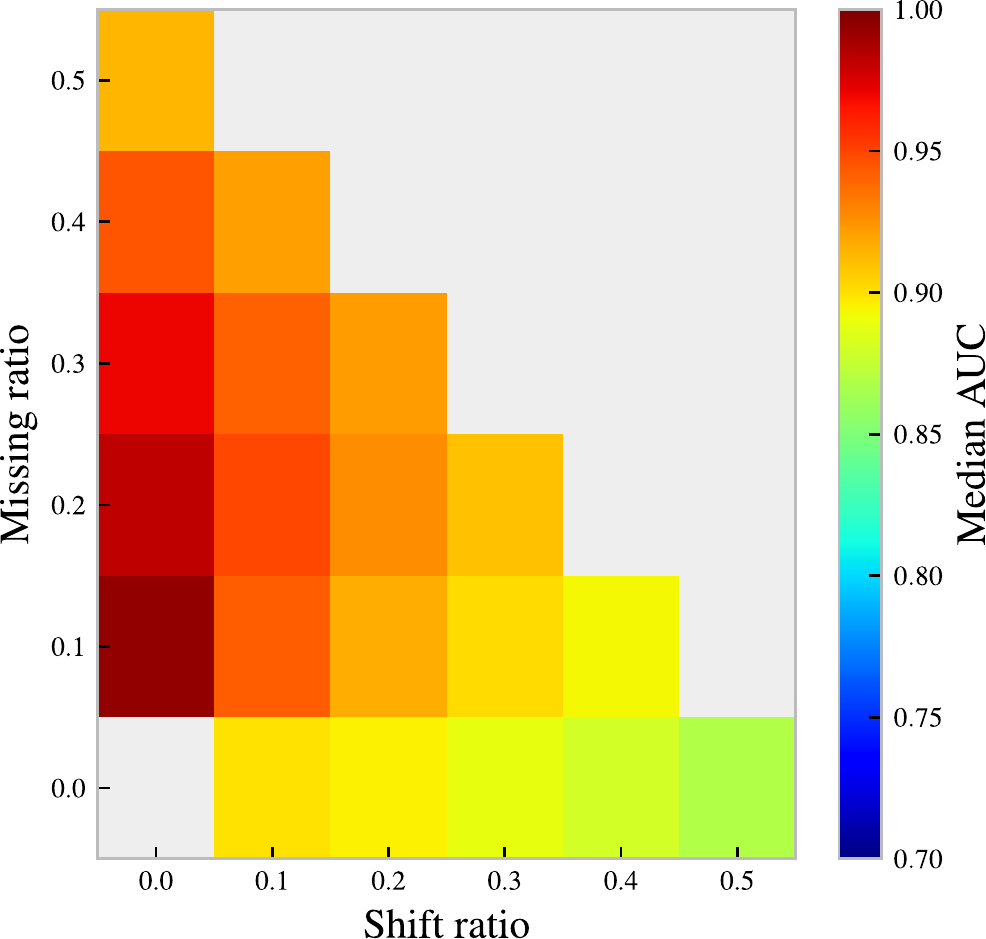}}\hspace{1ex}
	\subfloat[15 dBm]{\label{subfig:med_auc_sr_65_p_15}
		\includegraphics[width=0.23\linewidth]{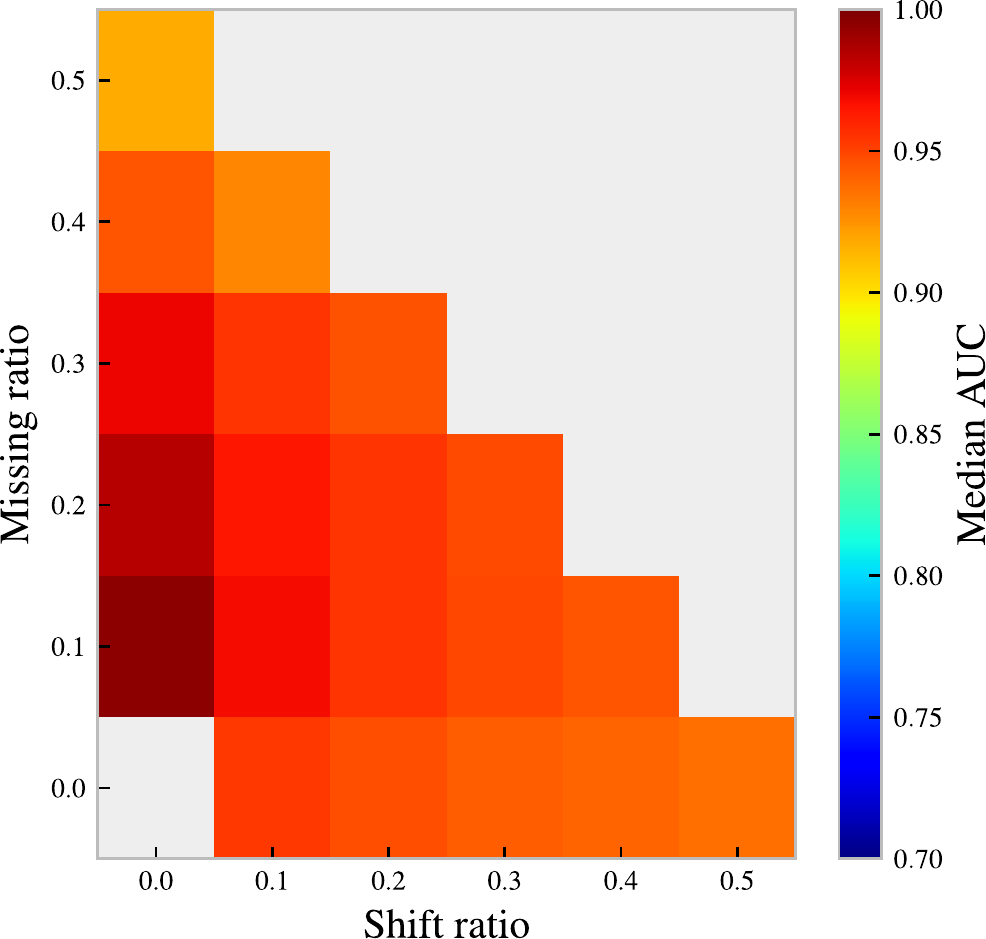}}
	\caption{The median value of \acs{auc} of the simulated changes. For each row, it presents the results of setting the sampling ratio to 45\%, 55\% and 65\%, respectively.}
	\label{fig:median_auc}
\end{figure}

We present the results for distinguishing different combinations of simulated changes in \figPref\ref{fig:median_auc}. On average, the \acs{auc} value is close to 0.9, \ie the achievable change detection accuracy is about 90\%. We can conclude from the results that: i) concerning the missing ratio or shift ratio, the \acs{auc} value reduces as the increasing of the missing or shift ratio. The larger amount of changes occurred in the measurable features makes it more challenging to identify. This is related to the degrading of positioning performance when an increasing amount of changes has occurred (see \figPref\ref{fig:cmp_ecdf_missing}). In case that the deteriorating of the positioning performance is too much, the contribution of the deviation of feature values from the positioning error dominates among the residuals; ii) the value shift of the features has a higher impact on the change detection than that of missing features. The reduction of \acs{auc} value when increasing the shift ratio is larger than that of increasing the missing ratio. This is because of two factors: one is the dissimilarity measure. \acs{cdm} is robust to missing features but ignores to alleviate the influence of value changes of the features. Another is that the missing features are easier to detect because those missed features would never be sampled and it is highly likely to yield large residual when comparing to the supposed values that retrieved from the \acs{rfm}; iii) the mixed changes of missing features and the value shift have an inconsistent influence on the change detection. For a given missing ratio of features, the \acs{auc} value slightly improves as increasing the shift ratio; iv) referring to the impact of the shift ratio on the change detection, reducing the feature value makes the change more difficult to be detected than that of increasing the feature value. One explanation is that the reduction of the feature value has the possibility that makes the feature from measurable to be missed. I.e. it increases the missing ratio when reducing the feature value. However, the pattern of the impact of feature value shift is not coherent. The reduction of 10~dBm of the feature value has a higher impact than that of reducing the feature value about 15~dBm. Our hypothesis is that a smaller reduction of feature value is more difficult to detect because it results in a small value of \acs{rss} (but not missing) which mitigates the discriminativeness of corresponding features; and v) the sampling ratio has limited impact on the change detection, though the performance is slightly improved as the sampling ratio increase from 45\% to 65\%.

\begin{figure}[!h]
	\centering
	\includegraphics[width=0.4\linewidth]{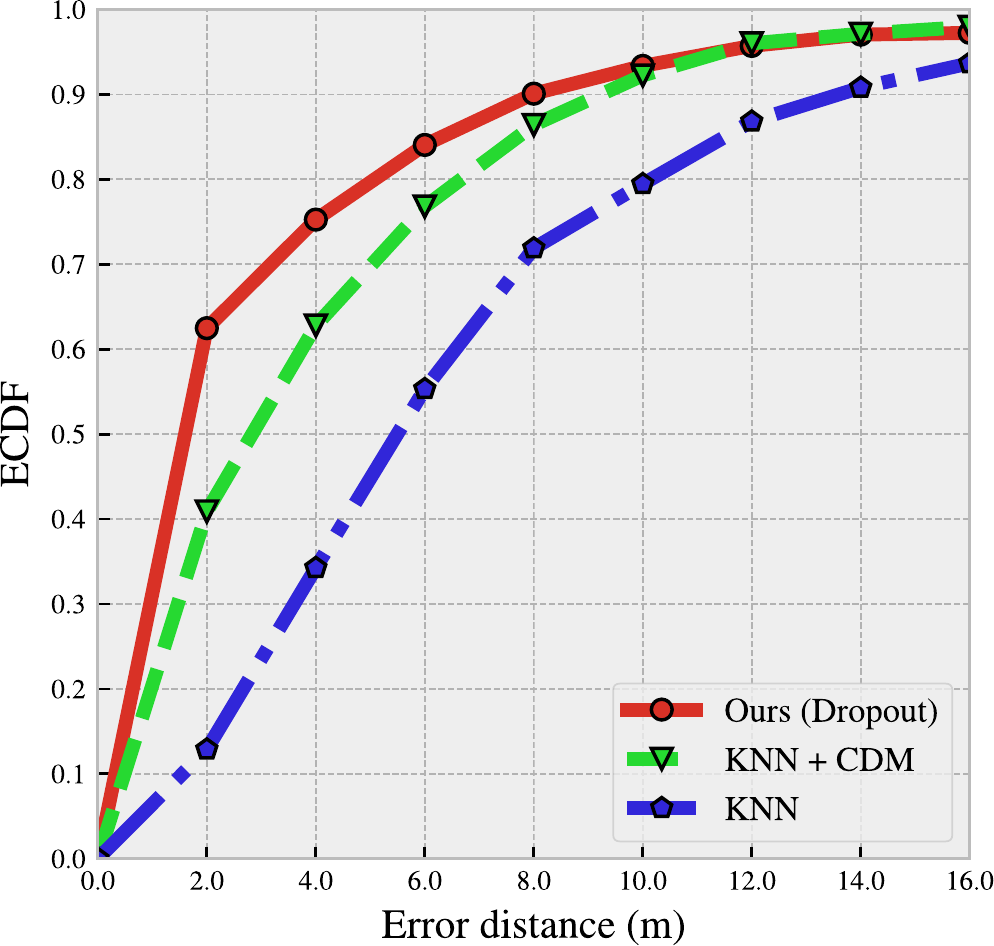}
	\caption{The comparison of \acs{ecdf} when dropping out the features that are detected as changed in case that the simulated changes is (50\%, 0, -15~dBm). The change belief threshold is set to 0.95.}
	\label{fig:w_o_ecdf}
\end{figure}

After identifying which subset of features has changed, we drop them out when performing the position estimation. \figPref\ref{fig:w_o_ecdf} compares the \acs{ecdf} \acs{wrt} to the positioning error between the baselines and the one obtained without taking the changed features into account. The positioning accuracy within 2~m and 4~m are up-to 62\% and 75\%, which improves about 20\% and over 10\% comparing to that of \acs{knn} with \acs{cdm}, respectively.

\begin{figure}[!h]
	\centering
	\includegraphics[width=0.4\linewidth]{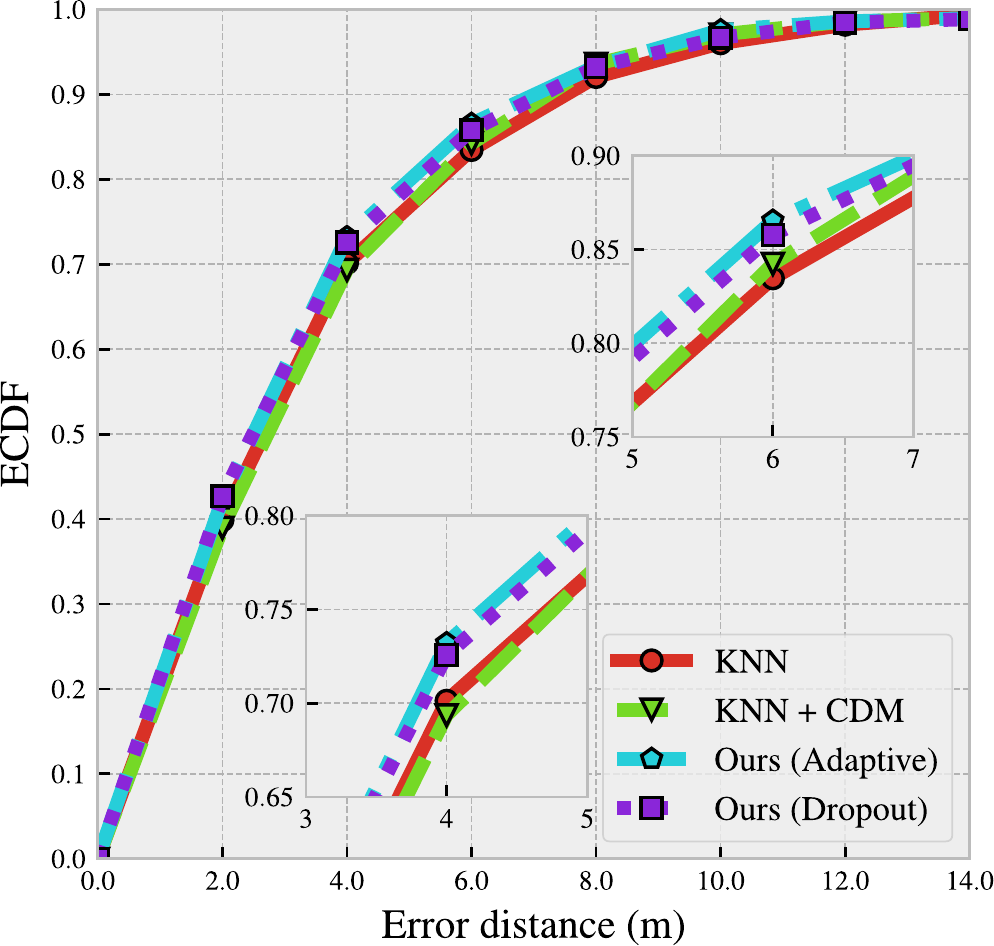}
	\caption{Comparison of \acs{ecdf} of dataset \textit{HIL-R2}}
	\label{fig:ecdf_r2}
\end{figure}

In addition, we also extend the experimental analysis to dataset \textit{HIL-R2}. We set the values of parameters the same as used for dataset \textit{HIL-R1}. For the generalization test, the sampling ratio is arbitrarily set to 45\%. \figPref\ref{fig:ecdf_r2} depicts the positioning performance using the \acs{ecdf} \acs{wrt} the error distance. Without any fine-tuning of the parameters for \textit{HIL-R2}, we can slightly improve the positioning performance. The positioning accuracy within 4~m increases by around 4\%. We would expect that the positioning performance can be improved if the parameters were tuned for this dataset.

\subsection{Generalization to long-term dataset}
Mendoza-Silva has contributed a long-term collected WiFi fingerprinting dataset and it is a benchmark database for studying robust indoor positioning \citep{mendoza2018long}. We use this dataset for validating the generalization capability of our proposal. We first describe the analysis of the baseline positioning performance and the change status labeling, respectively. We then evaluate the change detection performance by comparing the detected feature status to that of labeled by location-wisely analyzing the variability of the feature values.

\subsubsection{About the UJI long-term dataset}
\begin{figure}[!h]
	\centering
	\subfloat[Raw \acs{rfm}]{\label{subfig:uji_raw_rfm}
		\includegraphics[width=0.3\linewidth]{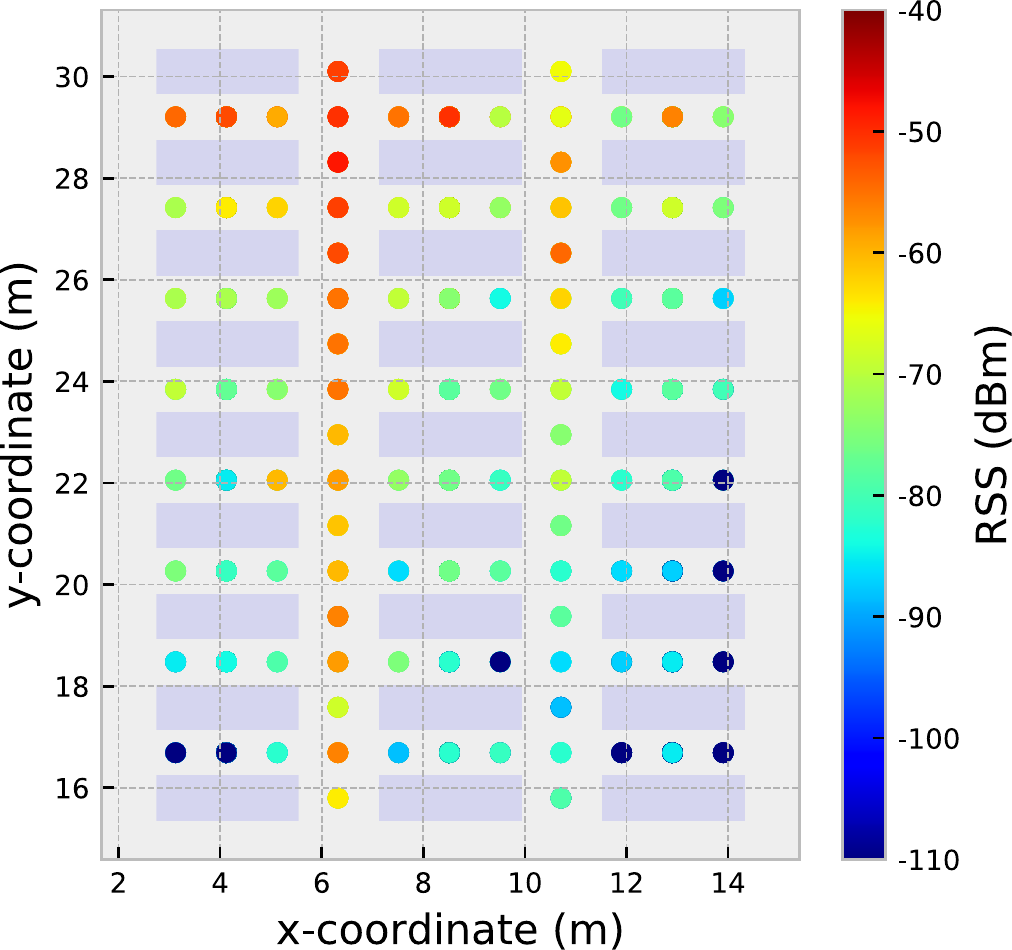}}\hspace{2ex}
	\subfloat[Smoothed \acs{rfm}]{\label{subfig:uji_ks_rfm}
		\includegraphics[width=0.3\linewidth]{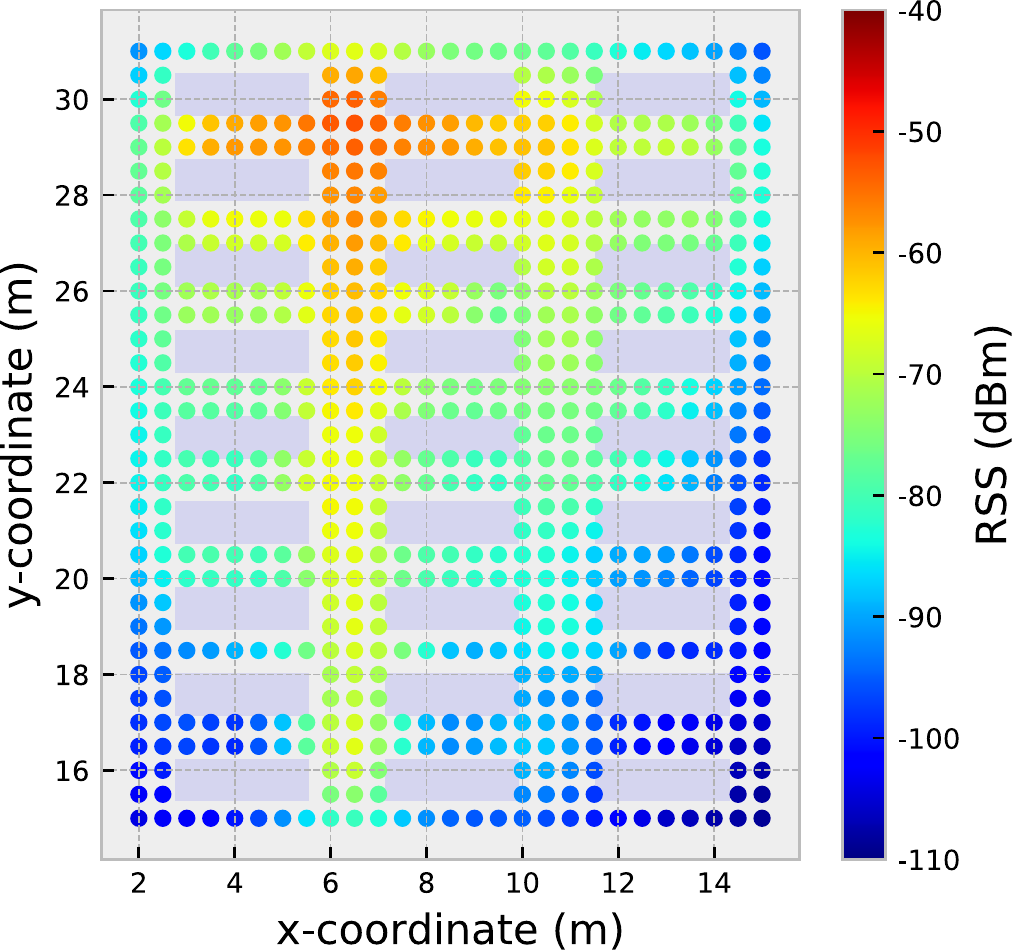}}\hspace{2ex}
	\subfloat[\acs{ecdf}]{\label{subfig:uji_ecdf_cmp}
		\includegraphics[width=0.3\linewidth]{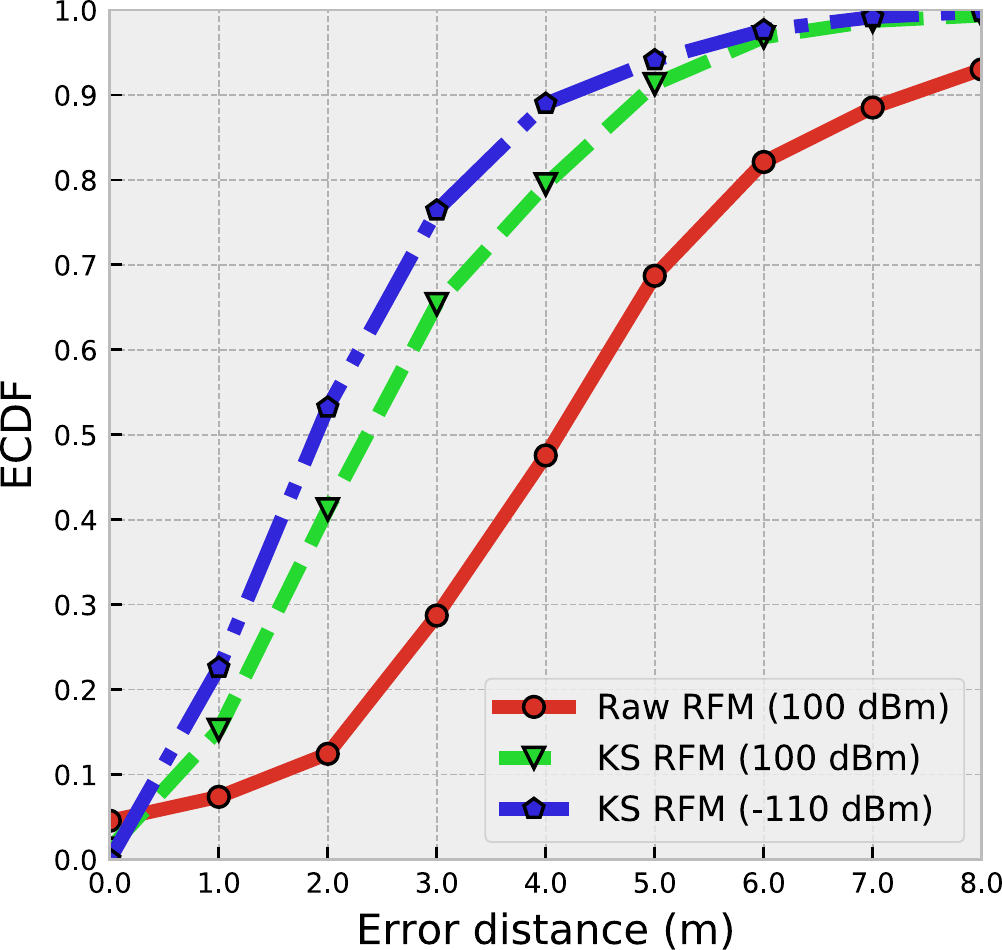}}
	\caption{The results of \acs{ks} interpolated \acs{rfm} and baseline positioning. The light violet rectangles in the first two plots denotes the bookshelves, which are inaccessible.}
	\label{fig_uji_rss_ecdf}
\end{figure}

We briefly describe the dataset. More details can be found in \citep{mendoza2018long}. This dataset is collected in a library over a period of 15 months. Each round of the data collection, a surveyor uses the mobile device to measure the \acs{rss} from the measurable \acsp{ap} and collect 6/12 consecutive samples at each location facing to a given direction statically. In order to evaluate the generalization performance, we use the dataset in the following way:
\begin{itemize}
	\item \acs{roi}: we evaluate the approach using the dataset that covers the 3rd floor without mitigating generosity;
	\item \acs{rfm}: we use the data collected during the first month for finding the variability of the feature values and for building the world model of the \acs{rfm} using \acs{ks}. Regarding the variability, we carry out the analysis as described in \secPref\ref{subsubsec:iden_feat} and estimate the variability model used for change belief approximation. Using the raw measured data, we generate the interpolated (0.5-by-0.5~$ \mathrm{m^2} $ grid) and smoothed \acs{rfm} by applying \acs{ks} ($ \norSymScript{\lambda}{}{LS}=2,\, \norSymScript{\lambda}{}{KS}=2$) (see \figPref\ref{subfig:uji_raw_rfm} and \figPref\ref{subfig:uji_ks_rfm});
	\item Missing indicator value: In the original dataset, the missed features are indicated using 100. We replace it with -110~dBm when performing positioning (see \figPref\ref{subfig:uji_ecdf_cmp}).
	\item The settings of parameters: For the values of other parameters, we fixed them the same as used in the previous subsection.
\end{itemize}

\subsubsection{Labeling the change status}
\begin{figure}[!h]
	\centering
	\subfloat[\acs{rss} range, \acs{ap} 1]{\label{subfig:rss_var_boxplot_26_1}
		\includegraphics[width=0.45\linewidth]{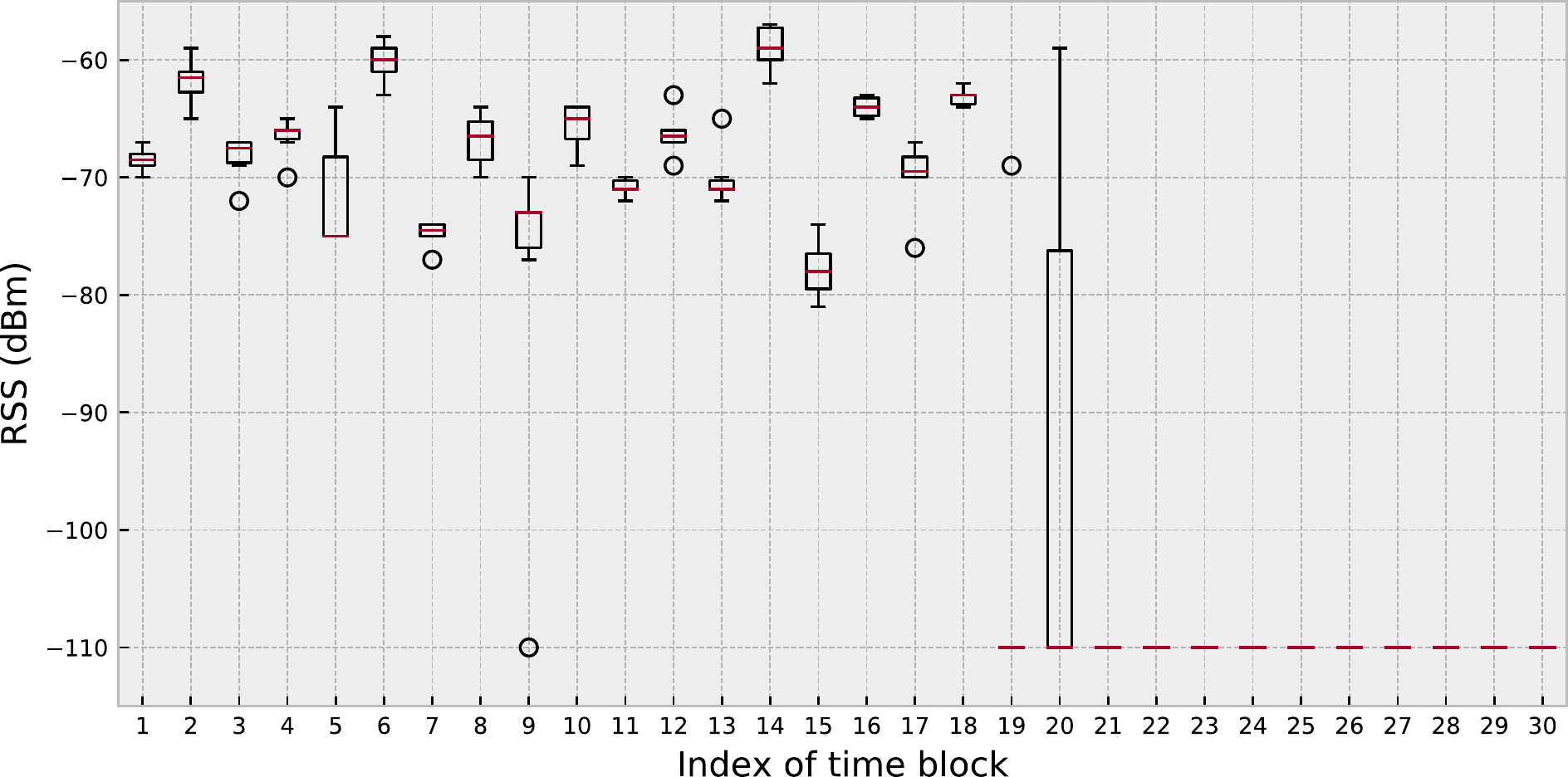}}\hspace{1ex}
	\subfloat[\acs{rss} range, \acs{ap} 8]{\label{subfig:rss_var_boxplot_26_8}
		\includegraphics[width=0.45\linewidth]{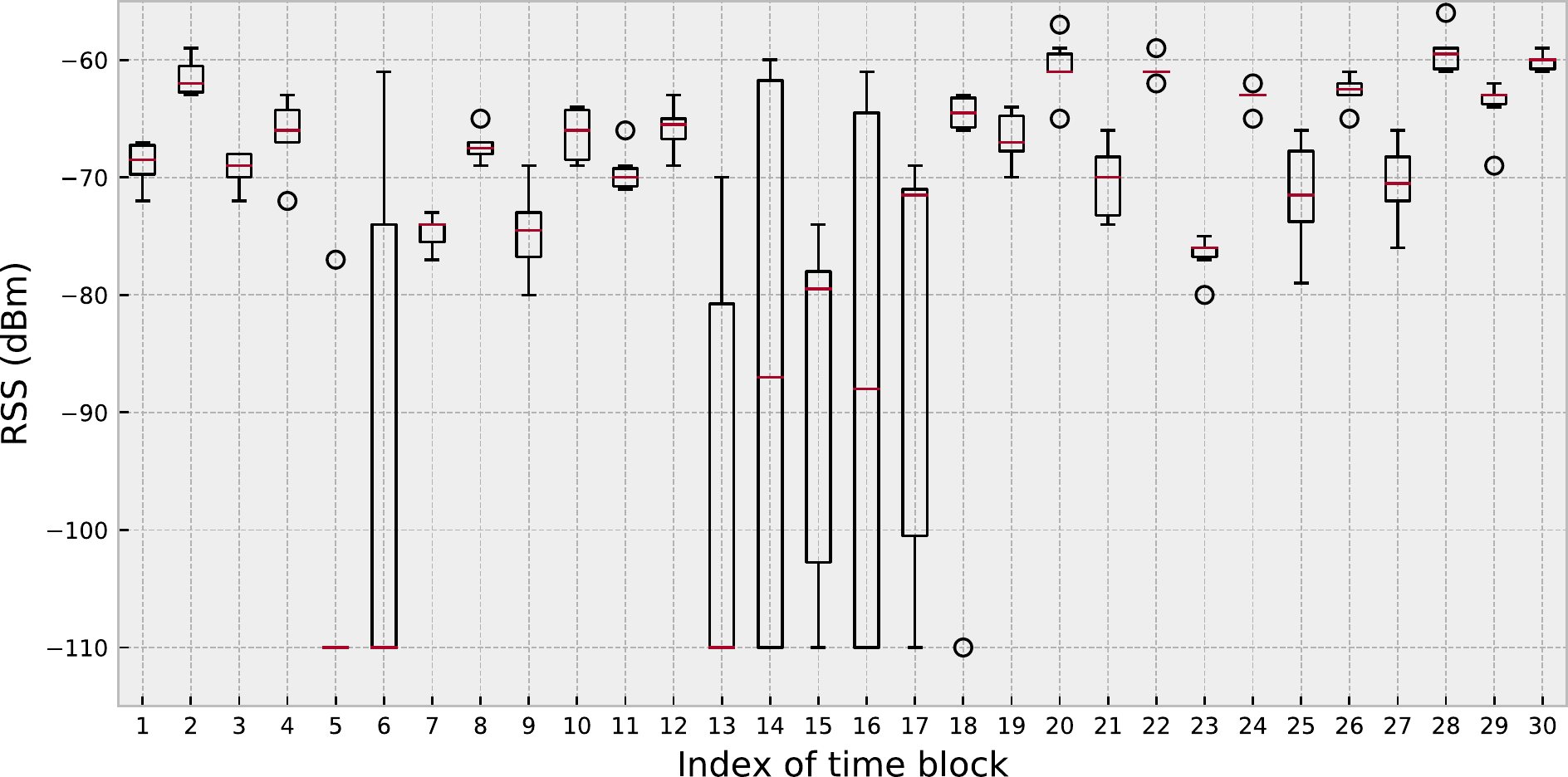}}\\
	\subfloat[Estimated variability, \acs{ap} 1]{\label{subfig:rss_var_rob_26_1}
		\includegraphics[width=0.45\linewidth]{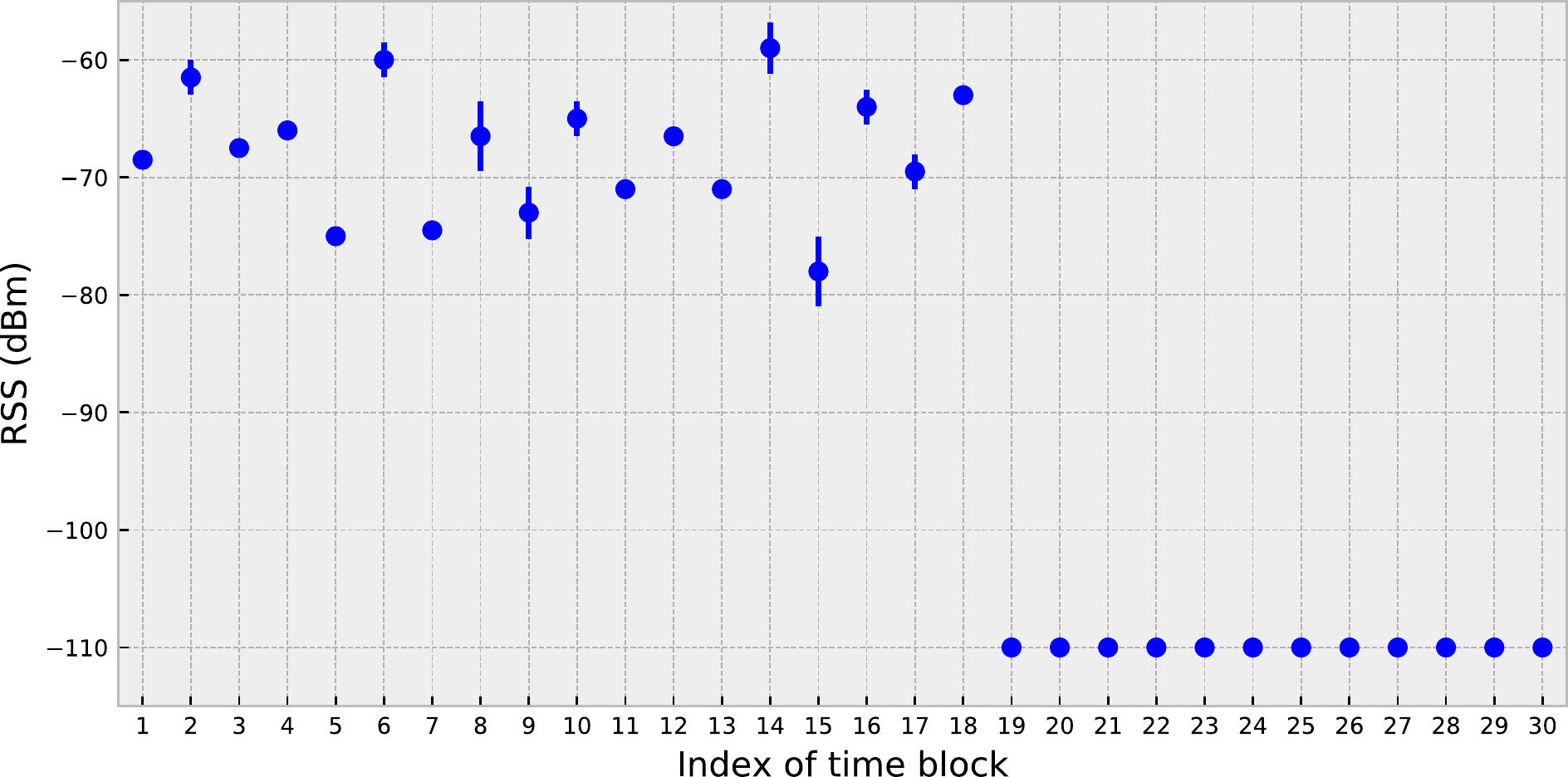}}\hspace{1ex}
	\subfloat[Estimated variability, \acs{ap} 8]{\label{subfig:rss_var_rob_26_8}
		\includegraphics[width=0.45\linewidth]{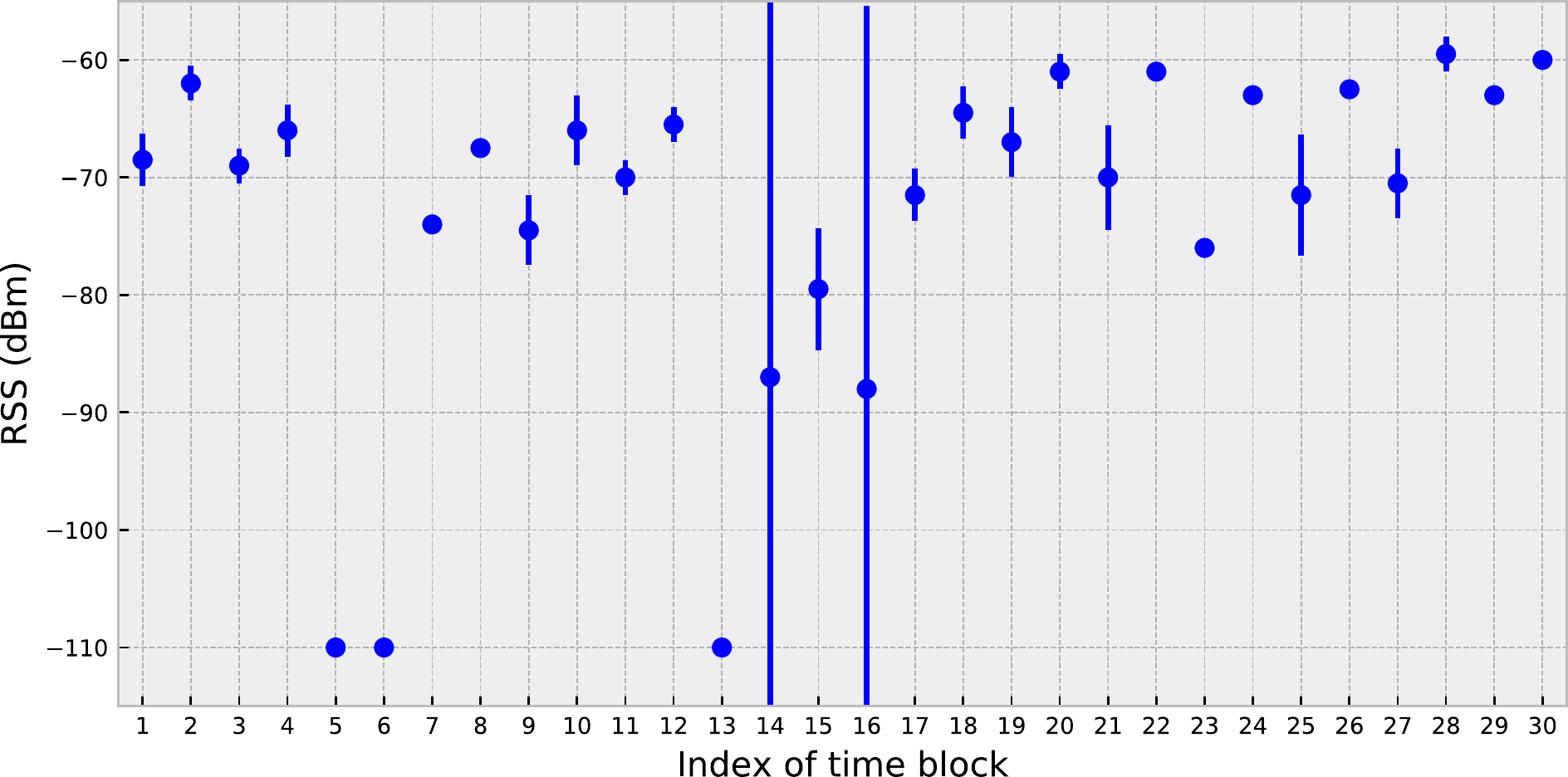}}\\
	\subfloat[Labeled status, \acs{ap} 1]{\label{subfig:rss_var_status_26_1}
		\includegraphics[width=0.45\linewidth]{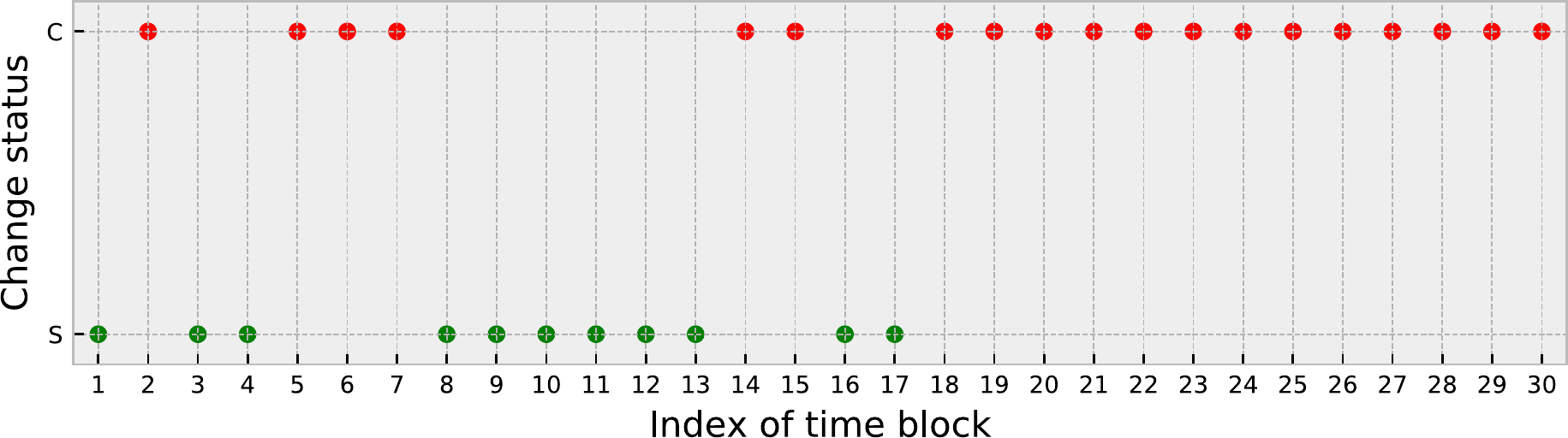}}\hspace{1ex}
	\subfloat[Labeled status, \acs{ap} 8]{\label{subfig:rss_var_status_26_8}
		\includegraphics[width=0.45\linewidth]{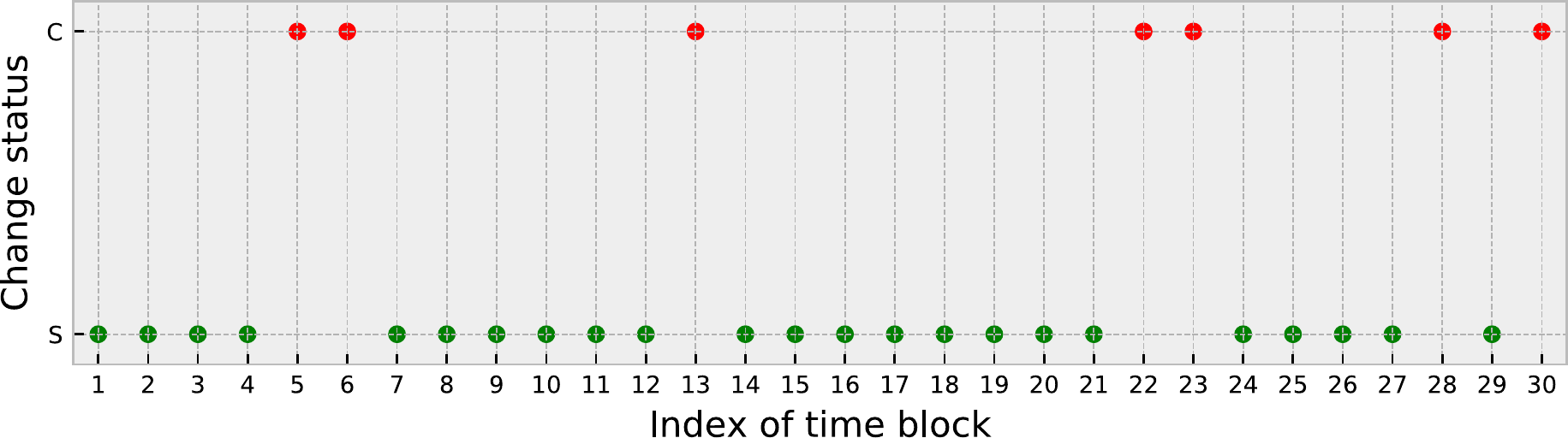}}\\
	\subfloat[Estimated status, \acs{ap} 1]{\label{subfig:rss_est_status_26_1}
		\includegraphics[width=0.45\linewidth]{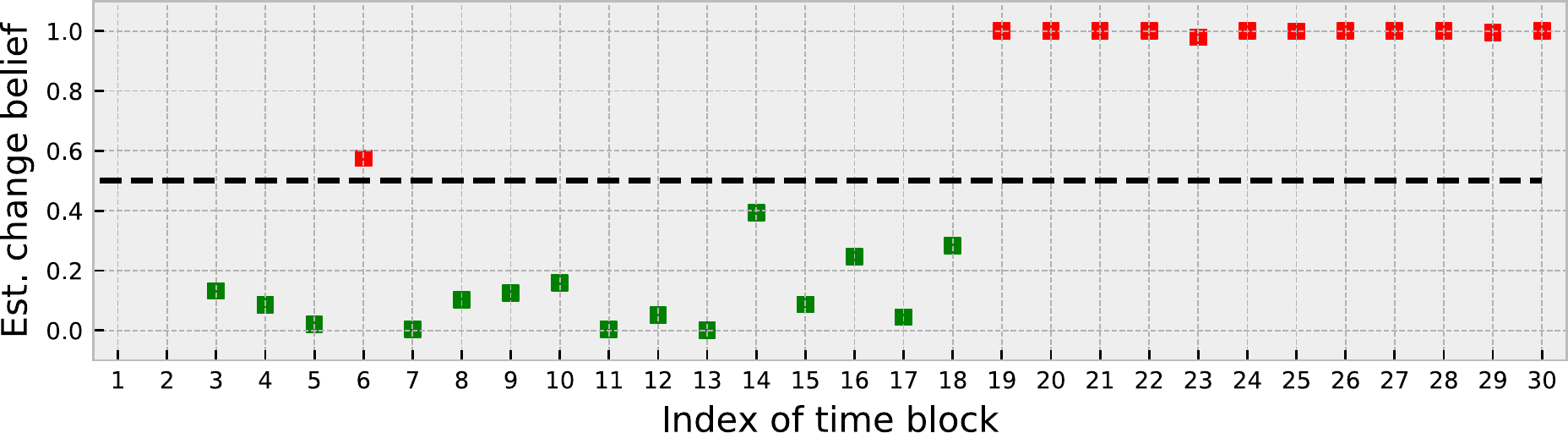}}\hspace{1ex}
	\subfloat[Estimated status, \acs{ap} 8]{\label{subfig:rss_est_status_26_8}
		\includegraphics[width=0.45\linewidth]{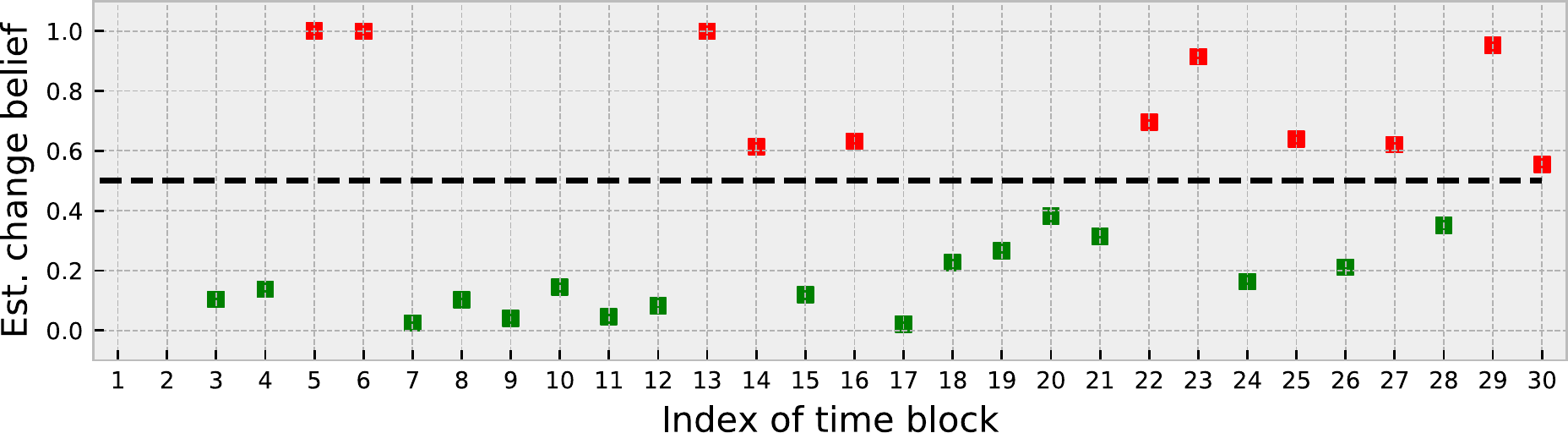}}
	\caption{Examples of change status labeling by comparing to the first time block (\ie index 1).  The labeled status is denoted as changed ('C', in red) or stable ('S', in green). The data collected in the first two time blocks (\ie the first month) is used for generating the smoothed and interpolated \acs{rfm}. The estimated status is marked in green (\ie stable)/red (\ie changed) by setting the detecting threshold to 0.5.}
	\label{fig:ch_labeling}
\end{figure}

Though we do not the ground truth of the status of the features we can approximately label the status by making full use of consecutively measured examples in the short time (\eg within one minute) at each location. We analyze the short time variability of the features using the available measurements at each location during the time block around one minute. At a fixed location, the consecutively collected values $ \vecSymScript{v}{i}{} $ of feature $ a $ of time block $ i $ is used to estimated the mean $ \norSymScript{\hat{\mu}}{i}{} $ and deviation $ \norSymScript{\hat{{\sigma}}}{i}{} $. In order to alleviate the impact of the noise in the raw measurement (see \figPref\ref{subfig:rss_var_boxplot_26_1} and \figPref\ref{subfig:rss_var_boxplot_26_8}), we estimated them using:
\begin{equation}
	\label{eq:rob_mean_std}
	\begin{split}
		&\norSymScript{\hat{\mu}}{i}{} = \operatorname{median}(\vecSymScript{v}{i}{})\\
		&\norSymScript{\hat{{\sigma}}}{i}{} = 1.4826\cdot\operatorname{MAD}(\vecSymScript{v}{i}{})
	\end{split}
\end{equation}
where $ \operatorname{MAD}(\cdot) $ is an operator for computing the \acf{mad}. We label the change status of each feature at two levels:
\begin{itemize}
	\item Within-time block: Any feature value deviates more than three times for the estimated standard deviation in a given time block is labeled as changed.
	\item Inter-time block: Any feature value between time block $ i $ and $ j $ at the fixed location for a given feature that satisfies:
	\begin{equation}
		\label{eq:inter_label}
		\frac{|\norSymScript{\hat{\mu}}{i}{} - \norSymScript{\hat{\mu}}{j}{}|}{3\cdot\sqrt{\norSymScript{\hat{{\sigma}}}{i}{2} + \norSymScript{\hat{{\sigma}}}{j}{2}}} \ge 1 \nonumber
	\end{equation}
	is labeled as changed.
\end{itemize}
The inter-time block status of the feature is computed by comparing to the first one (see \figPref\ref{subfig:rss_var_status_26_1} and \figPref\ref{subfig:rss_var_status_26_8}). Those labeled statuses are used as the ground truth for evaluating the change detection approach.

\subsubsection{Change detection result}
\begin{figure}[!h]
	\centering
	\subfloat[Time block 3]{\label{subfig:auc_spa_tms_blk_3}
		\includegraphics[width=0.3\linewidth]{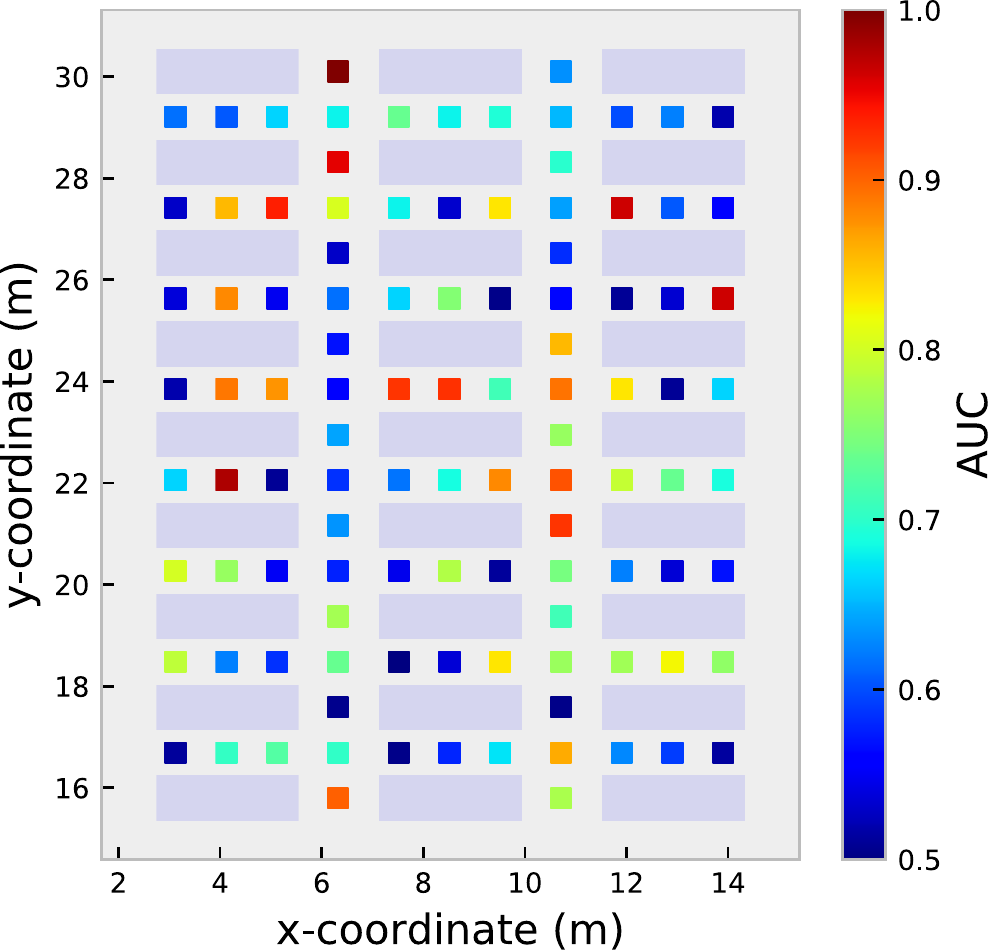}}\hspace{2ex}
	\subfloat[Time block 8]{\label{subfig:auc_spa_tms_blk_8}
		\includegraphics[width=0.3\linewidth]{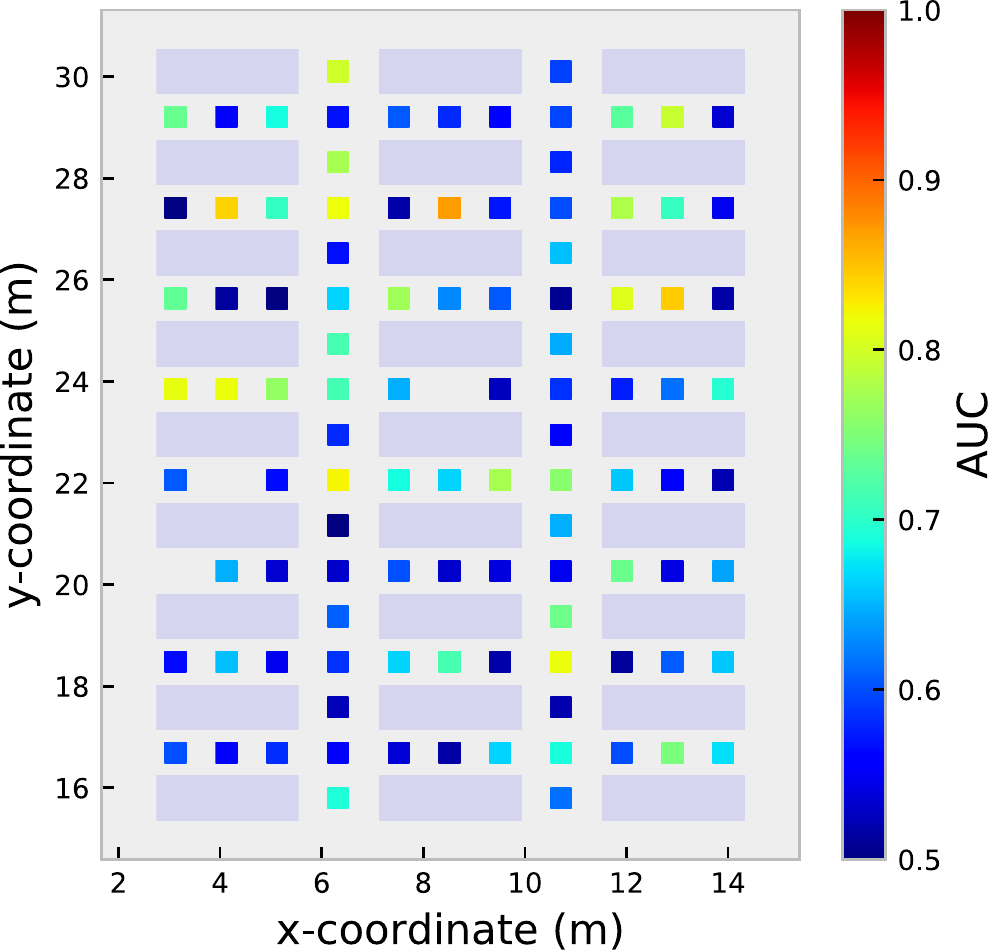}}\hspace{2ex}
	\subfloat[Time block 13]{\label{subfig:auc_spa_tms_blk_13}
		\includegraphics[width=0.3\linewidth]{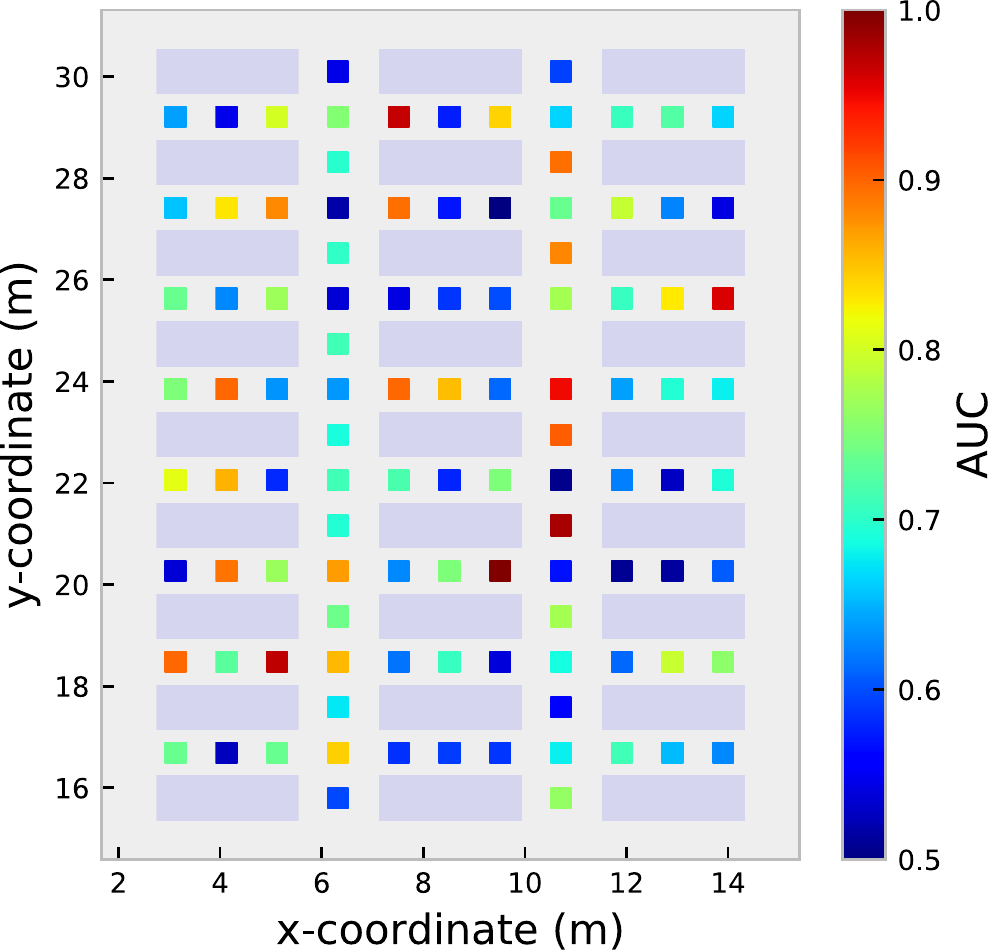}}\\
	\subfloat[Time block 18]{\label{subfig:auc_spa_tms_blk_18}
		\includegraphics[width=0.3\linewidth]{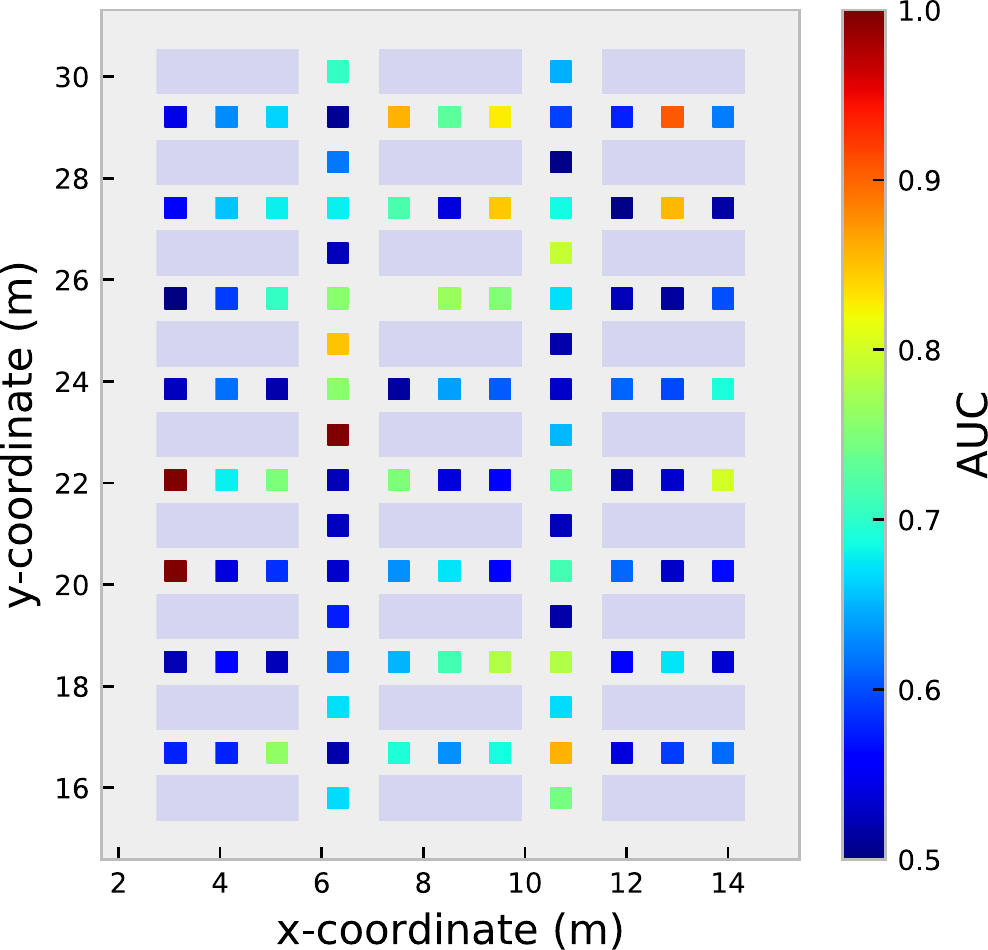}}\hspace{2ex}
	\subfloat[Time block 23]{\label{subfig:auc_spa_tms_blk_23}
		\includegraphics[width=0.3\linewidth]{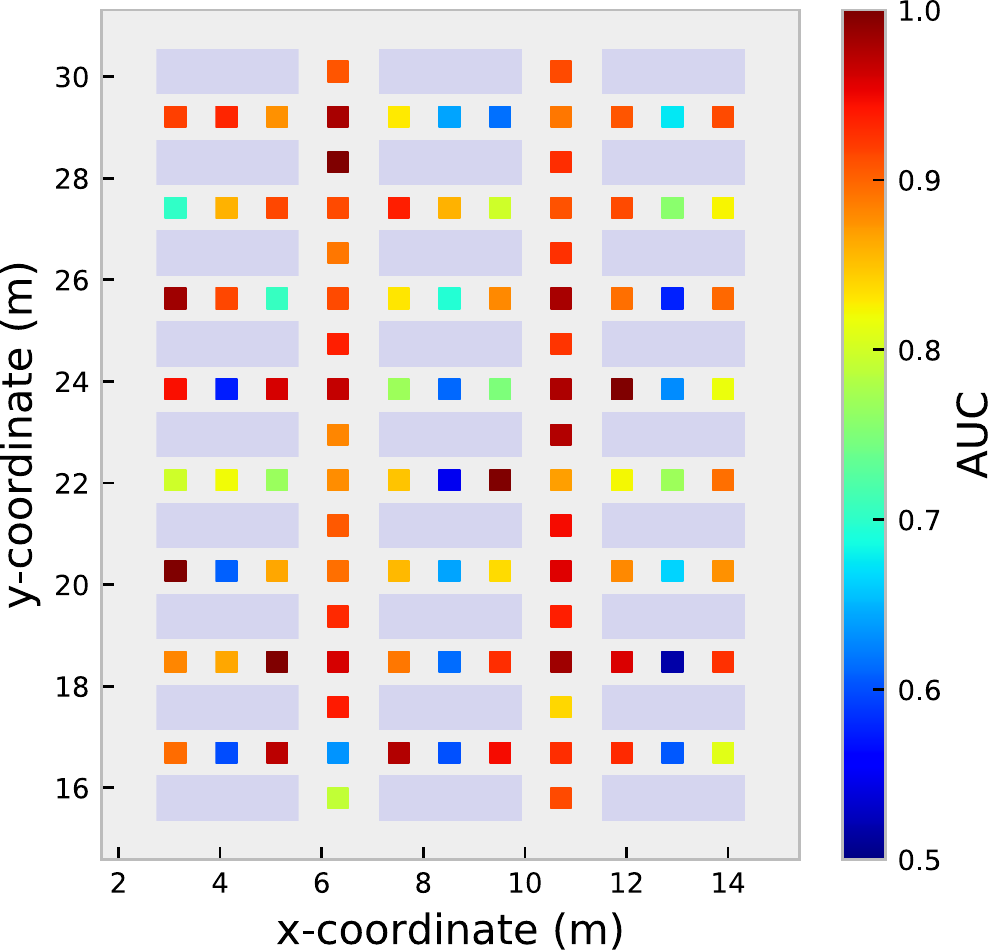}}\hspace{2ex}
	\subfloat[Time block 28]{\label{subfig:auc_spa_tms_blk_28}
		\includegraphics[width=0.3\linewidth]{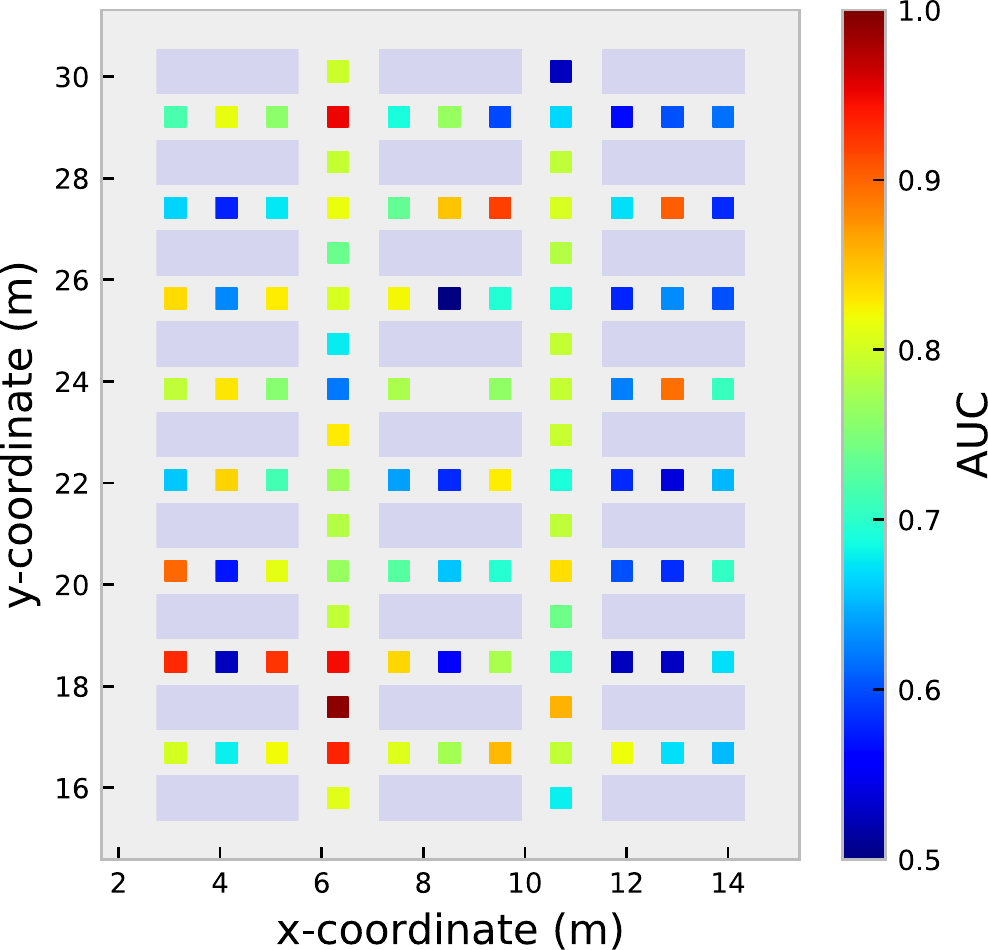}}
	\caption{The \acs{auc} over space of varies time blocks.}
	\label{fig:rep_spa}
\end{figure}

We apply our proposed feature-wise change detection and robust positioning approach to the long-term data and obtain a sequence of detected status of each feature location-wisely. \figPref\ref{subfig:rss_est_status_26_1} and \figPref\ref{subfig:rss_est_status_26_8} show two examples of the estimated change belief at the fixed location. The variability of the \acs{rss} values of \acs{ap} 1 suggests that directly adapting the \acs{rfm} with newly measured data without priorly determination is not benefit for keeping it up-to-date. Because the variations of the \acs{rss} are big contributed from varies sources (\eg device heterogeneity or collecting orientation) which are not essential motivations for updating the \acs{rfm}. Our estimation of the feature status is close to those labeled status by computing the \acs{auc} values. For \acs{ap} 1 and \acs{ap} 2, the \acs{auc} is 0.88 and 0.92, respectively. In addition, we found out that the estimated change belief is robust to the non-essential variations of the feature values: i) in case that \acs{ap} 1 is labeled as changed from time block 5 to 7 because the under-estimation of the variations, our approach is not strongly affected by them; and ii) our approach is capable of recognizing the changes in case that \acs{ap} 8 is mislabeled as stable from time block 15 to 16 due the over-estimation of the variation. Furthermore, we illustrate the change detection performance at different locations for all features using \acs{auc} values in \figPref\ref{fig:rep_spa}. From the results, we can conclude that: i) the change detection performance varies over space and there is no clear dependence between the \acs{auc} value and location;  and ii) the change detection performance varies over time. There is no obvious pattern of the temporal dependence though the average \acs{auc} at time block 23 is much higher than that of other five time blocks. According to \citep{mendoza2018long}, there is an upgrade of \acs{wlan} around time block 21, which causes large changes of features. Though the long-term collected dataset is not ideal for validating the proposed method because the pattern of the changes is unknown and is not abundant (mostly from missing to measurable or vice versa), it demonstrates the change distinguish capability of our approach.

\section{Conclusion}\label{sec:conclusion}
This paper concentrates on improving the quality of the \acf{rfm} for \acl{fbp} by using \acf{ransac}-like approach to jointly detect the feature-wise changes and estimate the position. The detected status of the features can be exploited for accumulatively and tactically update the \acs{rfm}, \eg only adapting the part of \acs{rfm} where accumulatively reported changes have occurred. The proposed method is self-supervised, \ie it does not require extra positioning system for providing the (approximate) ground truth location. Our extra-positioning-free solution is achieved by making full use of the redundancy of the opportunistically measurable location-relevant signal. Arbitrarily sampled features from the online measurement are used for estimating intermediate locations. The candidate locations that are closest to the ground truth one (though unknown at online positioning stage) are identified by using a threshold-free indicator value, \acf{mji} via comparing the online measurement to that of the expected measurement at the intermediate locations which is retrieved from the world model of the \acs{rfm}. In this way, a robust estimation of user's location can be calculated using these candidate locations and the feature-wise change status is able to be approximated using the estimated variability of features and the world model of the \acs{rfm}.

In order to validate our proposal, we have carried out two experimental analysis on i) simulated changes and ii) long-term collected dataset \citep{mendoza2018long}. On the simulated dataset, we can identify different simulated changes and the average value of the area under the \acf{roc} curve (\acs{auc}) is about 0.9, \ie the achievable change detection accuracy is about 90\%. Meanwhile, the positioning accuracy can be enhanced by dropping out features that are detected as change when performing the fingerprint matching with the \acs{rfm}. The improvement of positioning accuracy within 2~m and 4~m are up about 20\% and 10\% comparing to that of matching the full measurement to the \acs{rfm}. Regarding the long-term dataset, we use it for validating the generalization capability of our approach. Though there are no apparent dependences between the change detection performance and spatial and temporal variations on the long-term dataset, we have found out that our approach can detect the changes over time and the average \acs{auc} value is about 85\%. For future work, we would like to apply our approach to monitoring and detecting the feature changes in the real \acf{fips} and use those cumulated detected change information for updating the \acs{rfm}.

\section*{Acknowledgment}
The Chinese Scholarship Council has supported C. Zhou during his PhD studies at ETH, Z\"urich. Thanks for the inputs from Dr.~Andreas Wieser and Zan Gojcic.

\newpage
\appendix

\section{The world model of \acs{rfm} - \acf{ks}  and its variant}\label{subsec:ks_variant}
Herein we apply \acs{ks}, \ie kernel regression as the representation of the world model $ \funSym{F} $. Firstly we briefly describe the \acs{ks} and then introduce a variant version of it for reducing the computational complexity.

\subsection{Fundamentals of \acs{ks}}
\acs{ks} can be formulated as an optimization problem in \acf{rkhs} $ \funSym{H}^{\texSym{K}} $:
\begin{equation}
\label{eq:ks}
\funSym{F}^*=\underset{\funSym{F}\in\funSym{H}^{\texSym{K}}}{\operatorname{argmin}}{\,\,\sum_{i=1}^{|\setSym{F}|} \funSym{d}(\funSym{F}(\vecSymScript{l}{i}{}), \setSymScript{O}{i}{}) + \sigma^2\|\funSym{F}\|_{\funSym{H}^{\texSym{\texSym{K}}}}^{2}}
\end{equation}
where $ \funSym{d} $ is a dissimilarity metric, \ie $ \funSym{d}: \setSym{O}\times\setSym{O}\mapSym\varmathbb{R} $ and $ \sigma^2 $ is the regularization parameter. The solution of the above problem (treated as the world model of the \acs{rfm}) is:
\begin{equation}
\label{eq:sol_ks}
\funSym{F}^*(\vecSym{l}) = \vecSymScript{z}{\mathbf{l}}{'}(\vecSym{G}^{'}\vecSym{G} + \sigma^2\vecSymScript{I}{|\setSym{F}|}{})^{-1}\vecSym{o}
\end{equation}
where $ \vecSymScript{I}{|\setSym{F}|}{} $ is the identity matrix of size $ |\setSym{F}| $, $ \vecSym{o} \in\varmathbb{R}^{|\setSym{F}|}$ ($ \vecSym{o}_i = \setSymScript{O}{i}{}, i\in\intSet{|\setSym{F}|}$) and $ \vecSymScript{z}{\mathbf{l}}{} \in\varmathbb{R}^{|\setSym{F}|}$ is a column vector:
\begin{equation}
\label{eq:ks_z}
\vecSymScript{z}{\mathbf{l}}{}  = [\funSym{K}(\vecSym{l},\vecSymScript{l}{1}{}); \funSym{K}(\vecSym{l}, \vecSymScript{l}{2}{});
\cdots; \funSym{K}(\vecSym{l}, \vecSymScript{l}{|\setSym{F}|}{})]
\end{equation}
where $ \funSym{K}(\cdot, \cdot) $ is the kernel function. $ \vecSym{G} $ is Gram-matrix of a given kernel $ \funSym{K}(\cdot, \cdot)$. More details about \acs{ks} can be found, \eg in \cite{Berlinet2011}.

\subsection{The computational complexity of \acs{ks}}
In this paper, the world model $ \funSym{F} $ is used for pre-processing the collected raw measurements, for estimating the observation at the \acl{fbp} estimated location within a query radius and for long-term updating of the \acs{rfm}. In the middle case, it might need to be executed online (\ie real time or on a low-power and computational-resource mobile device). It is thus important that \acs{ks} can be executed in a low computational complexity. As the solution shown in \eqref{eq:sol_ks}, the computational complexity of \acs{ks} (mainly from the matrix inversion) is $ \compComp{|\setSym{F}|^{3}} $ ($ \compComp{|\setSym{F}|^{2.373}} $ using optimized algorithms) \cite{Cormen:2009:IAT:1614191} and $ \compComp{|\setSym{F}|^{2}} $ \acs{wrt} the time and storage, respectively. The time cost increases almost cubically \acs{wrt} the number of measurements contained in \acs{rfm}. It is expensive for executing \acs{ks} online in case of the size of the \acs{rfm} (\ie the area of the \acs{roi}) is big.

\subsection{Variant of \acs{ks}}
In order to reduce the computational complexity of \acs{ks}, especially for the online application, we apply a variant to \acs{ks} for making its computational complexity to a fix and adaptable level. Instead of using all measurements stored in the \acs{rfm} for solving the \acs{ks} problem, only a subset of the \acs{rfm} in the closeness of a given location is used, a queried \acs{rfm} $ \setSymScript{F}{i}{KS, u} (\norSymScript{r}{}{KS}) $ is extracted using only the reference points located in the given radius $ \norSymScript{r}{}{KS} $ and is defined as $ \norSymScript{r}{}{KS}:= \norSymScript{\lambda}{}{KS} \cdot \norSymScript{\lambda}{}{LS} $. The scale factor $  \norSymScript{\lambda}{}{KS} \in \varmathbb{R}^{+}  $ and $  \norSymScript{\lambda}{}{LS} $ is the length scale of the kernel.

As shown in the two figures in the first column of \figPref\ref{fig:raw_ks_interp_r12}, the raw measurements of the signals from \acs{wlan} \acsp{ap} are very noisy. We applied \acs{ks} as a pre-processing for filtering out the noisy of the raw measurements as shown in the next two columns of the plots in the figure. The output from \acs{ks} is much less noisy comparing to the raw measurements. However, a over-smoothed \acs{rfm} is not realistic because the spatial distribution of the features strongly correlated with the indoor environments (\eg the structure of the building or the furniture), especially in case of using radio frequency signals such as \acs{wlan} as the fingerprints. Because the propagation of the radio frequency is highly influenced by the obstacles that the signals penetrate through. As suggested in \cite{10.1007/978-3-642-32645-5_18}, a trade-off has to be found for balancing the discontinuity and smoothness of the spatial distribution of the value of the feature. Herein in the algorithm of \acs{ks}, a parameter $ \norSymScript{\lambda}{}{LS} $ of Mat\'ern kernel in \eqref{eq:matern_ks} is used for balancing this. We have tried different values of $ \norSymScript{\lambda}{}{LS} $ and qualitatively evaluate the balance between the discontinuity and smoothness of the spatial distribution. We came out with fixing $ \norSymScript{\lambda}{}{LS}=1 $ for \textit{HIL-R1} and \textit{HIL-R2}. In addition, comparing to the raw measurements collected at two different times (row-wise comparison in \figPref\ref{fig:raw_ks_interp_r12}), the measurability of an \acs{ap} does not change apparently. However, the value changes of the measurement are obvious and various within the \acs{roi}. The values of the feature measured in \textit{HIL-R2} are about 5 to 10 dB higher than that of \textit{HIL-R1}.

\begin{figure}[!htb]
	\centering
	\subfloat[\acs{mae}]{
		\label{subfig:radius_vs_residual}
		\includegraphics[width=0.3\linewidth]{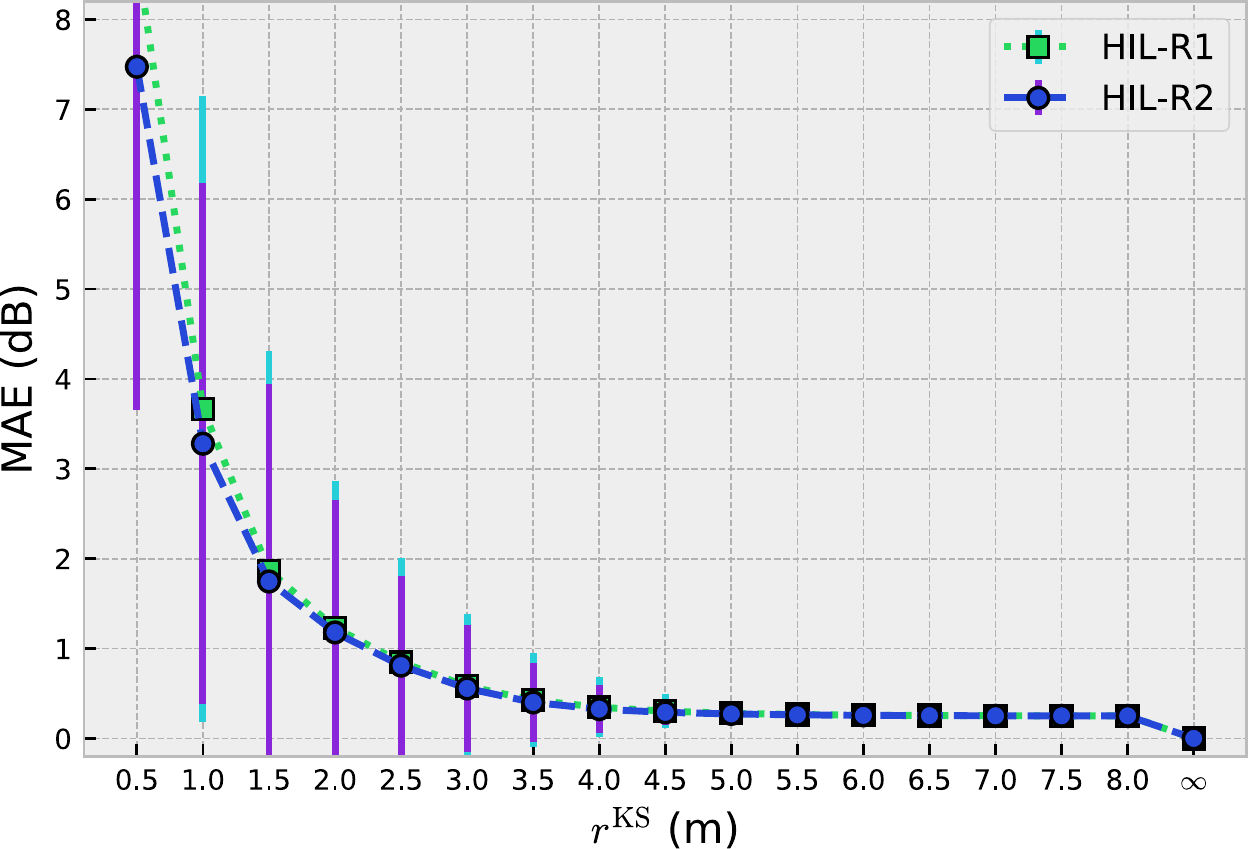}}\hspace{1ex}
	\subfloat[Process time]{
		\label{subfig:radius_vs_pTime}
		\includegraphics[width=0.3\linewidth]{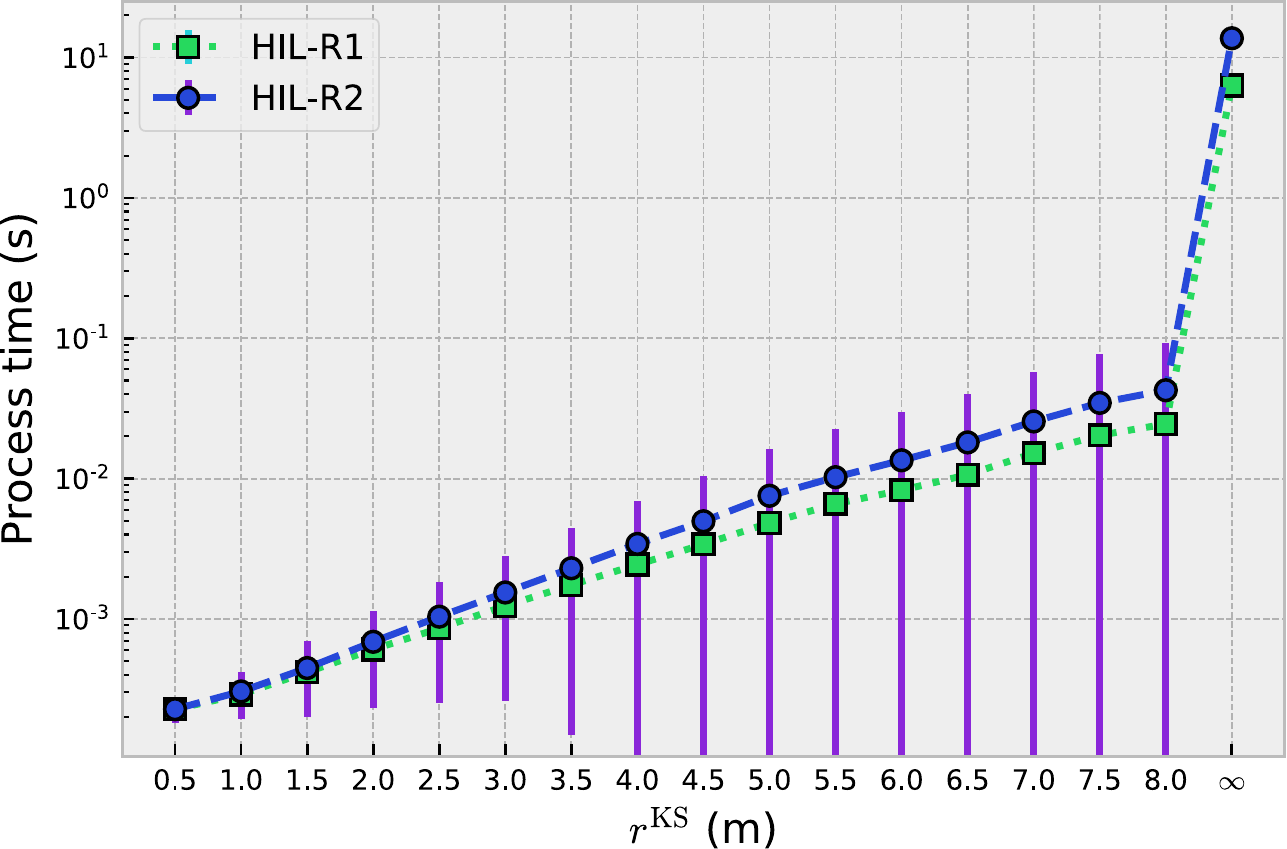}}\hspace{1ex}
	\subfloat[Storage space]{
		\label{subfig:radius_vs_storage}
		\includegraphics[width=0.3\linewidth]{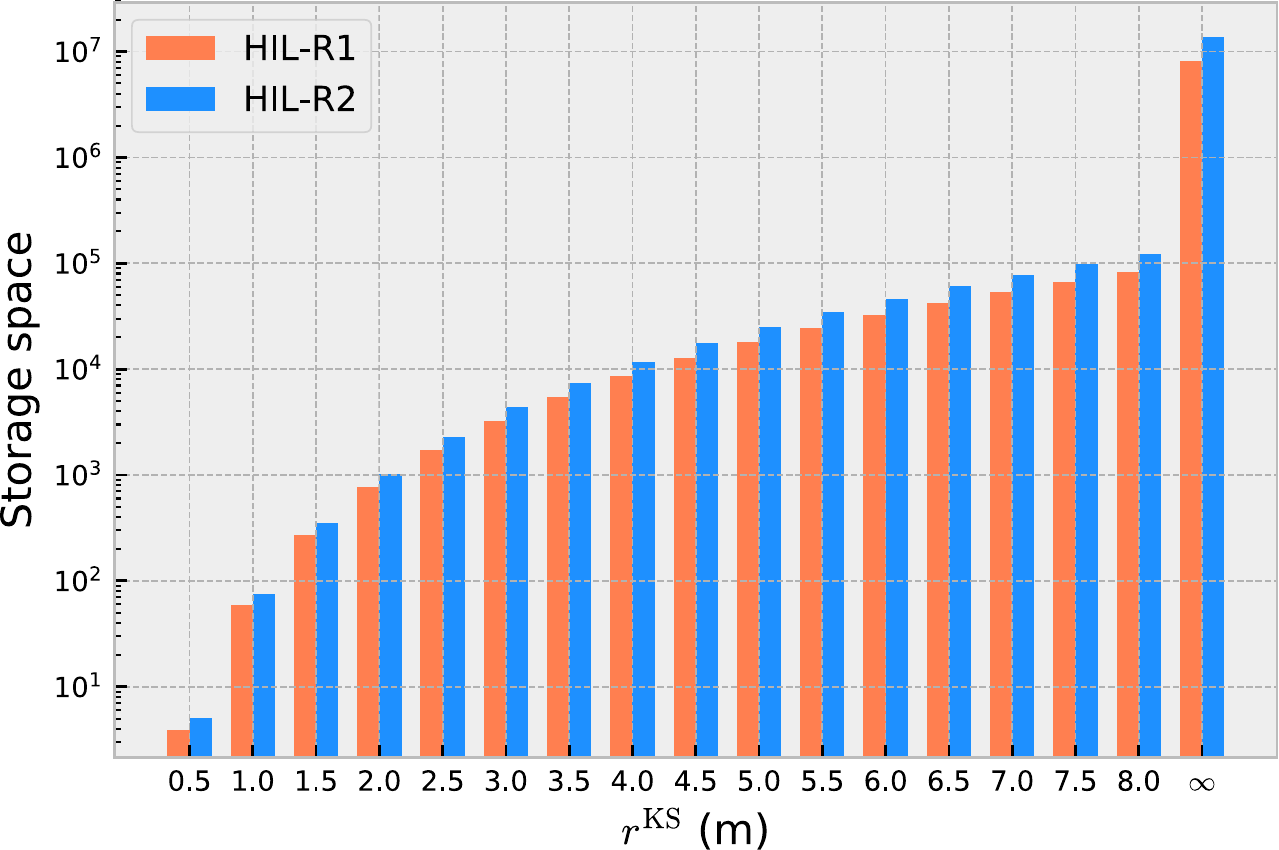}}
	\caption{The influence of the value of $ \norSymScript{r}{}{KS} $ on processing time, storage space and residuals. The value of $ \norSymScript{k}{}{}=\infty $ means that using all the training examples for \acs{ks}.}
	\label{fig:radius_vs_q_ks}
\end{figure}

\begin{figure}[!htb]
	\centering
	\subfloat[\textit{HIL-R1}]{
		\label{subfig:res_hil_r1}
		\includegraphics[width=0.3\linewidth]{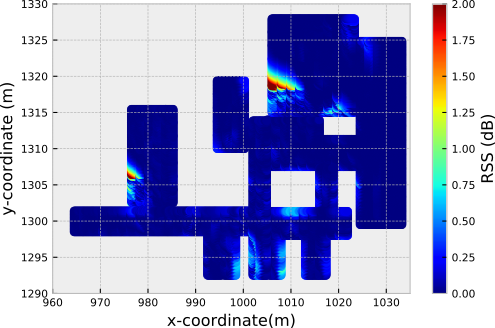}}\hspace{3ex}
	\subfloat[\textit{HIL-R2}]{
		\label{subfig:res_hil_r2}
		\includegraphics[width=0.3\linewidth]{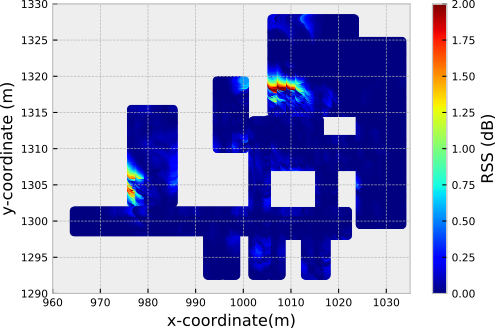}}
	\caption{An example (\acs{mac} is \textit{9c:50:ee:09:61:d0}) of the absolute value of residual between the interpolated \acs{rfm} using all the training samples  and the one using querying-based \acs{ks} ($ \norSymScript{r}{}{KS}=5.0 $).}
	\label{fig:res_q_ks}
\end{figure}

As aforementioned in \secPref\ref{subsec:config_exp}, we use the variant version \acs{ks} for reducing the computational complexity. We evaluate the impact of the querying radius $ \norSymScript{r}{}{KS} $ on the computational complexity \acs{wrt} the processing time and storage space, as well as the interpolation performance (defined as \acf{mae} of the residuals)  by calculating the residual between the estimated values using queried \acs{rfm} and the ones using the whole one. As shown in \figPref\ref{fig:radius_vs_q_ks}, the \acs{mae} of the interpolation value using \acs{ks} decreases by about 400 times as the value of $ \norSymScript{k}{}{KS} $ increases from 0.5~m to 8~m. Meanwhile, the processing time and storage cost increase by about 1000 times. In order to balance the value of residual and the computational complexity, we find out that the residual has no apparent reduction and the overall residual is low (\figPref\ref{fig:res_q_ks}) when $ \norSymScript{r}{}{KS} $ is larger than 5m. At the querying radius value for \acs{ks}, the process time is around tens of milliseconds.

\section{Formulation of Mat\'ern kernel}\label{subsec:matern}
Mat\'ern kernel $ \funSym{K}^{\mathrm{M}}_{v}(d) $ is formulated as \footnote{In this subsection, the symbol has only the pure mathematical meaning and is independent from what we defined in the other sections.} \cite{Rasmussen:2005:GPM:1162254}:
\begin{equation}
\label{eq:general_matern}
\funSym{K}^{\mathrm{M}}_{\nu}(d)=\sigma^2\frac{2^{1-\nu}}{\Gamma}\Big(\sqrt{2\nu}\frac{d}{\norSymScript{\lambda}{}{LS}}\Big)^\nu\funSym{B}_v\Big(\sqrt{2\nu}\frac{d}{\norSymScript{\lambda}{}{LS}}\Big)
\end{equation}
where $ \Gamma() $ is the Gamma function, $ \funSym{B}_\nu() $ is the modified Bessel function of the second kind. $ \sigma $, $ \norSymScript{\lambda}{}{LS} $ and $ \nu $ are non-negative parameters. In case of $ \nu\rightarrow\infty $, Mat\'ern kernel converges to the squared exponential kernel. Herein we set $ \nu =3/2 $ (a default value suggested in \cite{scikit-learn}), \eqref{eq:general_matern} results in:
\begin{equation}
\label{eq:matern_ks}
\funSym{K}^{\mathrm{M}}_{3/2}(d)=\sigma^2\Big(1 + \frac{\sqrt{3}d}{\norSymScript{\lambda}{}{LS}}\Big)\operatorname{exp}\Big(-\frac{\sqrt{3}d}{\norSymScript{\lambda}{}{LS}}\Big)
\end{equation}
The key parameter for the application of spatial interpolation is $ \norSymScript{\lambda}{}{LS} $, which determines the contributions of each reference measurement at different distance of a given location to the interpolated value. We fixed $ \norSymScript{\lambda}{}{LS}=1 $ in the experimental analysis for \textit{HIL-R1} and \textit{HIL-R2}.

\bibliographystyle{apalike}
\bibliography{../reference/rfm_update_ransac_sensors}



\end{document}

%% file: cz_acronyms.tex
\acrodef{vc}[VC]{Vapnik-Chervonenkis}
\acrodef{knn}[$ k $NN]{$ k $ nearest neighbors}
\acrodef{lbs}[LBS]{location-based service}
\acrodef{ilbs}[ILBS]{indoor location-based service}
\acrodef{fips}[FIPS]{fingerprinting-based indoor positioning system}
\acrodef{fbp}[FbP]{feature-based positioning}
\acrodef{gnss}[GNSS]{global navigation satellite system}
\acrodef{ips}[IPS]{indoor positioning system}
\acrodef{rfid}[RFID]{radio frequency identification}
\acrodef{uwb}[UWB]{ultra wideband}
\acrodef{wlan}[WLAN]{wireless local area network}
\acrodef{rss}[RSS]{received signal strength}
\acrodef{ap}[AP]{access point}
\acrodef{roi}[RoI]{region of interest}
\acrodef{lasso}[LASSO]{least absolute shrinkage and selection operator}
\acrodef{rfm}[RFM]{reference fingerprint map}
\acrodef{rfm1}[RFM]{reference feature map}
\acrodef{map}[MAP]{maximum a posteriori}
\acrodef{cpa}[CPA]{cumulative positioning accuracy}
\acrodef{mse}[MSE]{mean squared error}
\acrodef{tp}[TP]{test position}
\acrodef{wrt}[w.r.t.]{with respect to}
\acrodef{mac}[MAC]{media access control}
\acrodef{imu}[IMU]{inertial measurement unit}
\acrodef{foba}[adaFoBa]{adaptive forward-backward greedy}
\acrodef{rp}[RP]{reference point}
\acrodef{mji}[MJI]{modified Jaccard index}
\acrodef{ble}[BLE]{Bluetooth low energy}
\acrodef{oil}[OIL]{organic indoor localization}
\acrodef{will}[WILL]{wireless indoor localization without site survey}
\acrodef{hiwl}[HIWL]{hidden Morkov model-based indoor wireless localization}
\acrodef{svm}[SVM]{support vector machine}
\acrodef{lda}[LDA]{linear discriminant analysis}
\acrodef{sop}[SoP]{signal of opportunity}
\acrodef{ks}[KS]{Kolmogorov-Smirnov}
\acrodef{ecdf}[ECDF]{empirical cumulative distribution function}
\acrodef{cdm}[CDM]{compound dissimilarity measure}
\acrodef{rmse}[RMSE]{root mean squared error}
\acrodef{rcdm}[RCDM]{relatively weighted compound dissimilarity measure}
\acrodef{acdm}[ACDM]{averagely weighted compound dissimilarity measure}
\acrodef{cv}[CV]{cross validation}
\acrodef{pdf}[PDF]{probability density function}
\acrodef{laafu}[LAAFU]{localization with altered APs and fingerprint update}
\acrodef{amd}[AMD]{average mutual distance}
\acrodef{rpc}[RPC]{recall precision curve}
\acrodef{auc}[AUC]{area under the ROC curve}
\acrodef{bo}[BO]{Bayesian optimization}
\acrodef{roc}[ROC]{receiver operating characteristic}
\acrodef{rocNick}[ROC]{relative operating characteristic curve}
\acrodef{svr}[SVR]{support vector regression}
\acrodef{smbo}[SMBO]{sequential model-based optimization}
\acrodef{iou}[IoU]{intersection over union}
\acrodef{gp}[GP]{Gaussian process}
\acrodef{ei}[EI]{expected improvement}
\acrodef{tpr}[TPR]{true positive rate}
\acrodef{fpr}[FPR]{false positive rate}
\acrodef{ks}[KS]{kernel smoothing}
\acrodef{rkhs}[RKHS]{reproducing kernel Hilbert space}
\acrodef{qcd}[QCD]{querying-based change detection}
\acrodef{mae}[MAE]{mean absolute error}
\acrodef{kde}[KDE]{kernel-based density estimation}
\acrodef{pdr}[PDR]{pedestrian dead reckoning}
\acrodef{rsm}[RSM]{random subspace method}
\acrodef{ransac}[RANSAC]{random sampling consensus}
\acrodef{dbscan}[DBSCAN]{density based spatial clustering of applications with noise}
\acrodef{amise}[AMISE]{asymptotic mean integrated squared error}
\acrodef{epdf}[EPDF]{empirical probability density function}
\acrodef{cr}[CR]{confidence region}
\acrodef{ce}[CE]{circular error}
\acrodef{tp}[TP]{true positive}
\acrodef{fn}[FN]{false negative}
\acrodef{tn}[TN]{true negative}
\acrodef{fp}[FP]{false positive}
\acrodef{tpr}[TPR]{true positive rate}
\acrodef{tnr}[TNR]{true negative rate}
\acrodef{plsr}[PLSR]{partial least square regression}
\acrodef{gpr}[GPR]{Gaussian process regression}
\acrodef{std}[STD]{standard deviation}
\acrodef{bic}[BIC]{Bayesian information criterion}
\acrodef{ugv}[UGV]{unmanned ground vehicle}
\acrodef{ldplm}[LDPLM]{Log-distance path loss model}
\acrodef{slam}[SLAM]{simultaneous localization and mapping}
\acrodef{ekf}[EKF]{extended Kalman filter}
\acrodef{mad}[MAD]{median absolute deviation}
\acrodef{mle}[MLE]{maximum likelihood estimation}
\acrodef{mcd}[MCD]{minimum covariance determinant}
\acrodef{tf}[TF]{termination flag}

%% file: cz_command.tex
\newcommand{\todo}[1]{\btext{\textit{To be done: #1}}}
\newcommand{\tocite}{\btext{[CITE] }}
\newcommand{\toref}{\btext{[REF-FIG-TAB] }}
\newcommand{\appRef}{Appendix }
\newcommand{\secPref}{Section }
\newcommand{\figPref}{Fig.}
\newcommand{\tabPref}{TABLE }
\newcommand{\eg}{e.g. }
\newcommand{\ie}{i.e. }
\newcommand{\etal}{et~al.}
\newcommand{\rhl}[1]{\textcolor{red}{\hl{#1}}}
\newcommand{\btext}[1]{\textcolor{blue}{{#1}}}
\newcommand{\colRev}[1]{\btext{#1}}
\newcommand{\colRevOff}[1]{{#1}}
\newcommand{\convSetSym}[1]{\varmathbb{#1}}
\newcommand{\setSym}[1]{\mathbbm{#1}}
\newcommand{\vecSym}[1]{\mathbf{#1}}
\newcommand{\funSym}[1]{\mathpzc{#1}}
\newcommand{\textSym}[1]{\mathrm{#1}}
\newcommand{\mapSym}{\mapsto}
\newcommand{\texSym}[1]{\mathrm{#1}}
\newcommand{\intSet}[1]{\{1, 2, \cdots, #1\}}
\newcommand{\setSymScript}[3]{\setSym{#1}_{#2}^{\texSym{#3}}}
\newcommand{\vecSymScript}[3]{\vecSym{#1}_{#2}^{\texSym{#3}}}
\newcommand{\norSymScript}[3]{#1_{#2}^{\texSym{#3}}}
\newcommand{\compComp}[1]{\mathcal{O}(#1)}
\newcommand\blankpage{%
	\null
	\thispagestyle{empty}%
	\addtocounter{page}{-1}%
	\newpage}
\newlength{\tempdima}
\newcommand{\rowname}[1]
{\rotatebox{90}{\makebox[\tempdima][c]{{#1}}}}

\renewcommand{\thesubfigure}{\alph{subfigure}}
\newcommand{\mycaption}[1]
{\refstepcounter{subfigure}\textbf{(\thesubfigure) }{\ignorespaces #1}}